%% file: main.tex
\documentclass[11pt,dvipsnames]{article}
    
\input{preamble.tex}

\begin{document}

\maketitle

\begin{abstract}
    \input{sections/abstract.tex}

\end{abstract}

\input{sections/intro.tex}

\input{sections/problem_setup.tex}

\input{sections/proposed_method.tex}
\input{sections/results.tex}

\section{Conclusions}
In this work, we present an explicit study of repository-level code-editing retrieval. We propose a
method for training models specifically designed for this task and demonstrate that existing
retrieval models are suboptimal in this setting. We identify that this problem differs from the
traditional constrastive representation learning problem and propose a loss function that
substantially improves the performance of retrieval models compared to using standard contrastive
losses. Further, incorporating code context from neighbours in the call graph gives an additional
boost in performance. We speculate that further improvement may come with better strategies to
select relevant neighbours, \eg by leveraging topological properties of the call graph
\citet{tsourakakis2017predicting, chiang2014prediction, cesa2012correlation}.
We hope this work highlights the importance of retrieval for code editing and inspires further
research to advance models and techniques in this domain.

\section*{Limitations}
Our work can be extended in several ways. For instance, our experiments are restricted to Python, and are based on two datasets - SWE-bench and LCA. At the time of the submission they were the only datasets that
allowed for repository level code retrieval problems. Recently, a multi-language dataset called SWE-PolyBench\citep{polybench} has been released and we plan to include it in our future work on this topic.

Furthermore, SWE-bench Verified requires to modify, for the large part, one file in each test case. Thus, file recall can be very high. LCA features edits in multiple files, and thus is a better repository to benchmark.
The data preparation and chunking step,
while extensible, can be expensive to implement for languages other than Python.

Our investigation has been limited to encoder models for their feature prediction abilities. Several
modern LLMs have been modified to output feature
embeddings\citep{behnamghader2024llmvec,tao2024llms}. It is also likely that these LLMs
trained on large corpus can provide better baseline performance, but training them requires more resources.

\bibliography{custom}

\appendix

\begin{appendices}
    \input{sections/appendix.tex}
\end{appendices}

\end{document}

%% file: preamble.tex
\usepackage[final]{acl}
\usepackage{times}
\usepackage{latexsym}

\usepackage[T1]{fontenc}

\usepackage[utf8]{inputenc}

\usepackage{microtype}

\usepackage{inconsolata}

\usepackage{graphicx}
\usepackage{amsmath}
\usepackage{latexsym}
\usepackage{amssymb}
\usepackage{booktabs}
\usepackage[textsize=tiny]{todonotes}
\usepackage{color, colortbl}
\usepackage{pgf,caption,enumitem}
\usepackage[title]{appendix}
\usepackage{tabularx}
\usepackage{wrapfig, multirow}
\usepackage{bookmark}
\usepackage{float}

\hypersetup{
    colorlinks,linktoc=all,
    linkcolor=BrickRed,
    citecolor=PineGreen,
    filecolor=Mulberry,
    urlcolor=NavyBlue,
    menucolor=BrickRed,
    runcolor=Mulberry,
}

\usepackage[capitalize,noabbrev]{cleveref}
\usepackage{amsthm}
\theoremstyle{plain}

\theoremstyle{definition}

\theoremstyle{remark}

\definecolor{Gray}{gray}{0.9}

\DeclareUnicodeCharacter{2212}{-}

\NewCommandCopy{\ORIcitep}{\citep}
\DeclareRobustCommand{\citep}{\leavevmode\unskip~\ORIcitep}

\NewCommandCopy{\ORIcitet}{\citet}
\DeclareRobustCommand{\citet}{\leavevmode\unskip~\ORIcitet}
\def\cody{CoRet\xspace}
\def\localtitle{\cody: Improved Retriever for Code Editing}
\def\localauthors{Fabio Fehr\thanks{Work done during an internship at AWS, Berlin.} \\
    EPFL \& Idiap Research Institute \\
    \texttt{fabio.fehr@epfl.ch} \\
    \And
    Prabhu Teja Sivaprasad \\
    AWS AI Labs \\
    \texttt{prbuteja@amazon.de}
    \And
    Luca Franceschi\\
    AWS AI Labs \\
    \texttt{franuluc@amazon.de} \\
    \AND
    Giovanni Zappella \\
    AWS AI Labs \\
    \texttt{zappella@amazon.de}\\}
\title{\localtitle}
\author{\localauthors}

\usepackage{xspace}

\makeatletter
\DeclareRobustCommand\onedot{\futurelet\@let@token\@onedot}
\def\@onedot{\ifx\@let@token.\else.\null\fi\xspace}

\def\eg{\emph{e.g}\onedot} 
\def\ie{\emph{i.e}\onedot}

\makeatother

\input{defines.tex}

%% file: defines.tex
\newcommand{\repo}{\ensuremath{\mathcal{C}}\xspace}
\newcommand{\argtopk}{\mathop{\mathrm{arg\,top}\,k}}
\newcommand{\query}{\ensuremath{\mathbf{q}}\xspace}
\newcommand{\chunk}{\ensuremath{\mathbf{c}}\xspace}

\newcolumntype{H}{>{\setbox0=\hbox\bgroup}c<{\egroup}@{}}
\newcolumntype{Z}{>{\setbox0=\hbox\bgroup}c<{\egroup}@{\hspace*{-\tabcolsep}}}

%% file: sections/abstract.tex
In this paper, we introduce \cody, a dense retrieval model designed for code-editing tasks
that integrates code semantics, repository structure, and call graph dependencies. The model focuses
on retrieving relevant portions of a code repository based on natural language queries such as requests to implement  new
features or fix bugs. These retrieved code chunks can then be presented to a user or to
a second code-editing model or agent. To train \cody, we propose a loss function
explicitly designed for repository-level retrieval.
On SWE-bench and Long Code Arena's bug localisation datasets, we show that our model substantially
improves retrieval recall by at least $15$ percentage points over existing models, and ablate the design
choices to show their importance in achieving these results.

%% file: sections/intro.tex
\section{Introduction}
Code editing is an important task that often requires developers to make changes to code
repositories based on explicit natural language descriptions such as a GitHub pull request about a
bug, a new feature, or about an error in the code. Successful code editing requires correct
navigation and retrieval of relevant sections of the repository. This process demands a
repository-level understanding of the code functionality (semantics), organisation (repository
hierarchy), and relationships between various entities in the codebase (\eg runtime dependencies).
Retrieving multiple relevant code chunks is especially difficult in large, real-world
repositories\citep{jimenez2024swebench}. Retrieval is important for both coding agents and humans
and forms an important first step in the overall code editing process.

We find that existing pretrained encoder models perform poorly in repository-level retrieval for
code-editing. We conjecture the reason for this is threefold: the representations do not align the
problem statements to segments of code; the structure from the repository hierarchy is lost; and the
runtime dependencies within a repository are not captured. Models like CodeSage\citep{zhang2024code}
capture docstring-code relationships effectively. Yet, we find that this capacity does not readily
transfer to retrieval for code-editing tasks. Similarly, models like
CodeBERT\citep{feng-etal-2020-codebert}, GraphCodeBERT\citep{guo2021graphcodebert}, and UniXcoder
\citep{guo-etal-2022-unixcoder} focus on code semantics but miss repository-level structures,
limiting their generalisation to this task.

Incorporating additional context has been long known to improve the quality of
retrieval\citep{colbert2020khattab,zhu2024largelanguagemodelsinformation}. The nature of code
presents challenges in determining how to integrate context and what type of context to use. For
tasks like code completion and summarisation, using chunks from other files within the same
repository as context improves performance\citep{Bansal2021, ding-etal-2024-cocomic}.
\Citet{Bansal2023FunctionCG} argue that the call graph dependency structure is required to
understand the semantics of code and consider embedding code subroutines together. We follow this
direction.

Our work introduces a dense retrieval model for code-editing tasks. %
We show that
optimising directly the likelihood of the model to retrieve correct code sections brings significant
advantages over standard contrastive losses.
Moreover, we show how to effectively incorporate file hierarchy and call graph context. These
contributions lead to improvements of at least $15$ percentage points recall over the baseline
methods on SWE-bench and Long Code Arena.

%% file: sections/problem_setup.tex
\section{Problem setup}

\begin{figure}
    \centering
    \includegraphics[width=0.4\textwidth]{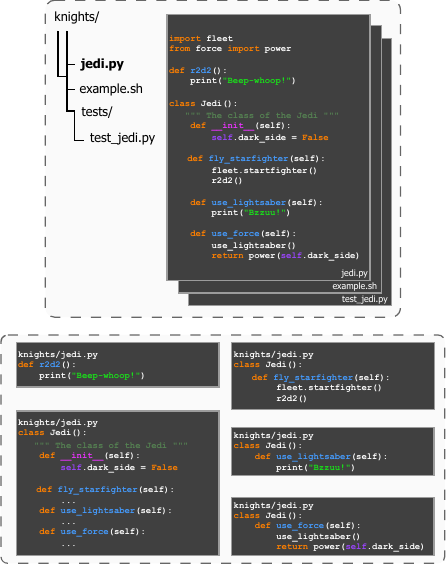}
    \caption{\textbf{Top}: Code repository before processing. \textbf{Bottom}: Code chunks after
        filtering and chunking.}
    \label{fig:data_processing}
\end{figure}

Representing a code repository in structurally-informed and semantically succinct units is
imperative for retrieval performance \citep{shrivastava2023repofusiontrainingcodemodels,
    zhang2023repocoder, liao2024a3codgenrepositorylevelcodegeneration}. An obvious atom of a repository
is a file. However, files may be particularly long and may lack semantic coherence. Therefore, we
further break down each file into its constituent functions, classes and methods of classes, as
illustrated in \cref{fig:data_processing} (bottom). We refer to each such atom as a \emph{chunk}.
Further details on the chunking procedure are available in Appendix~\ref{app:chunking}.

The objective of code retrieval is to return all parts of the codebase that are pertinent to the
current input. Let \repo represent a code repo as a set of atoms: $\repo = \{c_1, \cdots c_M\}$. We
represent with $q$ a natural text input that describes an issue in the repository. Our dataset
$\mathcal{D}$ consists of $N$ triplets of issue descriptions, associated codebase, and the ground
truth atoms \ie $\mathcal{D} = \{(q_i, \repo_i, \repo_i^*) \}_{i=1}^N$ where $\repo_i^* \subseteq
    \repo_i$. We refer to each triplet as an example, or instance. A retriever is a function $f: (q,
    c_i) \rightarrow [0, 1] \,\,\text{for}\,\, c_i \in \repo$ that assigns a numerical score to each of
the atoms in \repo. A ranking can be induced on the atoms by sorting the scores.  We parameterize
$f(q, c_i; \theta_q, \theta_c) = \mathrm{sim}(Q(q;\theta_q), C(c_i; \theta_c))$, where $Q$ and $C$
are embedding networks for $q$ and $c_i$ respectively, and `$\mathrm{sim}$' is the cosine similarity
between them. The ideal retriever induces a ranking that puts all the relevant atoms before the
irrelevant ones. In practice, we extract the top-$k$ most similar atoms to the query, denoted by
$\repo_F$, where $k \ll |\repo|$ \ie $\repo_F = \argtopk_{ c \in \repo} [f(q, c)]$. We leave the
dependence of  $\repo_F$ on $k$ implicit. For brevity, we represent $Q(q;\theta_q)$ with \query and
$C(c_i; \theta_c)$ with $\chunk_i$. During inference, given a (test) codebase and a query in natural
language, the top-$k$ most similar atoms retrieved are from the set $C = \{c_1, \dots, c_M\}$ where
$M$ is the number of atoms (chunks) in this codebase.

%% file: sections/proposed_method.tex
\section{Proposed method}
\label{sec:method}
Here we describe our method to train encoder models for the specific task of code retrieval.
\subsection{Training}

The goal of training the retrieval model $f$ is to learn an embedding space where the query and the
relevant code chunks have higher similarity
than irrelevant code chunks. Each instance $i \in [N]$, contains a single query $q_i$
and multiple ground truth code chunks $c^*_j$. These form positive pairs $(q_i ,
    c_j^*)$. The remaining pairs $(q_i , c_k)$, for $c_k\in \repo \setminus
    \repo^*$,  are negative pairs. We optimise the following loss function:
\begin{equation}
    \mathcal{L}(\theta) = \frac{1}{N}\sum^{N}_i \frac{1}{|\mathcal{C}_i^*|} \sum_{c^*\in \mathcal{C}_i^*}\log\dfrac{\exp{
            \left(\tfrac{\query_i \cdot \chunk^*} {\tau} \right)} }
    {\Gamma(\query, \mathcal{C}_i)},
    \label{eqn:loss_def}
\end{equation}
where $\theta=(\theta_q, \theta_c)$ are the parameters of the model and $\Gamma(\query,
    \mathcal{C})= \sum_{c\in\mathcal{C}} \exp{ \left(\tfrac{\query \cdot \chunk} {\tau} \right)}$ is the
normalising factor. This is the mean likelihood of retrieving the code chunk per model, and is akin
to the standard cross-entropy loss for multi-class classification.

Since the normalising factor involves a summation over all the chunks in a code repository (which
can be in the order of 10000 chunks), we implement an approximation by considering only a random
subset of instances:
\begin{equation}
    \tilde{\Gamma}(q, c^*, \mathcal{C}) =
    \exp{
        \left(\tfrac{\query_i \cdot \chunk^*}{\tau} \right)
    }
    +
    \sum_{c\in\mathcal{B}} \exp{
        \left(\tfrac{\query_i \cdot \chunk} {\tau} \right)}
    \label{eqn:final_loss}
\end{equation}
where $\mathcal{B}\subset \repo$ is random sample of within-instance negatives and $\tau$ is a
temperature parameter. Following \citet{zhang2024code}, we set $\tau=0.05$ throughout. Prior works
(see \cref{app:sec:related_work}) primarily use contrastive losses for feature learning, whereas we
apply a standard log-likelihood loss in this setting.

\subsection{Call graph context for code-editing}\label{subsec:cg_context}

\begin{figure}
    \includegraphics[width=\linewidth]{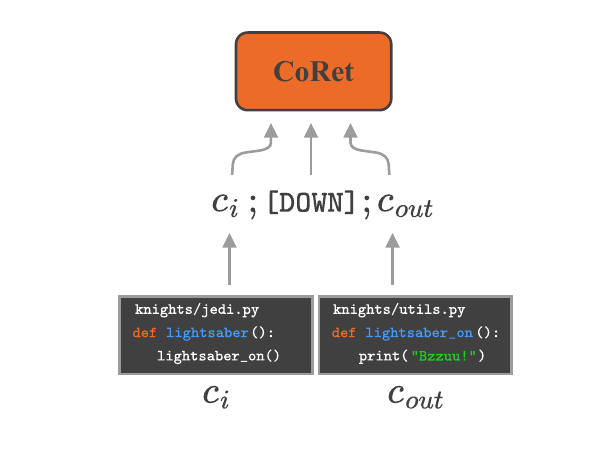}
    \caption{Given a function $c_i$ and its downstream neighbour $c_{\text{out}}$, we concatenate the
        strings as $c_i$; \texttt{[DOWN]}; $c_{\text{out}}$, including the special separation token, and
        fine-tune the model to obtain \textbf{CoRet}.}
    \label{fig:coret_illustration}
\end{figure}

Context has been shown to significantly improve the quality of
retrieval\citep{lewis2020,günther2024latechunkingcontextualchunk,pmlr-v162-borgeaud22a}.
A code repository has a natural relationship between its chunks endowed by the call
graph\citep{Ahn2009AWC, Bansal2023FunctionCG}. The neighbours in the call graph are the code
entities that are invoked by or invoke the current chunk of interest. We propose to enrich each
chunk $c_i$ in \repo with its call graph neighbours $\mathcal{N}(c_i)$. To incorporate this
information, we modify the chunk embedder $C$ to accept that information as $C(c_i,
    \mathcal{N}(c_i); \theta_c)$. Several implementations for this $C$ are possible: we retain the same
network as without the call graph information, but add the call graph chunks in the input itself \ie
$C([c_i; \mathcal{N}(c_i)]; \theta_c)$.  We also use only downstream neighbours by introducing a new
token \texttt{[DOWN]} that denotes this relationship. For instance, a chunk with one incoming and
one outgoing edge is represented as $c_i; \texttt{[DOWN]}; c_{out}$, where `$;$' denotes string
concatenation. See \cref{fig:coret_illustration} for an example of how the call graph context is fed
to the model. We follow BERT\citep{devlin-etal-2019-bert} in adding token segment type embeddings
\ie trainable embeddings that signify $c_i$ and $\mathcal{N}(c_i)$.

%% file: sections/results.tex
\section{Experiments}
\providecommand{\mathdefault}[1]{#1}

\subsection{Dataset}
For training, we consider repository-level code-editing problems which contain a language problem
statement $q$ and a code repository \repo from SWE-bench\citep{jimenez2024swebench}.
The ground truth pull requests are parsed to obtain the ground truth chunks $\repo^*$ which
correspond to the edited code chunks.
We evaluate our
trained models on SWE-bench Verified, and Long Code Arena (LCA) bug
localisation\citep{bogomolov2024longcodearenaset}.
We provide further dataset statistics in \cref{appendix:data}.

\subsection{Metrics}
We measure the performance of our model using standard retrieval metrics: recall@$k$ and mean
reciprocal rank (MRR). Recall@$k$ measures how many of the ground truth chunks in $\repo^*$ are
retrieved when $k$ most similar chunks are retrieved. MRR measures the minimum $k$ needed to
retrieve at least one correct code chunk; see \cref{app:sec:metrics} for formal
definitions. Additionally, when multiple chunks have to retrieved, as in the case of LCA, we show
the performance metric Perfect-Recall@$k$ which is a binary value for each instance if all correct
chunks were retrieved at $k$. This is better suited to measure improvements as partial retrievals
are not useful for subsequent code editing.
\begin{align}
    \small
    \text{Perf-Recall}@k= \frac{1}{N}\sum^N_{i=1}
    \begin{cases}
        1 & \text{if } {\repo_F}_i \cap \repo^*_i = \repo^*_i , \\
        0 & \text{otherwise.}
    \end{cases}
\end{align}

We prioritize recall in our choice of metrics because retrieving all the necessary code chunks is necessary to solve a task, while the impact of retrieving unnecessary chunks (\ie low precision) is not clear.

\subsection{Implementation}
We implement \cody using CodeSage Small~(S)\citep{zhang2024code} as our pre-trained backbone with the following
modification: we use mean pooling over all chunk tokens instead of using the standard \texttt{[CLS]}
token. This modification resulted in a moderate performance boost in early experiments. We also tie
the weights for both $C$ and $Q$ models, meaning $\theta_c= \theta_q$, which we initialise with the
publicly available CodeSage S weights. We found that weight tying performs consistently better in
our experiments than letting $C$ and $Q$ vary independently during training.
Further model and implementation details are in \cref{sec:app:model-details} and \cref{appendix:hyperparameters}.
\begin{figure}[t]
    \centering
    \resizebox{0.95\linewidth}{!}{\input{swe_results_main_paper.pgf}}
    \caption{Recall@$k$ for the baseline models and our proposed method  on instances of SWE-bench Verified.
        CodeSage-family models substantially outperform other baselines.
        \cody, described in \cref{sec:method},  consistently outperforms baselines across all $k$. The
        dashed line corresponds to the trained \cody model, whereas the solid lines correspond to the
        untrained baselines.}
    \label{fig:recall_plots}
\end{figure}
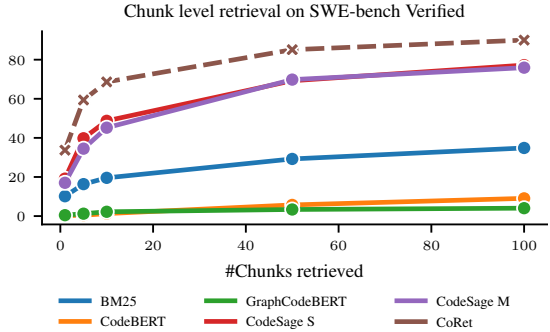

\subsection{Results}

\noindent\textbf{Existing models are sub-optimal when used for retrieval in code editing.} We
compare our model to standard methods such as BM25\citep{Trotman2014bm25}, several text-code encoder
models like CodeBERT\citep{feng-etal-2020-codebert}, GraphCodeBERT\citep{guo2021graphcodebert},
UniXcoder\citep{guo-etal-2022-unixcoder}, and CodeSage\citep{zhang2024code}.
In \cref{fig:recall_plots}, we present each baseline model retrieval performance on SWE-bench
Verified in solid lines. Among the models considered, CodeSage S and CodeSage M perform the best
with the caveat that CodeSage M's performance comes with a much higher computational resource
requirements. For this reason, we pick CodeSage S model as our pretrained backbone for further
finetuning.

\noindent\textbf{Representation learning focused on retrieval improves performance.}
In \cref{fig:recall_plots}, the dashed line reports the performance of \cody{} after training. We
report the perfect chunk recall in \cref{tab:results}.
It is evident that \cody improves upon the best baseline CodeSage S significantly. On SWE-bench
Verified, recall@5 improves by $52.9\%$ compared to CodeSage S for recall@5 and by about $35\%$ for
recall@20.

\begin{table}
    \resizebox{1.\linewidth}{!}{
        \begin{tabular}{l|ccc|ccc}
            \toprule
            \multicolumn{1}{c}{}  & \multicolumn{3}{c}{SWE Verified} & \multicolumn{3}{c}{LCA}                                                                 \\
            \cmidrule{1-7}
            {Model}               & {@5}                             & {@20}                   & {MRR}         & {@5}          & {@20}         & {MRR}         \\
            \midrule
            CodeSage S            & 0.34                             & 0.51                    & 0.35          & 0.26          & 0.34          & 0.28          \\
            \cody{} $-$ CG        & 0.52                             & 0.69                    & 0.52          & \textbf{0.32} & 0.41          & 0.45          \\
            \cody{} $-$ CG + file & \textbf{0.54}                    & 0.69                    & 0.52          & 0.29          & 0.38          & 0.44          \\
            \midrule
            \cody                 & \textbf{0.54}                    & \textbf{0.71}           & \textbf{0.53} & \textbf{0.32} & \textbf{0.47} & \textbf{0.47} \\
            \bottomrule
        \end{tabular}
    }
    \caption{Perfect recall at chunk level on code-editing retrieval tasks.}
    \label{tab:results}
\end{table}

\noindent\textbf{Call graph context improves multi-chunk retrieval.}
Next, we ablate on the call graph context to assess the contribution of including this additional
information during training and inference.
\cref{tab:results} reports perfect chunk recall on our two evaluation datasets. \cody{} - CG
indicates our method without the call graph context described in \cref{subsec:cg_context}. We
present results of additional baselines in \cref{app:fig:untrained} in the Appendix and provide a
summary here. We further perform an ablation (\cody{} - CG + file) where we include as additional
context a number of chunks randomly sampled from the same file as the target chunk, and we leave the
rest of the pipeline unchanged. We observe a substantial decrease in performances throughout,
further validating our design choice.

\noindent\textbf{Choice of negative samples in \texorpdfstring{\cref{eqn:final_loss}}{Equation 2}
    influences the performance.} Traditional methods for representation learning using contrastive
losses\citep{chopra_contrastive,chen2020simple,karpukhin-etal-2020-dense} form negative samples by
using positives from elements across the batch. We term this \emph{across-instance negatives}. In
\cref{eqn:final_loss}, we form the negatives from entirely within a problem's repository and not
across problems, like in \citet{sohn_classmining}, and we term this \emph{in-instance negatives}.
For our experiments we randomly draw up to a maximum of 1024 negative samples to compute the term
$\tilde{\Gamma}$ in our loss (\cref{eqn:final_loss}) and consider the influence of the number of
negatives in \cref{fig:negatives} in the Appendix. We find improvements for all $k$ when we include
in-instance negatives. Additionally, the impact of number of negatives is also evident; the larger
the number of negatives, the better the performance. Increasing the number of negatives from $8 \to
    1024$ improves the recall@20 by almost $10$ points. This provides further evidence that
\cref{eqn:loss_def} with the approximation from \cref{eqn:final_loss} mimics the goal of a
retriever: it explicitly models which code chunks are relevant and which are not.

\begin{table}
    \centering
    \resizebox{\linewidth}{!}{
        \begin{tabular}{l|ccc|ccc}
            \toprule
            \multicolumn{1}{c}{} & \multicolumn{3}{c}{{wo/filepath}} &
            \multicolumn{3}{c}{{w/filepath}}
            \\
            \cmidrule{1-7}

            {Model}              & {@5}                              & {@20}
                                 & {MRR}                             & {@5}          & {@20}         & {MRR}         \\
            \toprule
            BM25                 & 0.15                              & 0.21
                                 & 0.14                              & {0.16}        & {0.22}        & {0.16}        \\
            CodeSage S           & 0.40                              & \textbf{0.58}
                                 & 0.33                              & 0.40          & 0.57          & { 0.35}       \\
            \midrule
            \cody                & \textbf{0.42}                     & \textbf{0.58}
                                 & \textbf{0.42}                     & \textbf{0.53} & \textbf{0.70} & \textbf{0.53} \\
            \bottomrule
        \end{tabular}}
    \caption{Chunk-level accuracy on SWE-bench Verified, comparing CoRet without (wo/filepath) and with
        (w/filepath) file path input. \cody is fine-tuned with file paths, highlighting performance drop
        without them. \cody performance improves with file path context.}
    \label{tab:filepath}
\end{table}

\noindent\textbf{File hierarchy is an important feature for retrieval.} We prefix the file path to
each chunk as a part of the chunking strategy. We find that it is an important feature. In
\cref{tab:filepath}, we show the retrieval performance drops for \cody when file paths are removed
from the chunk representation for inference, as the models are originally trained with the file path
present. The performance difference is minimal for both BM25 and CodeSage S whereas \cody learns to
rely on the file name through training. Additional evidence from an attention matrix is presented in
\cref{appendix:filepath_exp}.

%% file: swe_results_main_paper.pgf
\begingroup%
\makeatletter%
\begin{pgfpicture}%
\pgfpathrectangle{\pgfpointorigin}{\pgfqpoint{3.266660in}{2.067667in}}%
\pgfusepath{use as bounding box, clip}%
\begin{pgfscope}%
\pgfsetbuttcap%
\pgfsetmiterjoin%
\definecolor{currentfill}{rgb}{1.000000,1.000000,1.000000}%
\pgfsetfillcolor{currentfill}%
\pgfsetlinewidth{0.000000pt}%
\definecolor{currentstroke}{rgb}{1.000000,1.000000,1.000000}%
\pgfsetstrokecolor{currentstroke}%
\pgfsetdash{}{0pt}%
\pgfpathmoveto{\pgfqpoint{0.000000in}{0.000000in}}%
\pgfpathlineto{\pgfqpoint{3.266660in}{0.000000in}}%
\pgfpathlineto{\pgfqpoint{3.266660in}{2.067667in}}%
\pgfpathlineto{\pgfqpoint{0.000000in}{2.067667in}}%
\pgfpathlineto{\pgfqpoint{0.000000in}{0.000000in}}%
\pgfpathclose%
\pgfusepath{fill}%
\end{pgfscope}%
\begin{pgfscope}%
\pgfsetbuttcap%
\pgfsetmiterjoin%
\definecolor{currentfill}{rgb}{1.000000,1.000000,1.000000}%
\pgfsetfillcolor{currentfill}%
\pgfsetlinewidth{0.000000pt}%
\definecolor{currentstroke}{rgb}{0.000000,0.000000,0.000000}%
\pgfsetstrokecolor{currentstroke}%
\pgfsetstrokeopacity{0.000000}%
\pgfsetdash{}{0pt}%
\pgfpathmoveto{\pgfqpoint{0.249073in}{0.706639in}}%
\pgfpathlineto{\pgfqpoint{3.216660in}{0.706639in}}%
\pgfpathlineto{\pgfqpoint{3.216660in}{1.849915in}}%
\pgfpathlineto{\pgfqpoint{0.249073in}{1.849915in}}%
\pgfpathlineto{\pgfqpoint{0.249073in}{0.706639in}}%
\pgfpathclose%
\pgfusepath{fill}%
\end{pgfscope}%
\begin{pgfscope}%
\pgfsetbuttcap%
\pgfsetroundjoin%
\definecolor{currentfill}{rgb}{0.000000,0.000000,0.000000}%
\pgfsetfillcolor{currentfill}%
\pgfsetlinewidth{0.803000pt}%
\definecolor{currentstroke}{rgb}{0.000000,0.000000,0.000000}%
\pgfsetstrokecolor{currentstroke}%
\pgfsetdash{}{0pt}%
\pgfsys@defobject{currentmarker}{\pgfqpoint{0.000000in}{-0.048611in}}{\pgfqpoint{0.000000in}{0.000000in}}{%
\pgfpathmoveto{\pgfqpoint{0.000000in}{0.000000in}}%
\pgfpathlineto{\pgfqpoint{0.000000in}{-0.048611in}}%
\pgfusepath{stroke,fill}%
}%
\begin{pgfscope}%
\pgfsys@transformshift{0.356712in}{0.706639in}%
\pgfsys@useobject{currentmarker}{}%
\end{pgfscope}%
\end{pgfscope}%
\begin{pgfscope}%
\definecolor{textcolor}{rgb}{0.000000,0.000000,0.000000}%
\pgfsetstrokecolor{textcolor}%
\pgfsetfillcolor{textcolor}%
\pgftext[x=0.356712in,y=0.609416in,,top]{\color{textcolor}{\rmfamily\fontsize{6.000000}{7.200000}\selectfont\catcode`\^=\active\def^{\ifmmode\sp\else\^{}\fi}\catcode`\%=\active\def
\end{pgfscope}%
\begin{pgfscope}%
\pgfsetbuttcap%
\pgfsetroundjoin%
\definecolor{currentfill}{rgb}{0.000000,0.000000,0.000000}%
\pgfsetfillcolor{currentfill}%
\pgfsetlinewidth{0.803000pt}%
\definecolor{currentstroke}{rgb}{0.000000,0.000000,0.000000}%
\pgfsetstrokecolor{currentstroke}%
\pgfsetdash{}{0pt}%
\pgfsys@defobject{currentmarker}{\pgfqpoint{0.000000in}{-0.048611in}}{\pgfqpoint{0.000000in}{0.000000in}}{%
\pgfpathmoveto{\pgfqpoint{0.000000in}{0.000000in}}%
\pgfpathlineto{\pgfqpoint{0.000000in}{-0.048611in}}%
\pgfusepath{stroke,fill}%
}%
\begin{pgfscope}%
\pgfsys@transformshift{0.901724in}{0.706639in}%
\pgfsys@useobject{currentmarker}{}%
\end{pgfscope}%
\end{pgfscope}%
\begin{pgfscope}%
\definecolor{textcolor}{rgb}{0.000000,0.000000,0.000000}%
\pgfsetstrokecolor{textcolor}%
\pgfsetfillcolor{textcolor}%
\pgftext[x=0.901724in,y=0.609416in,,top]{\color{textcolor}{\rmfamily\fontsize{6.000000}{7.200000}\selectfont\catcode`\^=\active\def^{\ifmmode\sp\else\^{}\fi}\catcode`\%=\active\def
\end{pgfscope}%
\begin{pgfscope}%
\pgfsetbuttcap%
\pgfsetroundjoin%
\definecolor{currentfill}{rgb}{0.000000,0.000000,0.000000}%
\pgfsetfillcolor{currentfill}%
\pgfsetlinewidth{0.803000pt}%
\definecolor{currentstroke}{rgb}{0.000000,0.000000,0.000000}%
\pgfsetstrokecolor{currentstroke}%
\pgfsetdash{}{0pt}%
\pgfsys@defobject{currentmarker}{\pgfqpoint{0.000000in}{-0.048611in}}{\pgfqpoint{0.000000in}{0.000000in}}{%
\pgfpathmoveto{\pgfqpoint{0.000000in}{0.000000in}}%
\pgfpathlineto{\pgfqpoint{0.000000in}{-0.048611in}}%
\pgfusepath{stroke,fill}%
}%
\begin{pgfscope}%
\pgfsys@transformshift{1.446735in}{0.706639in}%
\pgfsys@useobject{currentmarker}{}%
\end{pgfscope}%
\end{pgfscope}%
\begin{pgfscope}%
\definecolor{textcolor}{rgb}{0.000000,0.000000,0.000000}%
\pgfsetstrokecolor{textcolor}%
\pgfsetfillcolor{textcolor}%
\pgftext[x=1.446735in,y=0.609416in,,top]{\color{textcolor}{\rmfamily\fontsize{6.000000}{7.200000}\selectfont\catcode`\^=\active\def^{\ifmmode\sp\else\^{}\fi}\catcode`\%=\active\def
\end{pgfscope}%
\begin{pgfscope}%
\pgfsetbuttcap%
\pgfsetroundjoin%
\definecolor{currentfill}{rgb}{0.000000,0.000000,0.000000}%
\pgfsetfillcolor{currentfill}%
\pgfsetlinewidth{0.803000pt}%
\definecolor{currentstroke}{rgb}{0.000000,0.000000,0.000000}%
\pgfsetstrokecolor{currentstroke}%
\pgfsetdash{}{0pt}%
\pgfsys@defobject{currentmarker}{\pgfqpoint{0.000000in}{-0.048611in}}{\pgfqpoint{0.000000in}{0.000000in}}{%
\pgfpathmoveto{\pgfqpoint{0.000000in}{0.000000in}}%
\pgfpathlineto{\pgfqpoint{0.000000in}{-0.048611in}}%
\pgfusepath{stroke,fill}%
}%
\begin{pgfscope}%
\pgfsys@transformshift{1.991747in}{0.706639in}%
\pgfsys@useobject{currentmarker}{}%
\end{pgfscope}%
\end{pgfscope}%
\begin{pgfscope}%
\definecolor{textcolor}{rgb}{0.000000,0.000000,0.000000}%
\pgfsetstrokecolor{textcolor}%
\pgfsetfillcolor{textcolor}%
\pgftext[x=1.991747in,y=0.609416in,,top]{\color{textcolor}{\rmfamily\fontsize{6.000000}{7.200000}\selectfont\catcode`\^=\active\def^{\ifmmode\sp\else\^{}\fi}\catcode`\%=\active\def
\end{pgfscope}%
\begin{pgfscope}%
\pgfsetbuttcap%
\pgfsetroundjoin%
\definecolor{currentfill}{rgb}{0.000000,0.000000,0.000000}%
\pgfsetfillcolor{currentfill}%
\pgfsetlinewidth{0.803000pt}%
\definecolor{currentstroke}{rgb}{0.000000,0.000000,0.000000}%
\pgfsetstrokecolor{currentstroke}%
\pgfsetdash{}{0pt}%
\pgfsys@defobject{currentmarker}{\pgfqpoint{0.000000in}{-0.048611in}}{\pgfqpoint{0.000000in}{0.000000in}}{%
\pgfpathmoveto{\pgfqpoint{0.000000in}{0.000000in}}%
\pgfpathlineto{\pgfqpoint{0.000000in}{-0.048611in}}%
\pgfusepath{stroke,fill}%
}%
\begin{pgfscope}%
\pgfsys@transformshift{2.536758in}{0.706639in}%
\pgfsys@useobject{currentmarker}{}%
\end{pgfscope}%
\end{pgfscope}%
\begin{pgfscope}%
\definecolor{textcolor}{rgb}{0.000000,0.000000,0.000000}%
\pgfsetstrokecolor{textcolor}%
\pgfsetfillcolor{textcolor}%
\pgftext[x=2.536758in,y=0.609416in,,top]{\color{textcolor}{\rmfamily\fontsize{6.000000}{7.200000}\selectfont\catcode`\^=\active\def^{\ifmmode\sp\else\^{}\fi}\catcode`\%=\active\def
\end{pgfscope}%
\begin{pgfscope}%
\pgfsetbuttcap%
\pgfsetroundjoin%
\definecolor{currentfill}{rgb}{0.000000,0.000000,0.000000}%
\pgfsetfillcolor{currentfill}%
\pgfsetlinewidth{0.803000pt}%
\definecolor{currentstroke}{rgb}{0.000000,0.000000,0.000000}%
\pgfsetstrokecolor{currentstroke}%
\pgfsetdash{}{0pt}%
\pgfsys@defobject{currentmarker}{\pgfqpoint{0.000000in}{-0.048611in}}{\pgfqpoint{0.000000in}{0.000000in}}{%
\pgfpathmoveto{\pgfqpoint{0.000000in}{0.000000in}}%
\pgfpathlineto{\pgfqpoint{0.000000in}{-0.048611in}}%
\pgfusepath{stroke,fill}%
}%
\begin{pgfscope}%
\pgfsys@transformshift{3.081770in}{0.706639in}%
\pgfsys@useobject{currentmarker}{}%
\end{pgfscope}%
\end{pgfscope}%
\begin{pgfscope}%
\definecolor{textcolor}{rgb}{0.000000,0.000000,0.000000}%
\pgfsetstrokecolor{textcolor}%
\pgfsetfillcolor{textcolor}%
\pgftext[x=3.081770in,y=0.609416in,,top]{\color{textcolor}{\rmfamily\fontsize{6.000000}{7.200000}\selectfont\catcode`\^=\active\def^{\ifmmode\sp\else\^{}\fi}\catcode`\%=\active\def
\end{pgfscope}%
\begin{pgfscope}%
\definecolor{textcolor}{rgb}{0.000000,0.000000,0.000000}%
\pgfsetstrokecolor{textcolor}%
\pgfsetfillcolor{textcolor}%
\pgftext[x=1.732866in,y=0.473213in,,top]{\color{textcolor}{\rmfamily\fontsize{8.000000}{9.600000}\selectfont\catcode`\^=\active\def^{\ifmmode\sp\else\^{}\fi}\catcode`\%=\active\def
\end{pgfscope}%
\begin{pgfscope}%
\pgfsetbuttcap%
\pgfsetroundjoin%
\definecolor{currentfill}{rgb}{0.000000,0.000000,0.000000}%
\pgfsetfillcolor{currentfill}%
\pgfsetlinewidth{0.803000pt}%
\definecolor{currentstroke}{rgb}{0.000000,0.000000,0.000000}%
\pgfsetstrokecolor{currentstroke}%
\pgfsetdash{}{0pt}%
\pgfsys@defobject{currentmarker}{\pgfqpoint{-0.048611in}{0.000000in}}{\pgfqpoint{-0.000000in}{0.000000in}}{%
\pgfpathmoveto{\pgfqpoint{-0.000000in}{0.000000in}}%
\pgfpathlineto{\pgfqpoint{-0.048611in}{0.000000in}}%
\pgfusepath{stroke,fill}%
}%
\begin{pgfscope}%
\pgfsys@transformshift{0.249073in}{0.755945in}%
\pgfsys@useobject{currentmarker}{}%
\end{pgfscope}%
\end{pgfscope}%
\begin{pgfscope}%
\definecolor{textcolor}{rgb}{0.000000,0.000000,0.000000}%
\pgfsetstrokecolor{textcolor}%
\pgfsetfillcolor{textcolor}%
\pgftext[x=0.100925in, y=0.724288in, left, base]{\color{textcolor}{\rmfamily\fontsize{6.000000}{7.200000}\selectfont\catcode`\^=\active\def^{\ifmmode\sp\else\^{}\fi}\catcode`\%=\active\def
\end{pgfscope}%
\begin{pgfscope}%
\pgfsetbuttcap%
\pgfsetroundjoin%
\definecolor{currentfill}{rgb}{0.000000,0.000000,0.000000}%
\pgfsetfillcolor{currentfill}%
\pgfsetlinewidth{0.803000pt}%
\definecolor{currentstroke}{rgb}{0.000000,0.000000,0.000000}%
\pgfsetstrokecolor{currentstroke}%
\pgfsetdash{}{0pt}%
\pgfsys@defobject{currentmarker}{\pgfqpoint{-0.048611in}{0.000000in}}{\pgfqpoint{-0.000000in}{0.000000in}}{%
\pgfpathmoveto{\pgfqpoint{-0.000000in}{0.000000in}}%
\pgfpathlineto{\pgfqpoint{-0.048611in}{0.000000in}}%
\pgfusepath{stroke,fill}%
}%
\begin{pgfscope}%
\pgfsys@transformshift{0.249073in}{0.987460in}%
\pgfsys@useobject{currentmarker}{}%
\end{pgfscope}%
\end{pgfscope}%
\begin{pgfscope}%
\definecolor{textcolor}{rgb}{0.000000,0.000000,0.000000}%
\pgfsetstrokecolor{textcolor}%
\pgfsetfillcolor{textcolor}%
\pgftext[x=0.050000in, y=0.955803in, left, base]{\color{textcolor}{\rmfamily\fontsize{6.000000}{7.200000}\selectfont\catcode`\^=\active\def^{\ifmmode\sp\else\^{}\fi}\catcode`\%=\active\def
\end{pgfscope}%
\begin{pgfscope}%
\pgfsetbuttcap%
\pgfsetroundjoin%
\definecolor{currentfill}{rgb}{0.000000,0.000000,0.000000}%
\pgfsetfillcolor{currentfill}%
\pgfsetlinewidth{0.803000pt}%
\definecolor{currentstroke}{rgb}{0.000000,0.000000,0.000000}%
\pgfsetstrokecolor{currentstroke}%
\pgfsetdash{}{0pt}%
\pgfsys@defobject{currentmarker}{\pgfqpoint{-0.048611in}{0.000000in}}{\pgfqpoint{-0.000000in}{0.000000in}}{%
\pgfpathmoveto{\pgfqpoint{-0.000000in}{0.000000in}}%
\pgfpathlineto{\pgfqpoint{-0.048611in}{0.000000in}}%
\pgfusepath{stroke,fill}%
}%
\begin{pgfscope}%
\pgfsys@transformshift{0.249073in}{1.218975in}%
\pgfsys@useobject{currentmarker}{}%
\end{pgfscope}%
\end{pgfscope}%
\begin{pgfscope}%
\definecolor{textcolor}{rgb}{0.000000,0.000000,0.000000}%
\pgfsetstrokecolor{textcolor}%
\pgfsetfillcolor{textcolor}%
\pgftext[x=0.050000in, y=1.187318in, left, base]{\color{textcolor}{\rmfamily\fontsize{6.000000}{7.200000}\selectfont\catcode`\^=\active\def^{\ifmmode\sp\else\^{}\fi}\catcode`\%=\active\def
\end{pgfscope}%
\begin{pgfscope}%
\pgfsetbuttcap%
\pgfsetroundjoin%
\definecolor{currentfill}{rgb}{0.000000,0.000000,0.000000}%
\pgfsetfillcolor{currentfill}%
\pgfsetlinewidth{0.803000pt}%
\definecolor{currentstroke}{rgb}{0.000000,0.000000,0.000000}%
\pgfsetstrokecolor{currentstroke}%
\pgfsetdash{}{0pt}%
\pgfsys@defobject{currentmarker}{\pgfqpoint{-0.048611in}{0.000000in}}{\pgfqpoint{-0.000000in}{0.000000in}}{%
\pgfpathmoveto{\pgfqpoint{-0.000000in}{0.000000in}}%
\pgfpathlineto{\pgfqpoint{-0.048611in}{0.000000in}}%
\pgfusepath{stroke,fill}%
}%
\begin{pgfscope}%
\pgfsys@transformshift{0.249073in}{1.450490in}%
\pgfsys@useobject{currentmarker}{}%
\end{pgfscope}%
\end{pgfscope}%
\begin{pgfscope}%
\definecolor{textcolor}{rgb}{0.000000,0.000000,0.000000}%
\pgfsetstrokecolor{textcolor}%
\pgfsetfillcolor{textcolor}%
\pgftext[x=0.050000in, y=1.418833in, left, base]{\color{textcolor}{\rmfamily\fontsize{6.000000}{7.200000}\selectfont\catcode`\^=\active\def^{\ifmmode\sp\else\^{}\fi}\catcode`\%=\active\def
\end{pgfscope}%
\begin{pgfscope}%
\pgfsetbuttcap%
\pgfsetroundjoin%
\definecolor{currentfill}{rgb}{0.000000,0.000000,0.000000}%
\pgfsetfillcolor{currentfill}%
\pgfsetlinewidth{0.803000pt}%
\definecolor{currentstroke}{rgb}{0.000000,0.000000,0.000000}%
\pgfsetstrokecolor{currentstroke}%
\pgfsetdash{}{0pt}%
\pgfsys@defobject{currentmarker}{\pgfqpoint{-0.048611in}{0.000000in}}{\pgfqpoint{-0.000000in}{0.000000in}}{%
\pgfpathmoveto{\pgfqpoint{-0.000000in}{0.000000in}}%
\pgfpathlineto{\pgfqpoint{-0.048611in}{0.000000in}}%
\pgfusepath{stroke,fill}%
}%
\begin{pgfscope}%
\pgfsys@transformshift{0.249073in}{1.682005in}%
\pgfsys@useobject{currentmarker}{}%
\end{pgfscope}%
\end{pgfscope}%
\begin{pgfscope}%
\definecolor{textcolor}{rgb}{0.000000,0.000000,0.000000}%
\pgfsetstrokecolor{textcolor}%
\pgfsetfillcolor{textcolor}%
\pgftext[x=0.050000in, y=1.650348in, left, base]{\color{textcolor}{\rmfamily\fontsize{6.000000}{7.200000}\selectfont\catcode`\^=\active\def^{\ifmmode\sp\else\^{}\fi}\catcode`\%=\active\def
\end{pgfscope}%
\begin{pgfscope}%
\pgfpathrectangle{\pgfqpoint{0.249073in}{0.706639in}}{\pgfqpoint{2.967588in}{1.143277in}}%
\pgfusepath{clip}%
\pgfsetrectcap%
\pgfsetroundjoin%
\pgfsetlinewidth{2.007500pt}%
\definecolor{currentstroke}{rgb}{0.121569,0.466667,0.705882}%
\pgfsetstrokecolor{currentstroke}%
\pgfsetdash{}{0pt}%
\pgfpathmoveto{\pgfqpoint{0.383963in}{0.873476in}}%
\pgfpathlineto{\pgfqpoint{0.492965in}{0.945170in}}%
\pgfpathlineto{\pgfqpoint{0.629218in}{0.982397in}}%
\pgfpathlineto{\pgfqpoint{1.719241in}{1.094310in}}%
\pgfpathlineto{\pgfqpoint{3.081770in}{1.159257in}}%
\pgfusepath{stroke}%
\end{pgfscope}%
\begin{pgfscope}%
\pgfpathrectangle{\pgfqpoint{0.249073in}{0.706639in}}{\pgfqpoint{2.967588in}{1.143277in}}%
\pgfusepath{clip}%
\pgfsetbuttcap%
\pgfsetroundjoin%
\definecolor{currentfill}{rgb}{0.121569,0.466667,0.705882}%
\pgfsetfillcolor{currentfill}%
\pgfsetlinewidth{0.752812pt}%
\definecolor{currentstroke}{rgb}{1.000000,1.000000,1.000000}%
\pgfsetstrokecolor{currentstroke}%
\pgfsetdash{}{0pt}%
\pgfsys@defobject{currentmarker}{\pgfqpoint{-0.041667in}{-0.041667in}}{\pgfqpoint{0.041667in}{0.041667in}}{%
\pgfpathmoveto{\pgfqpoint{0.000000in}{-0.041667in}}%
\pgfpathcurveto{\pgfqpoint{0.011050in}{-0.041667in}}{\pgfqpoint{0.021649in}{-0.037276in}}{\pgfqpoint{0.029463in}{-0.029463in}}%
\pgfpathcurveto{\pgfqpoint{0.037276in}{-0.021649in}}{\pgfqpoint{0.041667in}{-0.011050in}}{\pgfqpoint{0.041667in}{0.000000in}}%
\pgfpathcurveto{\pgfqpoint{0.041667in}{0.011050in}}{\pgfqpoint{0.037276in}{0.021649in}}{\pgfqpoint{0.029463in}{0.029463in}}%
\pgfpathcurveto{\pgfqpoint{0.021649in}{0.037276in}}{\pgfqpoint{0.011050in}{0.041667in}}{\pgfqpoint{0.000000in}{0.041667in}}%
\pgfpathcurveto{\pgfqpoint{-0.011050in}{0.041667in}}{\pgfqpoint{-0.021649in}{0.037276in}}{\pgfqpoint{-0.029463in}{0.029463in}}%
\pgfpathcurveto{\pgfqpoint{-0.037276in}{0.021649in}}{\pgfqpoint{-0.041667in}{0.011050in}}{\pgfqpoint{-0.041667in}{0.000000in}}%
\pgfpathcurveto{\pgfqpoint{-0.041667in}{-0.011050in}}{\pgfqpoint{-0.037276in}{-0.021649in}}{\pgfqpoint{-0.029463in}{-0.029463in}}%
\pgfpathcurveto{\pgfqpoint{-0.021649in}{-0.037276in}}{\pgfqpoint{-0.011050in}{-0.041667in}}{\pgfqpoint{0.000000in}{-0.041667in}}%
\pgfpathlineto{\pgfqpoint{0.000000in}{-0.041667in}}%
\pgfpathclose%
\pgfusepath{stroke,fill}%
}%
\begin{pgfscope}%
\pgfsys@transformshift{0.383963in}{0.873476in}%
\pgfsys@useobject{currentmarker}{}%
\end{pgfscope}%
\begin{pgfscope}%
\pgfsys@transformshift{0.492965in}{0.945170in}%
\pgfsys@useobject{currentmarker}{}%
\end{pgfscope}%
\begin{pgfscope}%
\pgfsys@transformshift{0.629218in}{0.982397in}%
\pgfsys@useobject{currentmarker}{}%
\end{pgfscope}%
\begin{pgfscope}%
\pgfsys@transformshift{1.719241in}{1.094310in}%
\pgfsys@useobject{currentmarker}{}%
\end{pgfscope}%
\begin{pgfscope}%
\pgfsys@transformshift{3.081770in}{1.159257in}%
\pgfsys@useobject{currentmarker}{}%
\end{pgfscope}%
\end{pgfscope}%
\begin{pgfscope}%
\pgfpathrectangle{\pgfqpoint{0.249073in}{0.706639in}}{\pgfqpoint{2.967588in}{1.143277in}}%
\pgfusepath{clip}%
\pgfsetrectcap%
\pgfsetroundjoin%
\pgfsetlinewidth{2.007500pt}%
\definecolor{currentstroke}{rgb}{1.000000,0.498039,0.054902}%
\pgfsetstrokecolor{currentstroke}%
\pgfsetdash{}{0pt}%
\pgfpathmoveto{\pgfqpoint{0.383963in}{0.758606in}}%
\pgfpathlineto{\pgfqpoint{0.492965in}{0.760823in}}%
\pgfpathlineto{\pgfqpoint{0.629218in}{0.770692in}}%
\pgfpathlineto{\pgfqpoint{1.719241in}{0.822012in}}%
\pgfpathlineto{\pgfqpoint{3.081770in}{0.860906in}}%
\pgfusepath{stroke}%
\end{pgfscope}%
\begin{pgfscope}%
\pgfpathrectangle{\pgfqpoint{0.249073in}{0.706639in}}{\pgfqpoint{2.967588in}{1.143277in}}%
\pgfusepath{clip}%
\pgfsetbuttcap%
\pgfsetroundjoin%
\definecolor{currentfill}{rgb}{1.000000,0.498039,0.054902}%
\pgfsetfillcolor{currentfill}%
\pgfsetlinewidth{0.752812pt}%
\definecolor{currentstroke}{rgb}{1.000000,1.000000,1.000000}%
\pgfsetstrokecolor{currentstroke}%
\pgfsetdash{}{0pt}%
\pgfsys@defobject{currentmarker}{\pgfqpoint{-0.041667in}{-0.041667in}}{\pgfqpoint{0.041667in}{0.041667in}}{%
\pgfpathmoveto{\pgfqpoint{0.000000in}{-0.041667in}}%
\pgfpathcurveto{\pgfqpoint{0.011050in}{-0.041667in}}{\pgfqpoint{0.021649in}{-0.037276in}}{\pgfqpoint{0.029463in}{-0.029463in}}%
\pgfpathcurveto{\pgfqpoint{0.037276in}{-0.021649in}}{\pgfqpoint{0.041667in}{-0.011050in}}{\pgfqpoint{0.041667in}{0.000000in}}%
\pgfpathcurveto{\pgfqpoint{0.041667in}{0.011050in}}{\pgfqpoint{0.037276in}{0.021649in}}{\pgfqpoint{0.029463in}{0.029463in}}%
\pgfpathcurveto{\pgfqpoint{0.021649in}{0.037276in}}{\pgfqpoint{0.011050in}{0.041667in}}{\pgfqpoint{0.000000in}{0.041667in}}%
\pgfpathcurveto{\pgfqpoint{-0.011050in}{0.041667in}}{\pgfqpoint{-0.021649in}{0.037276in}}{\pgfqpoint{-0.029463in}{0.029463in}}%
\pgfpathcurveto{\pgfqpoint{-0.037276in}{0.021649in}}{\pgfqpoint{-0.041667in}{0.011050in}}{\pgfqpoint{-0.041667in}{0.000000in}}%
\pgfpathcurveto{\pgfqpoint{-0.041667in}{-0.011050in}}{\pgfqpoint{-0.037276in}{-0.021649in}}{\pgfqpoint{-0.029463in}{-0.029463in}}%
\pgfpathcurveto{\pgfqpoint{-0.021649in}{-0.037276in}}{\pgfqpoint{-0.011050in}{-0.041667in}}{\pgfqpoint{0.000000in}{-0.041667in}}%
\pgfpathlineto{\pgfqpoint{0.000000in}{-0.041667in}}%
\pgfpathclose%
\pgfusepath{stroke,fill}%
}%
\begin{pgfscope}%
\pgfsys@transformshift{0.383963in}{0.758606in}%
\pgfsys@useobject{currentmarker}{}%
\end{pgfscope}%
\begin{pgfscope}%
\pgfsys@transformshift{0.492965in}{0.760823in}%
\pgfsys@useobject{currentmarker}{}%
\end{pgfscope}%
\begin{pgfscope}%
\pgfsys@transformshift{0.629218in}{0.770692in}%
\pgfsys@useobject{currentmarker}{}%
\end{pgfscope}%
\begin{pgfscope}%
\pgfsys@transformshift{1.719241in}{0.822012in}%
\pgfsys@useobject{currentmarker}{}%
\end{pgfscope}%
\begin{pgfscope}%
\pgfsys@transformshift{3.081770in}{0.860906in}%
\pgfsys@useobject{currentmarker}{}%
\end{pgfscope}%
\end{pgfscope}%
\begin{pgfscope}%
\pgfpathrectangle{\pgfqpoint{0.249073in}{0.706639in}}{\pgfqpoint{2.967588in}{1.143277in}}%
\pgfusepath{clip}%
\pgfsetrectcap%
\pgfsetroundjoin%
\pgfsetlinewidth{2.007500pt}%
\definecolor{currentstroke}{rgb}{0.172549,0.627451,0.172549}%
\pgfsetstrokecolor{currentstroke}%
\pgfsetdash{}{0pt}%
\pgfpathmoveto{\pgfqpoint{0.383963in}{0.761267in}}%
\pgfpathlineto{\pgfqpoint{0.492965in}{0.770137in}}%
\pgfpathlineto{\pgfqpoint{0.629218in}{0.781447in}}%
\pgfpathlineto{\pgfqpoint{1.719241in}{0.794994in}}%
\pgfpathlineto{\pgfqpoint{3.081770in}{0.802756in}}%
\pgfusepath{stroke}%
\end{pgfscope}%
\begin{pgfscope}%
\pgfpathrectangle{\pgfqpoint{0.249073in}{0.706639in}}{\pgfqpoint{2.967588in}{1.143277in}}%
\pgfusepath{clip}%
\pgfsetbuttcap%
\pgfsetroundjoin%
\definecolor{currentfill}{rgb}{0.172549,0.627451,0.172549}%
\pgfsetfillcolor{currentfill}%
\pgfsetlinewidth{0.752812pt}%
\definecolor{currentstroke}{rgb}{1.000000,1.000000,1.000000}%
\pgfsetstrokecolor{currentstroke}%
\pgfsetdash{}{0pt}%
\pgfsys@defobject{currentmarker}{\pgfqpoint{-0.041667in}{-0.041667in}}{\pgfqpoint{0.041667in}{0.041667in}}{%
\pgfpathmoveto{\pgfqpoint{0.000000in}{-0.041667in}}%
\pgfpathcurveto{\pgfqpoint{0.011050in}{-0.041667in}}{\pgfqpoint{0.021649in}{-0.037276in}}{\pgfqpoint{0.029463in}{-0.029463in}}%
\pgfpathcurveto{\pgfqpoint{0.037276in}{-0.021649in}}{\pgfqpoint{0.041667in}{-0.011050in}}{\pgfqpoint{0.041667in}{0.000000in}}%
\pgfpathcurveto{\pgfqpoint{0.041667in}{0.011050in}}{\pgfqpoint{0.037276in}{0.021649in}}{\pgfqpoint{0.029463in}{0.029463in}}%
\pgfpathcurveto{\pgfqpoint{0.021649in}{0.037276in}}{\pgfqpoint{0.011050in}{0.041667in}}{\pgfqpoint{0.000000in}{0.041667in}}%
\pgfpathcurveto{\pgfqpoint{-0.011050in}{0.041667in}}{\pgfqpoint{-0.021649in}{0.037276in}}{\pgfqpoint{-0.029463in}{0.029463in}}%
\pgfpathcurveto{\pgfqpoint{-0.037276in}{0.021649in}}{\pgfqpoint{-0.041667in}{0.011050in}}{\pgfqpoint{-0.041667in}{0.000000in}}%
\pgfpathcurveto{\pgfqpoint{-0.041667in}{-0.011050in}}{\pgfqpoint{-0.037276in}{-0.021649in}}{\pgfqpoint{-0.029463in}{-0.029463in}}%
\pgfpathcurveto{\pgfqpoint{-0.021649in}{-0.037276in}}{\pgfqpoint{-0.011050in}{-0.041667in}}{\pgfqpoint{0.000000in}{-0.041667in}}%
\pgfpathlineto{\pgfqpoint{0.000000in}{-0.041667in}}%
\pgfpathclose%
\pgfusepath{stroke,fill}%
}%
\begin{pgfscope}%
\pgfsys@transformshift{0.383963in}{0.761267in}%
\pgfsys@useobject{currentmarker}{}%
\end{pgfscope}%
\begin{pgfscope}%
\pgfsys@transformshift{0.492965in}{0.770137in}%
\pgfsys@useobject{currentmarker}{}%
\end{pgfscope}%
\begin{pgfscope}%
\pgfsys@transformshift{0.629218in}{0.781447in}%
\pgfsys@useobject{currentmarker}{}%
\end{pgfscope}%
\begin{pgfscope}%
\pgfsys@transformshift{1.719241in}{0.794994in}%
\pgfsys@useobject{currentmarker}{}%
\end{pgfscope}%
\begin{pgfscope}%
\pgfsys@transformshift{3.081770in}{0.802756in}%
\pgfsys@useobject{currentmarker}{}%
\end{pgfscope}%
\end{pgfscope}%
\begin{pgfscope}%
\pgfpathrectangle{\pgfqpoint{0.249073in}{0.706639in}}{\pgfqpoint{2.967588in}{1.143277in}}%
\pgfusepath{clip}%
\pgfsetrectcap%
\pgfsetroundjoin%
\pgfsetlinewidth{2.007500pt}%
\definecolor{currentstroke}{rgb}{0.839216,0.152941,0.156863}%
\pgfsetstrokecolor{currentstroke}%
\pgfsetdash{}{0pt}%
\pgfpathmoveto{\pgfqpoint{0.383963in}{0.977641in}}%
\pgfpathlineto{\pgfqpoint{0.492965in}{1.216754in}}%
\pgfpathlineto{\pgfqpoint{0.629218in}{1.319951in}}%
\pgfpathlineto{\pgfqpoint{1.719241in}{1.557321in}}%
\pgfpathlineto{\pgfqpoint{3.081770in}{1.649955in}}%
\pgfusepath{stroke}%
\end{pgfscope}%
\begin{pgfscope}%
\pgfpathrectangle{\pgfqpoint{0.249073in}{0.706639in}}{\pgfqpoint{2.967588in}{1.143277in}}%
\pgfusepath{clip}%
\pgfsetbuttcap%
\pgfsetroundjoin%
\definecolor{currentfill}{rgb}{0.839216,0.152941,0.156863}%
\pgfsetfillcolor{currentfill}%
\pgfsetlinewidth{0.752812pt}%
\definecolor{currentstroke}{rgb}{1.000000,1.000000,1.000000}%
\pgfsetstrokecolor{currentstroke}%
\pgfsetdash{}{0pt}%
\pgfsys@defobject{currentmarker}{\pgfqpoint{-0.041667in}{-0.041667in}}{\pgfqpoint{0.041667in}{0.041667in}}{%
\pgfpathmoveto{\pgfqpoint{0.000000in}{-0.041667in}}%
\pgfpathcurveto{\pgfqpoint{0.011050in}{-0.041667in}}{\pgfqpoint{0.021649in}{-0.037276in}}{\pgfqpoint{0.029463in}{-0.029463in}}%
\pgfpathcurveto{\pgfqpoint{0.037276in}{-0.021649in}}{\pgfqpoint{0.041667in}{-0.011050in}}{\pgfqpoint{0.041667in}{0.000000in}}%
\pgfpathcurveto{\pgfqpoint{0.041667in}{0.011050in}}{\pgfqpoint{0.037276in}{0.021649in}}{\pgfqpoint{0.029463in}{0.029463in}}%
\pgfpathcurveto{\pgfqpoint{0.021649in}{0.037276in}}{\pgfqpoint{0.011050in}{0.041667in}}{\pgfqpoint{0.000000in}{0.041667in}}%
\pgfpathcurveto{\pgfqpoint{-0.011050in}{0.041667in}}{\pgfqpoint{-0.021649in}{0.037276in}}{\pgfqpoint{-0.029463in}{0.029463in}}%
\pgfpathcurveto{\pgfqpoint{-0.037276in}{0.021649in}}{\pgfqpoint{-0.041667in}{0.011050in}}{\pgfqpoint{-0.041667in}{0.000000in}}%
\pgfpathcurveto{\pgfqpoint{-0.041667in}{-0.011050in}}{\pgfqpoint{-0.037276in}{-0.021649in}}{\pgfqpoint{-0.029463in}{-0.029463in}}%
\pgfpathcurveto{\pgfqpoint{-0.021649in}{-0.037276in}}{\pgfqpoint{-0.011050in}{-0.041667in}}{\pgfqpoint{0.000000in}{-0.041667in}}%
\pgfpathlineto{\pgfqpoint{0.000000in}{-0.041667in}}%
\pgfpathclose%
\pgfusepath{stroke,fill}%
}%
\begin{pgfscope}%
\pgfsys@transformshift{0.383963in}{0.977641in}%
\pgfsys@useobject{currentmarker}{}%
\end{pgfscope}%
\begin{pgfscope}%
\pgfsys@transformshift{0.492965in}{1.216754in}%
\pgfsys@useobject{currentmarker}{}%
\end{pgfscope}%
\begin{pgfscope}%
\pgfsys@transformshift{0.629218in}{1.319951in}%
\pgfsys@useobject{currentmarker}{}%
\end{pgfscope}%
\begin{pgfscope}%
\pgfsys@transformshift{1.719241in}{1.557321in}%
\pgfsys@useobject{currentmarker}{}%
\end{pgfscope}%
\begin{pgfscope}%
\pgfsys@transformshift{3.081770in}{1.649955in}%
\pgfsys@useobject{currentmarker}{}%
\end{pgfscope}%
\end{pgfscope}%
\begin{pgfscope}%
\pgfpathrectangle{\pgfqpoint{0.249073in}{0.706639in}}{\pgfqpoint{2.967588in}{1.143277in}}%
\pgfusepath{clip}%
\pgfsetrectcap%
\pgfsetroundjoin%
\pgfsetlinewidth{2.007500pt}%
\definecolor{currentstroke}{rgb}{0.580392,0.403922,0.741176}%
\pgfsetstrokecolor{currentstroke}%
\pgfsetdash{}{0pt}%
\pgfpathmoveto{\pgfqpoint{0.383963in}{0.953053in}}%
\pgfpathlineto{\pgfqpoint{0.492965in}{1.155158in}}%
\pgfpathlineto{\pgfqpoint{0.629218in}{1.279297in}}%
\pgfpathlineto{\pgfqpoint{1.719241in}{1.564064in}}%
\pgfpathlineto{\pgfqpoint{3.081770in}{1.634728in}}%
\pgfusepath{stroke}%
\end{pgfscope}%
\begin{pgfscope}%
\pgfpathrectangle{\pgfqpoint{0.249073in}{0.706639in}}{\pgfqpoint{2.967588in}{1.143277in}}%
\pgfusepath{clip}%
\pgfsetbuttcap%
\pgfsetroundjoin%
\definecolor{currentfill}{rgb}{0.580392,0.403922,0.741176}%
\pgfsetfillcolor{currentfill}%
\pgfsetlinewidth{0.752812pt}%
\definecolor{currentstroke}{rgb}{1.000000,1.000000,1.000000}%
\pgfsetstrokecolor{currentstroke}%
\pgfsetdash{}{0pt}%
\pgfsys@defobject{currentmarker}{\pgfqpoint{-0.041667in}{-0.041667in}}{\pgfqpoint{0.041667in}{0.041667in}}{%
\pgfpathmoveto{\pgfqpoint{0.000000in}{-0.041667in}}%
\pgfpathcurveto{\pgfqpoint{0.011050in}{-0.041667in}}{\pgfqpoint{0.021649in}{-0.037276in}}{\pgfqpoint{0.029463in}{-0.029463in}}%
\pgfpathcurveto{\pgfqpoint{0.037276in}{-0.021649in}}{\pgfqpoint{0.041667in}{-0.011050in}}{\pgfqpoint{0.041667in}{0.000000in}}%
\pgfpathcurveto{\pgfqpoint{0.041667in}{0.011050in}}{\pgfqpoint{0.037276in}{0.021649in}}{\pgfqpoint{0.029463in}{0.029463in}}%
\pgfpathcurveto{\pgfqpoint{0.021649in}{0.037276in}}{\pgfqpoint{0.011050in}{0.041667in}}{\pgfqpoint{0.000000in}{0.041667in}}%
\pgfpathcurveto{\pgfqpoint{-0.011050in}{0.041667in}}{\pgfqpoint{-0.021649in}{0.037276in}}{\pgfqpoint{-0.029463in}{0.029463in}}%
\pgfpathcurveto{\pgfqpoint{-0.037276in}{0.021649in}}{\pgfqpoint{-0.041667in}{0.011050in}}{\pgfqpoint{-0.041667in}{0.000000in}}%
\pgfpathcurveto{\pgfqpoint{-0.041667in}{-0.011050in}}{\pgfqpoint{-0.037276in}{-0.021649in}}{\pgfqpoint{-0.029463in}{-0.029463in}}%
\pgfpathcurveto{\pgfqpoint{-0.021649in}{-0.037276in}}{\pgfqpoint{-0.011050in}{-0.041667in}}{\pgfqpoint{0.000000in}{-0.041667in}}%
\pgfpathlineto{\pgfqpoint{0.000000in}{-0.041667in}}%
\pgfpathclose%
\pgfusepath{stroke,fill}%
}%
\begin{pgfscope}%
\pgfsys@transformshift{0.383963in}{0.953053in}%
\pgfsys@useobject{currentmarker}{}%
\end{pgfscope}%
\begin{pgfscope}%
\pgfsys@transformshift{0.492965in}{1.155158in}%
\pgfsys@useobject{currentmarker}{}%
\end{pgfscope}%
\begin{pgfscope}%
\pgfsys@transformshift{0.629218in}{1.279297in}%
\pgfsys@useobject{currentmarker}{}%
\end{pgfscope}%
\begin{pgfscope}%
\pgfsys@transformshift{1.719241in}{1.564064in}%
\pgfsys@useobject{currentmarker}{}%
\end{pgfscope}%
\begin{pgfscope}%
\pgfsys@transformshift{3.081770in}{1.634728in}%
\pgfsys@useobject{currentmarker}{}%
\end{pgfscope}%
\end{pgfscope}%
\begin{pgfscope}%
\pgfpathrectangle{\pgfqpoint{0.249073in}{0.706639in}}{\pgfqpoint{2.967588in}{1.143277in}}%
\pgfusepath{clip}%
\pgfsetbuttcap%
\pgfsetroundjoin%
\pgfsetlinewidth{2.007500pt}%
\definecolor{currentstroke}{rgb}{0.549020,0.337255,0.294118}%
\pgfsetstrokecolor{currentstroke}%
\pgfsetdash{{8.000000pt}{3.000000pt}}{0.000000pt}%
\pgfpathmoveto{\pgfqpoint{0.383963in}{1.146335in}}%
\pgfpathlineto{\pgfqpoint{0.492965in}{1.443362in}}%
\pgfpathlineto{\pgfqpoint{0.629218in}{1.550348in}}%
\pgfpathlineto{\pgfqpoint{1.719241in}{1.741228in}}%
\pgfpathlineto{\pgfqpoint{3.081770in}{1.797948in}}%
\pgfusepath{stroke}%
\end{pgfscope}%
\begin{pgfscope}%
\pgfpathrectangle{\pgfqpoint{0.249073in}{0.706639in}}{\pgfqpoint{2.967588in}{1.143277in}}%
\pgfusepath{clip}%
\pgfsetbuttcap%
\pgfsetmiterjoin%
\definecolor{currentfill}{rgb}{0.549020,0.337255,0.294118}%
\pgfsetfillcolor{currentfill}%
\pgfsetlinewidth{0.752812pt}%
\definecolor{currentstroke}{rgb}{1.000000,1.000000,1.000000}%
\pgfsetstrokecolor{currentstroke}%
\pgfsetdash{}{0pt}%
\pgfsys@defobject{currentmarker}{\pgfqpoint{-0.041667in}{-0.041667in}}{\pgfqpoint{0.041667in}{0.041667in}}{%
\pgfpathmoveto{\pgfqpoint{-0.020833in}{-0.041667in}}%
\pgfpathlineto{\pgfqpoint{0.000000in}{-0.020833in}}%
\pgfpathlineto{\pgfqpoint{0.020833in}{-0.041667in}}%
\pgfpathlineto{\pgfqpoint{0.041667in}{-0.020833in}}%
\pgfpathlineto{\pgfqpoint{0.020833in}{0.000000in}}%
\pgfpathlineto{\pgfqpoint{0.041667in}{0.020833in}}%
\pgfpathlineto{\pgfqpoint{0.020833in}{0.041667in}}%
\pgfpathlineto{\pgfqpoint{0.000000in}{0.020833in}}%
\pgfpathlineto{\pgfqpoint{-0.020833in}{0.041667in}}%
\pgfpathlineto{\pgfqpoint{-0.041667in}{0.020833in}}%
\pgfpathlineto{\pgfqpoint{-0.020833in}{0.000000in}}%
\pgfpathlineto{\pgfqpoint{-0.041667in}{-0.020833in}}%
\pgfpathlineto{\pgfqpoint{-0.020833in}{-0.041667in}}%
\pgfpathclose%
\pgfusepath{stroke,fill}%
}%
\begin{pgfscope}%
\pgfsys@transformshift{0.383963in}{1.146335in}%
\pgfsys@useobject{currentmarker}{}%
\end{pgfscope}%
\begin{pgfscope}%
\pgfsys@transformshift{0.492965in}{1.443362in}%
\pgfsys@useobject{currentmarker}{}%
\end{pgfscope}%
\begin{pgfscope}%
\pgfsys@transformshift{0.629218in}{1.550348in}%
\pgfsys@useobject{currentmarker}{}%
\end{pgfscope}%
\begin{pgfscope}%
\pgfsys@transformshift{1.719241in}{1.741228in}%
\pgfsys@useobject{currentmarker}{}%
\end{pgfscope}%
\begin{pgfscope}%
\pgfsys@transformshift{3.081770in}{1.797948in}%
\pgfsys@useobject{currentmarker}{}%
\end{pgfscope}%
\end{pgfscope}%
\begin{pgfscope}%
\pgfsetrectcap%
\pgfsetmiterjoin%
\pgfsetlinewidth{0.803000pt}%
\definecolor{currentstroke}{rgb}{0.000000,0.000000,0.000000}%
\pgfsetstrokecolor{currentstroke}%
\pgfsetdash{}{0pt}%
\pgfpathmoveto{\pgfqpoint{0.249073in}{0.706639in}}%
\pgfpathlineto{\pgfqpoint{0.249073in}{1.849915in}}%
\pgfusepath{stroke}%
\end{pgfscope}%
\begin{pgfscope}%
\pgfsetrectcap%
\pgfsetmiterjoin%
\pgfsetlinewidth{0.803000pt}%
\definecolor{currentstroke}{rgb}{0.000000,0.000000,0.000000}%
\pgfsetstrokecolor{currentstroke}%
\pgfsetdash{}{0pt}%
\pgfpathmoveto{\pgfqpoint{0.249073in}{0.706639in}}%
\pgfpathlineto{\pgfqpoint{3.216660in}{0.706639in}}%
\pgfusepath{stroke}%
\end{pgfscope}%
\begin{pgfscope}%
\definecolor{textcolor}{rgb}{0.000000,0.000000,0.000000}%
\pgfsetstrokecolor{textcolor}%
\pgfsetfillcolor{textcolor}%
\pgftext[x=1.732866in,y=1.933249in,,base]{\color{textcolor}{\rmfamily\fontsize{8.000000}{9.600000}\selectfont\catcode`\^=\active\def^{\ifmmode\sp\else\^{}\fi}\catcode`\%=\active\def
\end{pgfscope}%
\begin{pgfscope}%
\pgfsetrectcap%
\pgfsetroundjoin%
\pgfsetlinewidth{2.007500pt}%
\definecolor{currentstroke}{rgb}{0.121569,0.466667,0.705882}%
\pgfsetstrokecolor{currentstroke}%
\pgfsetdash{}{0pt}%
\pgfpathmoveto{\pgfqpoint{0.360202in}{0.254508in}}%
\pgfpathlineto{\pgfqpoint{0.443535in}{0.254508in}}%
\pgfpathlineto{\pgfqpoint{0.526869in}{0.254508in}}%
\pgfusepath{stroke}%
\end{pgfscope}%
\begin{pgfscope}%
\definecolor{textcolor}{rgb}{0.000000,0.000000,0.000000}%
\pgfsetstrokecolor{textcolor}%
\pgfsetfillcolor{textcolor}%
\pgftext[x=0.593535in,y=0.225342in,left,base]{\color{textcolor}{\rmfamily\fontsize{6.000000}{7.200000}\selectfont\catcode`\^=\active\def^{\ifmmode\sp\else\^{}\fi}\catcode`\%=\active\def
\end{pgfscope}%
\begin{pgfscope}%
\pgfsetrectcap%
\pgfsetroundjoin%
\pgfsetlinewidth{2.007500pt}%
\definecolor{currentstroke}{rgb}{1.000000,0.498039,0.054902}%
\pgfsetstrokecolor{currentstroke}%
\pgfsetdash{}{0pt}%
\pgfpathmoveto{\pgfqpoint{0.360202in}{0.132194in}}%
\pgfpathlineto{\pgfqpoint{0.443535in}{0.132194in}}%
\pgfpathlineto{\pgfqpoint{0.526869in}{0.132194in}}%
\pgfusepath{stroke}%
\end{pgfscope}%
\begin{pgfscope}%
\definecolor{textcolor}{rgb}{0.000000,0.000000,0.000000}%
\pgfsetstrokecolor{textcolor}%
\pgfsetfillcolor{textcolor}%
\pgftext[x=0.593535in,y=0.103027in,left,base]{\color{textcolor}{\rmfamily\fontsize{6.000000}{7.200000}\selectfont\catcode`\^=\active\def^{\ifmmode\sp\else\^{}\fi}\catcode`\%=\active\def
\end{pgfscope}%
\begin{pgfscope}%
\pgfsetrectcap%
\pgfsetroundjoin%
\pgfsetlinewidth{2.007500pt}%
\definecolor{currentstroke}{rgb}{0.172549,0.627451,0.172549}%
\pgfsetstrokecolor{currentstroke}%
\pgfsetdash{}{0pt}%
\pgfpathmoveto{\pgfqpoint{1.215728in}{0.254508in}}%
\pgfpathlineto{\pgfqpoint{1.299061in}{0.254508in}}%
\pgfpathlineto{\pgfqpoint{1.382394in}{0.254508in}}%
\pgfusepath{stroke}%
\end{pgfscope}%
\begin{pgfscope}%
\definecolor{textcolor}{rgb}{0.000000,0.000000,0.000000}%
\pgfsetstrokecolor{textcolor}%
\pgfsetfillcolor{textcolor}%
\pgftext[x=1.449061in,y=0.225342in,left,base]{\color{textcolor}{\rmfamily\fontsize{6.000000}{7.200000}\selectfont\catcode`\^=\active\def^{\ifmmode\sp\else\^{}\fi}\catcode`\%=\active\def
\end{pgfscope}%
\begin{pgfscope}%
\pgfsetrectcap%
\pgfsetroundjoin%
\pgfsetlinewidth{2.007500pt}%
\definecolor{currentstroke}{rgb}{0.839216,0.152941,0.156863}%
\pgfsetstrokecolor{currentstroke}%
\pgfsetdash{}{0pt}%
\pgfpathmoveto{\pgfqpoint{1.215728in}{0.132194in}}%
\pgfpathlineto{\pgfqpoint{1.299061in}{0.132194in}}%
\pgfpathlineto{\pgfqpoint{1.382394in}{0.132194in}}%
\pgfusepath{stroke}%
\end{pgfscope}%
\begin{pgfscope}%
\definecolor{textcolor}{rgb}{0.000000,0.000000,0.000000}%
\pgfsetstrokecolor{textcolor}%
\pgfsetfillcolor{textcolor}%
\pgftext[x=1.449061in,y=0.103027in,left,base]{\color{textcolor}{\rmfamily\fontsize{6.000000}{7.200000}\selectfont\catcode`\^=\active\def^{\ifmmode\sp\else\^{}\fi}\catcode`\%=\active\def
\end{pgfscope}%
\begin{pgfscope}%
\pgfsetrectcap%
\pgfsetroundjoin%
\pgfsetlinewidth{2.007500pt}%
\definecolor{currentstroke}{rgb}{0.580392,0.403922,0.741176}%
\pgfsetstrokecolor{currentstroke}%
\pgfsetdash{}{0pt}%
\pgfpathmoveto{\pgfqpoint{2.334355in}{0.254508in}}%
\pgfpathlineto{\pgfqpoint{2.417689in}{0.254508in}}%
\pgfpathlineto{\pgfqpoint{2.501022in}{0.254508in}}%
\pgfusepath{stroke}%
\end{pgfscope}%
\begin{pgfscope}%
\definecolor{textcolor}{rgb}{0.000000,0.000000,0.000000}%
\pgfsetstrokecolor{textcolor}%
\pgfsetfillcolor{textcolor}%
\pgftext[x=2.567689in,y=0.225342in,left,base]{\color{textcolor}{\rmfamily\fontsize{6.000000}{7.200000}\selectfont\catcode`\^=\active\def^{\ifmmode\sp\else\^{}\fi}\catcode`\%=\active\def
\end{pgfscope}%
\begin{pgfscope}%
\pgfsetrectcap%
\pgfsetroundjoin%
\pgfsetlinewidth{2.007500pt}%
\definecolor{currentstroke}{rgb}{0.549020,0.337255,0.294118}%
\pgfsetstrokecolor{currentstroke}%
\pgfsetdash{}{0pt}%
\pgfpathmoveto{\pgfqpoint{2.334355in}{0.131014in}}%
\pgfpathlineto{\pgfqpoint{2.417689in}{0.131014in}}%
\pgfpathlineto{\pgfqpoint{2.501022in}{0.131014in}}%
\pgfusepath{stroke}%
\end{pgfscope}%
\begin{pgfscope}%
\definecolor{textcolor}{rgb}{0.000000,0.000000,0.000000}%
\pgfsetstrokecolor{textcolor}%
\pgfsetfillcolor{textcolor}%
\pgftext[x=2.567689in,y=0.101847in,left,base]{\color{textcolor}{\rmfamily\fontsize{6.000000}{7.200000}\selectfont\catcode`\^=\active\def^{\ifmmode\sp\else\^{}\fi}\catcode`\%=\active\def
\end{pgfscope}%
\end{pgfpicture}%
\makeatother%
\endgroup%

%% file: sections/appendix.tex
\pagebreak
\input{sections/related_work.tex}

\section{Chunking methodology}\label{app:chunking}

\subsection{Representing file \& class information}\label{subsec:filehierarchy}
We form \emph{chunks} of code by splitting each file into its constituent classes, methods and functions as
shown in \cref{fig:data_processing}.

\noindent\textbf{File hierarchy:} These code chunks do not contain any information about their locality in the
code repository. To add this information, we insert the file path at the beginning of all code
chunks. This is an important bias for code-editing retrieval as the problem statements $q$'s may often
contain relevant file paths. For instance, on SWE-bench Verified\citep{jimenez2024swebench},
$26\%$ of the problem statements contain the path of at least one ground truth file (i.e. one
file that needs to be edited) (see \cref{tab:file_eda}). Empirically we verify the efficacy of this representation by showing improved
retrieval performance and visualising the attention maps of a trained retrieval model in
\cref{appendix:filepath_exp}.

\noindent\textbf{Class representation:} Similarly, we preserve the class hierarchy within the chunks. We
represent a class by its documentation string, its constructor and declaration of class methods. Each
method is represented by including the class it belongs to.

\section{Metric Definitions}\label{app:sec:metrics}
We measure the performance of our model using standard retrieval metrics: recall@$k$ and mean
reciprocal rank (MRR). We report recall at both a file and a chunk level. We compute file recall by
retrieving $k$ chunks and take the files those chunks have been extracted from. A similar process is
done to the ground truth chunks as well.

\begin{align}
    \text{Recall@}k & = \frac{1}{N}\sum^N_{i=1}\frac{|{\repo_F}_i \cap \repo_i^*|} {|\repo^*_i|} \\
    \text{MRR}      & = \frac{1}{N}\sum^N_{i=1} \frac{1}{\text{rank}(\repo^*_i, {\repo_F}_i)}
\end{align}

Here $\text{rank}(\cdot, \cdot)$ refers to the rank of the first element that is common to both the
arguments. This metric measures the minimum value of $k$ needed to retrieve at least one correct
code chunk, %
and is also known as the First Answer Reciprocal Rank\cite{radev-etal-2002-evaluating}.

\section{Dataset statistics}
\label{appendix:data}
We report relevant dataset statistics
in \cref{tab:data_analysis} and \cref{tab:file_eda}. If an instance does not have at
least one modified function, class or method of class, we discard it for the purpose of computing
reported statistics.
For all token calculations we use the SentencePiece (SP) tokenizer employed by CodeSage
\citep{zhang2024code}.
We extract ground truth (GT) files and chunks using the ground truth patch for each instance,
namely a file is in the ground truth if it is edited in the patch, and the same for other chunks.
Overall, training instances (SWE-bench train) have slightly larger ground truth sets, comprising
on average 5 chunks, than test instances (SWE-bench verified and LCA).
File overlap (GT file overlap in \cref{tab:file_eda}) is overall high across datasets,
suggesting that typically changes are restricted to a single or a small number of files.
Finally, more than a quarter of queries in SWE-bench and more than a third in LCA of queries
contain the path of at least one file to edit (GT file in query, \cref{tab:file_eda}).

\begin{table*}
    \centering
    \resizebox*{\linewidth}{!}{
        \begin{tabular}{ll|ccc|ccc|c|Z}
            \multicolumn{5}{c}{} & \multicolumn{3}{c}{\textbf{Chunk-level}} & \multicolumn{1}{c}{\textbf{Query}} &                                                                                                                                           \\
            \cmidrule{6-9}
            \textbf{Dataset}     &                                          & \textbf{Instances}                 & \textbf{Files} & \textbf{Chunks} & \textbf{GT} & \textbf{Lines} & \textbf{Tokens} & \textbf{Tokens} & \textbf{License}                    \\
            \midrule
            SWE-bench            & Train                                    & $8349$                             & $263_{(269)}$  & $3628_{(3150)}$ & $5_{(18)}$  & $20_{(6)}$     & $198_{(60)}$    & $631_{(1974)}$  & \citet[Tab 12]{jimenez2024swebench} \\
            SWE-bench            & Verified                                 & $435$                              & $420_{(222)}$  & $8551_{(6118)}$ & $2_{(3)}$   & $19_{(7)}$     & $195_{(84)}$    & $539_{(687)}$   & \citet[Tab 12]{jimenez2024swebench} \\
            \midrule
            LCA                  & Test                                     & $34$                               & $210_{(238)}$  & $3243_{(5010)}$ & $4_{(5)}$   & $19_{(8)}$     & $206_{(142)}$   & $911_{(878)}$   & Apache 2.0                          \\
        \end{tabular}}
    \caption{General dataset statistics for SWE-bench dataset. \textbf{GT} is the ground truth number of
        chunks that are edited, \textbf{Lines} the average number of lines and \textbf{Tokens} the average number of
        tokens (with SP tokenizer) edited. Standard deviation are reported in brackets.}
    \label{tab:data_analysis}
\end{table*}

\begin{table*}[t]
    \centering
    \begin{tabular}{ll|cccc}
        \textbf{Dataset} &                           & \textbf{Chunks per File} & \textbf{Files per GT} & \textbf{GT file
        overlap}         & \textbf{GT file in query}                                                                      \\
        \midrule
        SWE-bench        & Train                     & $17_{(8)}$               & $1.97_{(3.06)}$       & $0.83$
                         & $0.28$                                                                                         \\
        SWE-bench        & Verified                  & $22_{(9)}$               & $1.24_{(1.06)}$       & $0.82$
                         & $0.26$                                                                                         \\
        \midrule
        LCA              & Test                      & $16_{(7)}$               & $2.4_{(2.24)}$        & $0.70$
                         & $0.38$                                                                                         \\
    \end{tabular}
    \caption{\textbf{GT file overlap}: On instances where there are multiple chunks to edit, we report the
        empirical probability that at least one file is in common between the chunks to edit.
        \textbf{GT file in query} is the empirical probability that the query contains at least one file path of the ground
        truth files to edit.
    }
    \label{tab:file_eda}
\end{table*}

\begin{table*}[t]
    \centering
    \begin{tabular}{ll|cc}
        \textbf{Dataset} &          & \textbf{Calls}  & \textbf{GT call overlap} \\
        \midrule
        SWE-bench        & Train    & $1.40_{(0.70)}$ & 0.46                     \\
        SWE-bench        & Verified & $1.92_{(2.22)}$ & 0.38                     \\
        SWE-bench        & Lite     & $2.37_{(2.90)}$ & 0.16                     \\
        \midrule
        LCA              & Test     & $1.15_{(0.49)}$ & 0.40                     \\
    \end{tabular}
    \caption{\textbf{GT call overlap}: When there are multiple chunks to edit what is the probability of
        at least 1 chunk being connected by the call graph. The standard deviation is represented in the
        brackets.}
    \label{tab:call graph_eda}
\end{table*}

\section{Baseline Model Details}

\label{sec:app:model-details}

We report in \cref{app:tab:baselines} various statistics about the models and methods we choose as baselines.
Given the trade-offs between the performance and resource requirements, we choose CodeSage S as our base model to build
\cody upon.

\begin{table*}[t]
    \centering
    \begin{tabular}{l|cccrHHHHZZ}
        \multicolumn{9}{c}{}       &                                                                                                                                                                                 \\
        \cmidrule{1-11}

        {Model}                    & {Parms} & {Tok} & {Length}                           & {Encoding} & \textbf{@5} & \textbf{@20} & \textbf{@5} & \textbf{@20} & \textbf{MRR} & {License}                          \\
        \midrule
        BM25                       & -       & words & -                                  & -          & 0.16        & 0.22         & 0.36        & 0.56         & 0.16         & -                                  \\
        CodeBERT                   & 125M    & BPE   & 512                                & 0.5min     & 0.00        & 0.02         &
        0.08                       & 0.27    & 0.01  & Model Unspecified - Code under MIT                                                                                                                            \\
        GraphCodeBERT              & 125M    & BPE   & 512                                & 0.5min     & 0.01        & 0.02         & 0.03        & 0.10         & 0.01         & Model Unspecified - Code under MIT \\
        UniXcoder                  & 125M    & BPE   & 512                                & 0.5min     & 0.30        & 0.45         & 0.57        & 0.77         & 0.27         & Apache-2.0                         \\
        \rowcolor{Gray} CodeSage S & 130M    & SP    & 1024                               & 1.5min     & 0.40        & 0.57         & 0.70        & 0.83         & 0.35         & Apache-2.0                         \\
        CodeSage M                 & 400M    & SP    & 1024                               & 5min       & 0.34        & 0.57         & 0.70        & 0.84         & 0.32         & Apache-2.0                         \\
        CodeSage L                 & 1.3B    & SP    & 1024                               & 22.5min    & -           & -            & -           & -            & -            & Apache-2.0                         \\
        \bottomrule
    \end{tabular}
    \caption{Statistics about baseline models and methods. Parms: number of model parameters; Tok:
        tokenizer with SentencePiece (SP), Byte-Pair Encoding (BPE); Length: maximum number input tokens
        (maximum context length); Encoding the approximate inference time in minutes per 10K chunks, on a
        single GPU. The model we chose for fine-tuning (CodeSage S) is highlighted. Additionally, the models
        have been released under permissive licenses, or their official huggingface repositories do not have
        a license specific.}
    \label{app:tab:baselines}
\end{table*}

\section{Baseline performance}
\begin{figure}[h]
    \centering
    \begin{minipage}{0.49\textwidth}
        \resizebox{.95\linewidth}{!}{
            \input{SWE-Bench-Verified-Chunk_recall_graph.pgf}}
    \end{minipage}
    \begin{minipage}{0.49\textwidth}
        \resizebox{.95\linewidth}{!}{

            \input{SWE-Bench-Verified-File_recall_graph.pgf}}
    \end{minipage}
    \caption{Performance of various models on SWE-bench Verified before fine-tuning. Top: chuck level
        recall; bottom: file-level recall. It is evident that the modern encoder models like the ones from
        the CodeSage family perform substantially better than other baselines. However, those models reach
        acceptable performance (say around $80\%$ chunk recall) only when retrieving a large number of
        chunks ($k\geq 50$). This motivates our need to train models that are specific for code retrieval. }
    \label{app:fig:untrained}
\end{figure}
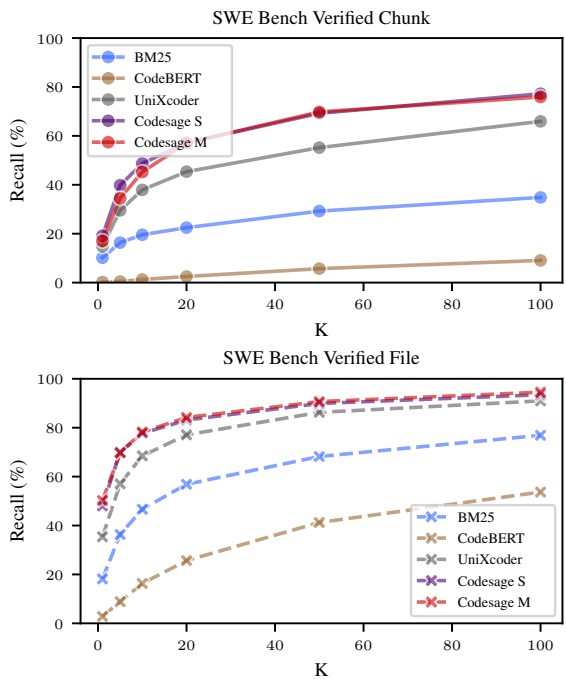

We report in \cref{app:fig:untrained} the performance of various models and methods on SWE-bench Verified (before fine-tuning).

\section{Fine-tuning details}
\label{appendix:hyperparameters}
All experiments use the model CodeSage small\footnote{We use the initial version
    \url{https://huggingface.co/codesage/codesage-small} as the second version was released after
    completion of the project.} \citep{zhang2024code} for comparability and reproducibility. This
Transformer encoder has 6 layers with 8 attention heads per layer. The size for the word embedding
vectors and model projections are 1024, feed-forward dimensions are 4096 which leads to models of
approximately 130 million trainable parameters. We use bf16-mixed precision to reduce the memory
footprint. The model was pre-trained on natural language and code tokenized with a Sentence Piece
Tokenizer with a vocabulary of approximately 50K tokens. Each fine-tuning experiment takes
approximately 24 hours to run on 8 NVIDIA A10G Tensor Core GPUs. During fine-tuning we use the
hyperparameters reported in \cref{tab:hyperparameters}.

\begin{table*}[t]
    \centering
    \begin{tabular}{l|cc}

        \textbf{Hyperparameters} & \textbf{Full} & \textbf{Late Fusion} \\
        \midrule
        Optimiser                & RAdam         & RAdam                \\
        Learning rate            & $5e^{-4}$     & $5e^{-4}$            \\
        Scheduler                & cosine decay  & cosine decay         \\
        Batch size               & $8$           & $8$                  \\
        Gradient acc.            & $32$          & $32$                 \\
        Negatives                & $1024$        & $1024$               \\
        Epochs                   & $4$           & $8$                  \\
    \end{tabular}
    \caption{Hyperparameters for fine-tuning}
    \label{tab:hyperparameters}
\end{table*}

We choose the optimiser hyperparameters to increase stability and remove the need of initial
learning rate warm up. Learning rate was
selected over the grid
$$\{1e^{-3}, 5e^{-4}, 1e^{-4}, 5e^{-5}, 1e^{-5}, 5e^{-6}, 1e^{-6}\}$$
The batch size is 1 instance per GPU over 8 GPUs with gradient
accumulation of 32; namely, we use an effective batch size of 256 instances.
For each instance we sample 1024 (in-repository) negatives during training.

\section{Choice of negative samples in \texorpdfstring{\cref{eqn:final_loss}}{Equation 2}
  influences the performance}
\begin{figure}
    \resizebox{0.9\linewidth}{!}{
        \input{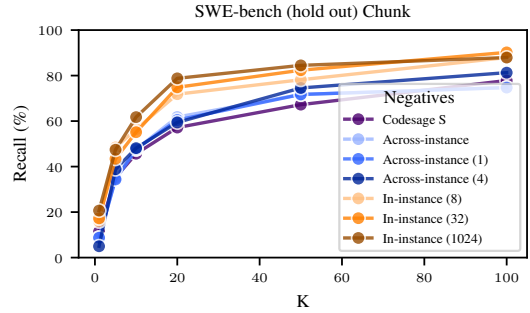}}
    \caption{The influence of number of negatives $|\mathcal{B}|$ and their source.}
    \label{fig:negatives}
\end{figure}
In \cref{fig:negatives}, we show the impact of choice of negatives.
\Citet{karpukhin-etal-2020-dense} propose including hard-negative chunks from BM25 which we select
from the same instance -- we found marginal improvement when considering a single BM25 negative. In
summary, in-instance negatives show a clear advantage, and including more negatives is advantageous
for the performance. These results have been computed on a hold-out set from the SWE-bench train
dataset.

\section{Repository hierarchy experiment}
\label{appendix:filepath_exp}
In this experiment we show that including the file path to the chunks representation
is important for the retrieval
task and aids in maintaining the repository hierarchy. In \cref{tab:filepath}, we show the
retrieval performance of models tested with and without including the file path.
The BM25 and not-fine-tuned CodeSage model achieve minor improvements in retrieval performance,
indicating that fine-tuning is crucial for the model to make use of the added information.

\begin{figure*}[h]
    \centering
    \includegraphics[width=0.85\textwidth]{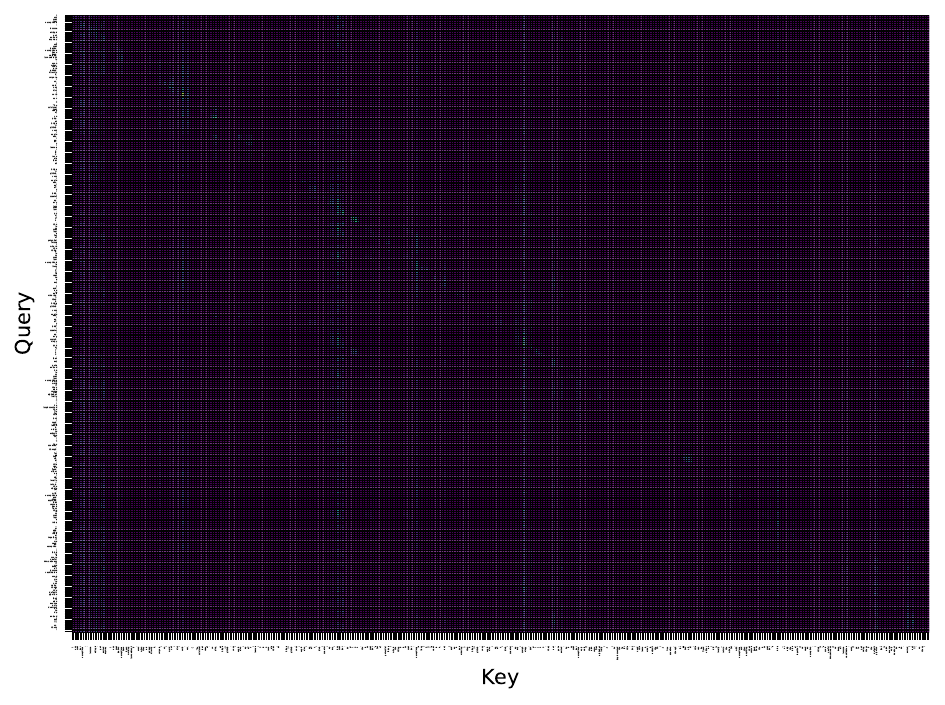} \hfill
    \centering
    \includegraphics[width=0.5\textwidth]{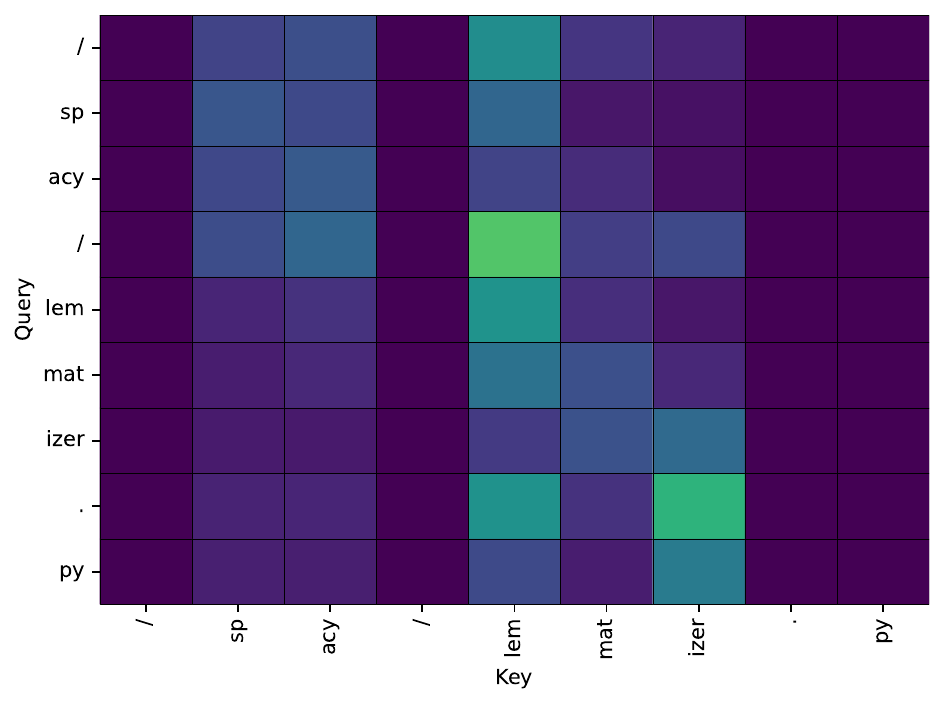}
    \caption{Attention map: query (problem statement) containing file path. Top: full average attention map (best viewed on a screen); bottom: zoom-in portion of the attention map containing the file path. Best viewed in colour on a digital display}
    \label{fig:attention_query}
\end{figure*}

\begin{figure*}[h]
    \centering
    \includegraphics[width=0.5\textwidth]{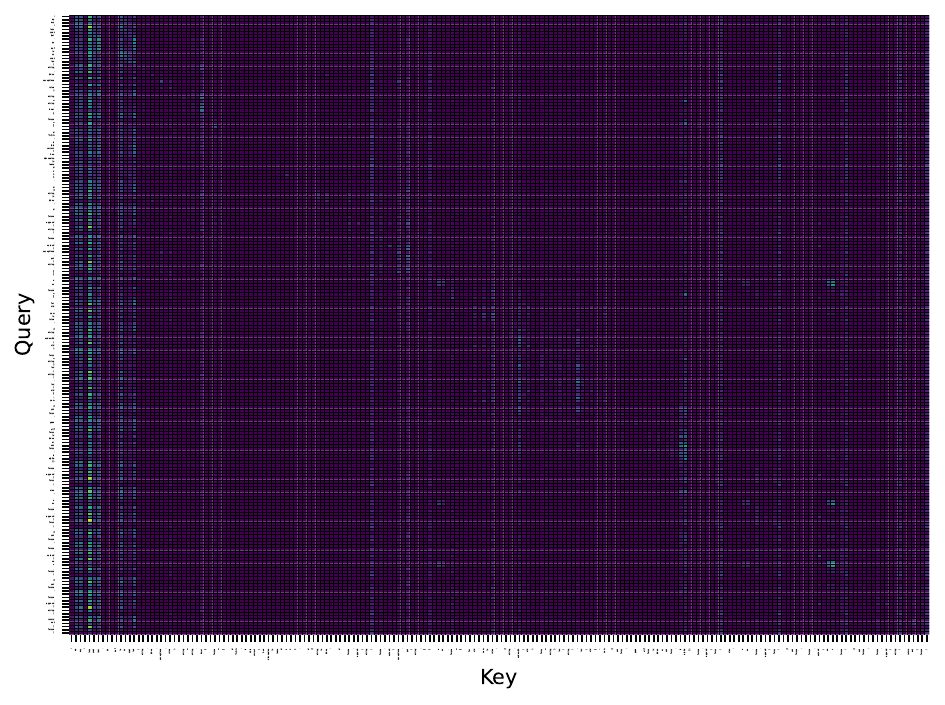}
    \includegraphics[width=0.85\textwidth]{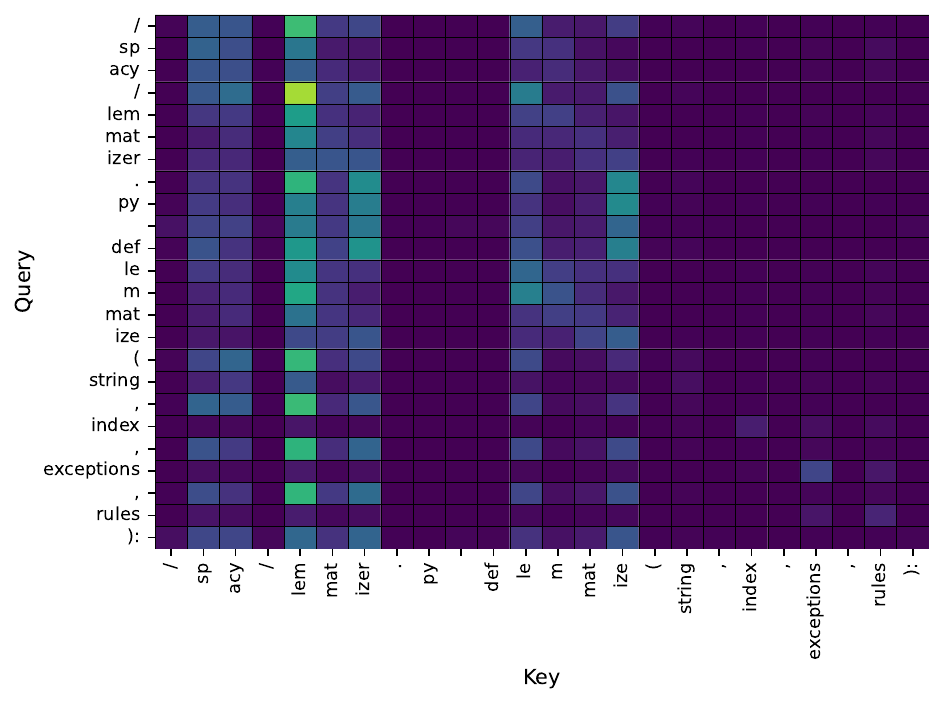} \hfill
    \centering
    \caption{Attention map: code chunk containing file path. Top: full average attention map (best viewed on a screen); bottom: zoom-in portion of the attention map containing the file path. Best viewed in colour on a digital display}
    \label{fig:attention_code}
\end{figure*}

We also investigate to what extent a model trained with file paths rely on them. In
\cref{tab:filepath} in the main paper, we show the retrieval performance drop of a model trained
with file paths once they are removed. We further analyse this by showing the attention plots of a
model trained with file paths when given a query, visualised in \cref{fig:attention_query}. For
this, we consider a repository from the SWE-bench train subset which was not used in our training
set (it is a validation example) and contains the correct file path in the query. We select the
second from the last layer of the model and averaged over all attention heads.
\Cref{fig:attention_query} and \cref{fig:attention_code} shows that the model attends to the file
path both for queries and code chunks and highlights that the model has learned to find and leverage
the path information in a natural language query.

\section{License of artefacts used}
SWE-bench\citep{jimenez2024swebench} dataset packages multiple repositories that are based on BSD
3-Clause, MIT, Apache-2.0, Custom (based on BSD-2 and BSD-3), GPLv2 licenses. The exact details are
available in the original paper in Table 12. The LCA dataset is released under Apache-2, with its
constituent data point from repositories with  MIT and Apache 2.0 licenses.

The code for training the models for CodeBERT and GraphCodeBERT is released under a MIT license at
\url{https://github.com/microsoft/CodeBERT/blob/master/LICENSE} and their weights are released under
with no license specified at \url{https://huggingface.co/microsoft/codebert-base} and
\url{https://huggingface.co/microsoft/graphcodebert-base} respectively. UniXcoder's weights are
released under Apache 2.0 at \url{https://huggingface.co/microsoft/unixcoder-base}. Our primary
baselines, the CodeSage family, have been released under a permissive Apache 2.0 license at
\url{https://huggingface.co/codesage/codesage-small}.

%% file: sections/related_work.tex
\section{Related work}\label{app:sec:related_work}
Here, we present some works related to ours.
\paragraph{Benchmarks for code retrieval}
The scope of standard benchmarks for semantic matching or code search between a natural language
query and code such as {CoSQA}\citep{huang-etal-2021-cosqa},
{CodeSearchNet}\citep{husain_2019_codesearchnet}, {AdvTest}\citep{lu2021codexglue},
{CodeXGLUE}\citep{lu2021codexglue} are based on docstrings and corresponding code.
The repository-level retrieval benchmarks such as {RepoEval} \citep{Zhang2023RepoCoderRC},
{Repobench} \citep{liu2024repobench}, {CodeRAG-Bench}
\citep{wang2024coderagbenchretrievalaugmentcode}, {EvoCodeBench} \citep{li2024evocodebench} are
typically used for code completion tasks where they do not have natural language queries or the
retrieval is only for context and not editing. CoIR\citep{li2024coircomprehensivebenchmarkcode} is a
recent benchmark that curated a collection of multiple datasets for various scenarios, including
text-to-code, code-to-text, code-to-code retrieval, thus going beyond our target setting. Therefore,
we focus our work on real-world GitHub repositories with natural language queries such as
    {SWE-bench} \citep{jimenez2024swebench} {and Long Code Arena} for bug localisation
\citep{bogomolov2024longcodearenaset}. {CoRNStack}
\citep{suresh2024cornstackhighqualitycontrastivedata} is the another dataset that is relevant to us,
however this is a concurrent work to ours and we do not experiment with it.

\paragraph{Models for code-editing retrieval}

Generative Large Language Models (LLMs) are showing improvements on many code related tasks
\citep{li2023starcodersourceyou, Guo2024DeepSeekCoderWT, yang2024qwen2technicalreport}.
\citet{behnamghader2024llmvec} proposes LLM2Vec which allows LLMs to be used for dense retrieval.
LLM2Vec comes at a cost of a much higher parameter count for the model model, making it slower
during inference, more costly to store embeddings, and challenging to fine-tune. \citet{agentless}
propose a simple LLM-only framework to code-editing. Their approach prompts the model to localise,
repair, and validate their solution. Most current approaches use agents which allows an LLM to
interact with the repository through the use of tools such as reading, editing files, and running
bash commands\citep{yang2024sweagentagentcomputerinterfacesenable}. A notable LLM-agent
{AutocodeRover} is provided with specific code search APIs which iteratively retrieve code context
and locate bugs\citep{zhang2024autocoderoverautonomousprogramimprovement}. Further improvements are
seen by {SpecRover} which generates summaries and feedback messages during agent
steps\citep{ruan2024specrovercodeintentextraction}. To improve repository-level navigation,
\citet{Ma2024HowTU} propose an agent {RepoUnderstander} that condenses the codebase into a knowledge
graph and exploit the structure of the repository using Monte Carlo tree search.
\citet{liu-etal-2025-codexgraph} parses a repository into code entities and establishes
relationships between them through a dataflow analysis, forming a repo-specific context graph. This
is shown to improve code completion accuracy.  \citet{liu-etal-2025-codexgraph}
integrate LLM agents with graph database interfaces extracted from code repositories. By leveraging
the structural properties of graph databases and the flexibility of the graph query language,
CodexGraph. RepoGraph\citep{ouyang2024repograph} constructs a graph of code lines, with the nodes
being code lines that capture  definition-reference dependencies. These works are orthogonal ways of
approaching the code retrieval problem through prompting LLMs.

\paragraph{Representation learning for code}
Representation learning for programming languages has benefited many downstream applications.
Different techniques have been applied to learning representation such as masked language modelling
(MLM) \citep{feng-etal-2020-codebert, li2023starcodersourceyou}, next token prediction
\citep{kanade2020pretrained, li2023starcodersourceyou} and contrastive learning
\citep{guo-etal-2022-unixcoder,zhang-etal-2022-syntax, zhang2024code}.
Representation learning seeks to induce meaningful (vector) embeddings of inputs, and is motivated
by applications such as information retrieval via semantic search \citep{Reimers2019SentenceBERTSE,
    Izacard2021UnsupervisedDI, zhang2024code}. Successful application of unsupervised contrastive
learning leverages text-code pairs, mined from docstrings \citep{husain_2019_codesearchnet,
    guo-etal-2022-unixcoder,zhang-etal-2022-syntax, zhang2024code}. These works involve models that
semantically align the embeddings of code to its natural language description. However, they do not
consider the alignment or abstraction required for retrieval for code-editing queries at the
repository-level. Concurrent to our work, CoCoMic\citep{ding-etal-2024-cocomic} shows that including
relevant cross-file context based on import statements significantly improves retrieval.

\paragraph{Repository-level feature learning}
The standard loss for code retrieval is contrastive multi-class N-pair loss or InfoNCE
\citep{sohn2016_npairloss, Oord2018RepresentationLW,chen2020simple}. This loss maximises similarity
between positive pairs while reducing similarity to all other pairs. This loss is used in many
retrieval or semantic search applications \citep{husain_2019_codesearchnet,
    karpukhin-etal-2020-dense, zhang2024code}.
Hard-negative mining selects negative examples that differ from the anchor but have similar
embeddings, making them the most challenging for the model to distinguish
\citep{robinson2021contrastive}. Supervised contrastive learning with hard-negatives has been show
to generally improve retrieval performance \citep{karpukhin-etal-2020-dense, lewis2020_paraphrasing,
    lewis_rag_2020}. However, \citet{xiong2021approximate} argues that in-batch negatives are unlikely
to be hard negatives when the mini-batch size is far smaller than the corpus size and when only a
few negatives are informative.
This has been shown to produce suboptimal training signals for dense passage retrieval across
independent documents. A solution is to use negatives from lexical models such as BM25 during
training\citep{karpukhin-etal-2020-dense, chen2021_complementLexical, luan-etal-2021-sparse}.
Similarly, for code retrieval across independent repositories, we reduce in-batch negatives in
favour of sampled hard negatives within the instance repository. Our formulation closely resembles
standard maximum likelihood estimation. For each repository, the likelihood function is modeled as a
categorical distribution over pairs, each consisting of the query and a repository code chunk. To
reduce computational complexity, we approximate the normalisation factor by sampling a number of
`negative' pairs. Our training loss is then computed by averaging across all repositories. On the
architecture side, we note that other kinds of fusion like feature fusion have been explored in
literature in other contexts\citep{günther2024latechunkingcontextualchunk}, however for this work,
we limit ourselves to input string concatenation for its ease of implementation.

%% file: SWE-Bench-Verified-Chunk_recall_graph.pgf
\begingroup%
\makeatletter%
\begin{pgfpicture}%
\pgfpathrectangle{\pgfpointorigin}{\pgfqpoint{3.196660in}{1.946660in}}%
\pgfusepath{use as bounding box, clip}%
\begin{pgfscope}%
\pgfsetbuttcap%
\pgfsetmiterjoin%
\definecolor{currentfill}{rgb}{1.000000,1.000000,1.000000}%
\pgfsetfillcolor{currentfill}%
\pgfsetlinewidth{0.000000pt}%
\definecolor{currentstroke}{rgb}{1.000000,1.000000,1.000000}%
\pgfsetstrokecolor{currentstroke}%
\pgfsetdash{}{0pt}%
\pgfpathmoveto{\pgfqpoint{0.000000in}{0.000000in}}%
\pgfpathlineto{\pgfqpoint{3.196660in}{0.000000in}}%
\pgfpathlineto{\pgfqpoint{3.196660in}{1.946660in}}%
\pgfpathlineto{\pgfqpoint{0.000000in}{1.946660in}}%
\pgfpathlineto{\pgfqpoint{0.000000in}{0.000000in}}%
\pgfpathclose%
\pgfusepath{fill}%
\end{pgfscope}%
\begin{pgfscope}%
\pgfsetbuttcap%
\pgfsetmiterjoin%
\definecolor{currentfill}{rgb}{1.000000,1.000000,1.000000}%
\pgfsetfillcolor{currentfill}%
\pgfsetlinewidth{0.000000pt}%
\definecolor{currentstroke}{rgb}{0.000000,0.000000,0.000000}%
\pgfsetstrokecolor{currentstroke}%
\pgfsetstrokeopacity{0.000000}%
\pgfsetdash{}{0pt}%
\pgfpathmoveto{\pgfqpoint{0.428083in}{0.355956in}}%
\pgfpathlineto{\pgfqpoint{3.181660in}{0.355956in}}%
\pgfpathlineto{\pgfqpoint{3.181660in}{1.763908in}}%
\pgfpathlineto{\pgfqpoint{0.428083in}{1.763908in}}%
\pgfpathlineto{\pgfqpoint{0.428083in}{0.355956in}}%
\pgfpathclose%
\pgfusepath{fill}%
\end{pgfscope}%
\begin{pgfscope}%
\pgfsetbuttcap%
\pgfsetroundjoin%
\definecolor{currentfill}{rgb}{0.000000,0.000000,0.000000}%
\pgfsetfillcolor{currentfill}%
\pgfsetlinewidth{0.803000pt}%
\definecolor{currentstroke}{rgb}{0.000000,0.000000,0.000000}%
\pgfsetstrokecolor{currentstroke}%
\pgfsetdash{}{0pt}%
\pgfsys@defobject{currentmarker}{\pgfqpoint{0.000000in}{-0.048611in}}{\pgfqpoint{0.000000in}{0.000000in}}{%
\pgfpathmoveto{\pgfqpoint{0.000000in}{0.000000in}}%
\pgfpathlineto{\pgfqpoint{0.000000in}{-0.048611in}}%
\pgfusepath{stroke,fill}%
}%
\begin{pgfscope}%
\pgfsys@transformshift{0.527961in}{0.355956in}%
\pgfsys@useobject{currentmarker}{}%
\end{pgfscope}%
\end{pgfscope}%
\begin{pgfscope}%
\definecolor{textcolor}{rgb}{0.000000,0.000000,0.000000}%
\pgfsetstrokecolor{textcolor}%
\pgfsetfillcolor{textcolor}%
\pgftext[x=0.527961in,y=0.258733in,,top]{\color{textcolor}{\rmfamily\fontsize{6.000000}{7.200000}\selectfont\catcode`\^=\active\def^{\ifmmode\sp\else\^{}\fi}\catcode`\%=\active\def
\end{pgfscope}%
\begin{pgfscope}%
\pgfsetbuttcap%
\pgfsetroundjoin%
\definecolor{currentfill}{rgb}{0.000000,0.000000,0.000000}%
\pgfsetfillcolor{currentfill}%
\pgfsetlinewidth{0.803000pt}%
\definecolor{currentstroke}{rgb}{0.000000,0.000000,0.000000}%
\pgfsetstrokecolor{currentstroke}%
\pgfsetdash{}{0pt}%
\pgfsys@defobject{currentmarker}{\pgfqpoint{0.000000in}{-0.048611in}}{\pgfqpoint{0.000000in}{0.000000in}}{%
\pgfpathmoveto{\pgfqpoint{0.000000in}{0.000000in}}%
\pgfpathlineto{\pgfqpoint{0.000000in}{-0.048611in}}%
\pgfusepath{stroke,fill}%
}%
\begin{pgfscope}%
\pgfsys@transformshift{1.033668in}{0.355956in}%
\pgfsys@useobject{currentmarker}{}%
\end{pgfscope}%
\end{pgfscope}%
\begin{pgfscope}%
\definecolor{textcolor}{rgb}{0.000000,0.000000,0.000000}%
\pgfsetstrokecolor{textcolor}%
\pgfsetfillcolor{textcolor}%
\pgftext[x=1.033668in,y=0.258733in,,top]{\color{textcolor}{\rmfamily\fontsize{6.000000}{7.200000}\selectfont\catcode`\^=\active\def^{\ifmmode\sp\else\^{}\fi}\catcode`\%=\active\def
\end{pgfscope}%
\begin{pgfscope}%
\pgfsetbuttcap%
\pgfsetroundjoin%
\definecolor{currentfill}{rgb}{0.000000,0.000000,0.000000}%
\pgfsetfillcolor{currentfill}%
\pgfsetlinewidth{0.803000pt}%
\definecolor{currentstroke}{rgb}{0.000000,0.000000,0.000000}%
\pgfsetstrokecolor{currentstroke}%
\pgfsetdash{}{0pt}%
\pgfsys@defobject{currentmarker}{\pgfqpoint{0.000000in}{-0.048611in}}{\pgfqpoint{0.000000in}{0.000000in}}{%
\pgfpathmoveto{\pgfqpoint{0.000000in}{0.000000in}}%
\pgfpathlineto{\pgfqpoint{0.000000in}{-0.048611in}}%
\pgfusepath{stroke,fill}%
}%
\begin{pgfscope}%
\pgfsys@transformshift{1.539375in}{0.355956in}%
\pgfsys@useobject{currentmarker}{}%
\end{pgfscope}%
\end{pgfscope}%
\begin{pgfscope}%
\definecolor{textcolor}{rgb}{0.000000,0.000000,0.000000}%
\pgfsetstrokecolor{textcolor}%
\pgfsetfillcolor{textcolor}%
\pgftext[x=1.539375in,y=0.258733in,,top]{\color{textcolor}{\rmfamily\fontsize{6.000000}{7.200000}\selectfont\catcode`\^=\active\def^{\ifmmode\sp\else\^{}\fi}\catcode`\%=\active\def
\end{pgfscope}%
\begin{pgfscope}%
\pgfsetbuttcap%
\pgfsetroundjoin%
\definecolor{currentfill}{rgb}{0.000000,0.000000,0.000000}%
\pgfsetfillcolor{currentfill}%
\pgfsetlinewidth{0.803000pt}%
\definecolor{currentstroke}{rgb}{0.000000,0.000000,0.000000}%
\pgfsetstrokecolor{currentstroke}%
\pgfsetdash{}{0pt}%
\pgfsys@defobject{currentmarker}{\pgfqpoint{0.000000in}{-0.048611in}}{\pgfqpoint{0.000000in}{0.000000in}}{%
\pgfpathmoveto{\pgfqpoint{0.000000in}{0.000000in}}%
\pgfpathlineto{\pgfqpoint{0.000000in}{-0.048611in}}%
\pgfusepath{stroke,fill}%
}%
\begin{pgfscope}%
\pgfsys@transformshift{2.045083in}{0.355956in}%
\pgfsys@useobject{currentmarker}{}%
\end{pgfscope}%
\end{pgfscope}%
\begin{pgfscope}%
\definecolor{textcolor}{rgb}{0.000000,0.000000,0.000000}%
\pgfsetstrokecolor{textcolor}%
\pgfsetfillcolor{textcolor}%
\pgftext[x=2.045083in,y=0.258733in,,top]{\color{textcolor}{\rmfamily\fontsize{6.000000}{7.200000}\selectfont\catcode`\^=\active\def^{\ifmmode\sp\else\^{}\fi}\catcode`\%=\active\def
\end{pgfscope}%
\begin{pgfscope}%
\pgfsetbuttcap%
\pgfsetroundjoin%
\definecolor{currentfill}{rgb}{0.000000,0.000000,0.000000}%
\pgfsetfillcolor{currentfill}%
\pgfsetlinewidth{0.803000pt}%
\definecolor{currentstroke}{rgb}{0.000000,0.000000,0.000000}%
\pgfsetstrokecolor{currentstroke}%
\pgfsetdash{}{0pt}%
\pgfsys@defobject{currentmarker}{\pgfqpoint{0.000000in}{-0.048611in}}{\pgfqpoint{0.000000in}{0.000000in}}{%
\pgfpathmoveto{\pgfqpoint{0.000000in}{0.000000in}}%
\pgfpathlineto{\pgfqpoint{0.000000in}{-0.048611in}}%
\pgfusepath{stroke,fill}%
}%
\begin{pgfscope}%
\pgfsys@transformshift{2.550790in}{0.355956in}%
\pgfsys@useobject{currentmarker}{}%
\end{pgfscope}%
\end{pgfscope}%
\begin{pgfscope}%
\definecolor{textcolor}{rgb}{0.000000,0.000000,0.000000}%
\pgfsetstrokecolor{textcolor}%
\pgfsetfillcolor{textcolor}%
\pgftext[x=2.550790in,y=0.258733in,,top]{\color{textcolor}{\rmfamily\fontsize{6.000000}{7.200000}\selectfont\catcode`\^=\active\def^{\ifmmode\sp\else\^{}\fi}\catcode`\%=\active\def
\end{pgfscope}%
\begin{pgfscope}%
\pgfsetbuttcap%
\pgfsetroundjoin%
\definecolor{currentfill}{rgb}{0.000000,0.000000,0.000000}%
\pgfsetfillcolor{currentfill}%
\pgfsetlinewidth{0.803000pt}%
\definecolor{currentstroke}{rgb}{0.000000,0.000000,0.000000}%
\pgfsetstrokecolor{currentstroke}%
\pgfsetdash{}{0pt}%
\pgfsys@defobject{currentmarker}{\pgfqpoint{0.000000in}{-0.048611in}}{\pgfqpoint{0.000000in}{0.000000in}}{%
\pgfpathmoveto{\pgfqpoint{0.000000in}{0.000000in}}%
\pgfpathlineto{\pgfqpoint{0.000000in}{-0.048611in}}%
\pgfusepath{stroke,fill}%
}%
\begin{pgfscope}%
\pgfsys@transformshift{3.056497in}{0.355956in}%
\pgfsys@useobject{currentmarker}{}%
\end{pgfscope}%
\end{pgfscope}%
\begin{pgfscope}%
\definecolor{textcolor}{rgb}{0.000000,0.000000,0.000000}%
\pgfsetstrokecolor{textcolor}%
\pgfsetfillcolor{textcolor}%
\pgftext[x=3.056497in,y=0.258733in,,top]{\color{textcolor}{\rmfamily\fontsize{6.000000}{7.200000}\selectfont\catcode`\^=\active\def^{\ifmmode\sp\else\^{}\fi}\catcode`\%=\active\def
\end{pgfscope}%
\begin{pgfscope}%
\definecolor{textcolor}{rgb}{0.000000,0.000000,0.000000}%
\pgfsetstrokecolor{textcolor}%
\pgfsetfillcolor{textcolor}%
\pgftext[x=1.804872in,y=0.122530in,,top]{\color{textcolor}{\rmfamily\fontsize{8.000000}{9.600000}\selectfont\catcode`\^=\active\def^{\ifmmode\sp\else\^{}\fi}\catcode`\%=\active\def
\end{pgfscope}%
\begin{pgfscope}%
\pgfsetbuttcap%
\pgfsetroundjoin%
\definecolor{currentfill}{rgb}{0.000000,0.000000,0.000000}%
\pgfsetfillcolor{currentfill}%
\pgfsetlinewidth{0.803000pt}%
\definecolor{currentstroke}{rgb}{0.000000,0.000000,0.000000}%
\pgfsetstrokecolor{currentstroke}%
\pgfsetdash{}{0pt}%
\pgfsys@defobject{currentmarker}{\pgfqpoint{-0.048611in}{0.000000in}}{\pgfqpoint{-0.000000in}{0.000000in}}{%
\pgfpathmoveto{\pgfqpoint{-0.000000in}{0.000000in}}%
\pgfpathlineto{\pgfqpoint{-0.048611in}{0.000000in}}%
\pgfusepath{stroke,fill}%
}%
\begin{pgfscope}%
\pgfsys@transformshift{0.428083in}{0.355956in}%
\pgfsys@useobject{currentmarker}{}%
\end{pgfscope}%
\end{pgfscope}%
\begin{pgfscope}%
\definecolor{textcolor}{rgb}{0.000000,0.000000,0.000000}%
\pgfsetstrokecolor{textcolor}%
\pgfsetfillcolor{textcolor}%
\pgftext[x=0.279936in, y=0.324299in, left, base]{\color{textcolor}{\rmfamily\fontsize{6.000000}{7.200000}\selectfont\catcode`\^=\active\def^{\ifmmode\sp\else\^{}\fi}\catcode`\%=\active\def
\end{pgfscope}%
\begin{pgfscope}%
\pgfsetbuttcap%
\pgfsetroundjoin%
\definecolor{currentfill}{rgb}{0.000000,0.000000,0.000000}%
\pgfsetfillcolor{currentfill}%
\pgfsetlinewidth{0.803000pt}%
\definecolor{currentstroke}{rgb}{0.000000,0.000000,0.000000}%
\pgfsetstrokecolor{currentstroke}%
\pgfsetdash{}{0pt}%
\pgfsys@defobject{currentmarker}{\pgfqpoint{-0.048611in}{0.000000in}}{\pgfqpoint{-0.000000in}{0.000000in}}{%
\pgfpathmoveto{\pgfqpoint{-0.000000in}{0.000000in}}%
\pgfpathlineto{\pgfqpoint{-0.048611in}{0.000000in}}%
\pgfusepath{stroke,fill}%
}%
\begin{pgfscope}%
\pgfsys@transformshift{0.428083in}{0.637546in}%
\pgfsys@useobject{currentmarker}{}%
\end{pgfscope}%
\end{pgfscope}%
\begin{pgfscope}%
\definecolor{textcolor}{rgb}{0.000000,0.000000,0.000000}%
\pgfsetstrokecolor{textcolor}%
\pgfsetfillcolor{textcolor}%
\pgftext[x=0.229011in, y=0.605889in, left, base]{\color{textcolor}{\rmfamily\fontsize{6.000000}{7.200000}\selectfont\catcode`\^=\active\def^{\ifmmode\sp\else\^{}\fi}\catcode`\%=\active\def
\end{pgfscope}%
\begin{pgfscope}%
\pgfsetbuttcap%
\pgfsetroundjoin%
\definecolor{currentfill}{rgb}{0.000000,0.000000,0.000000}%
\pgfsetfillcolor{currentfill}%
\pgfsetlinewidth{0.803000pt}%
\definecolor{currentstroke}{rgb}{0.000000,0.000000,0.000000}%
\pgfsetstrokecolor{currentstroke}%
\pgfsetdash{}{0pt}%
\pgfsys@defobject{currentmarker}{\pgfqpoint{-0.048611in}{0.000000in}}{\pgfqpoint{-0.000000in}{0.000000in}}{%
\pgfpathmoveto{\pgfqpoint{-0.000000in}{0.000000in}}%
\pgfpathlineto{\pgfqpoint{-0.048611in}{0.000000in}}%
\pgfusepath{stroke,fill}%
}%
\begin{pgfscope}%
\pgfsys@transformshift{0.428083in}{0.919137in}%
\pgfsys@useobject{currentmarker}{}%
\end{pgfscope}%
\end{pgfscope}%
\begin{pgfscope}%
\definecolor{textcolor}{rgb}{0.000000,0.000000,0.000000}%
\pgfsetstrokecolor{textcolor}%
\pgfsetfillcolor{textcolor}%
\pgftext[x=0.229011in, y=0.887480in, left, base]{\color{textcolor}{\rmfamily\fontsize{6.000000}{7.200000}\selectfont\catcode`\^=\active\def^{\ifmmode\sp\else\^{}\fi}\catcode`\%=\active\def
\end{pgfscope}%
\begin{pgfscope}%
\pgfsetbuttcap%
\pgfsetroundjoin%
\definecolor{currentfill}{rgb}{0.000000,0.000000,0.000000}%
\pgfsetfillcolor{currentfill}%
\pgfsetlinewidth{0.803000pt}%
\definecolor{currentstroke}{rgb}{0.000000,0.000000,0.000000}%
\pgfsetstrokecolor{currentstroke}%
\pgfsetdash{}{0pt}%
\pgfsys@defobject{currentmarker}{\pgfqpoint{-0.048611in}{0.000000in}}{\pgfqpoint{-0.000000in}{0.000000in}}{%
\pgfpathmoveto{\pgfqpoint{-0.000000in}{0.000000in}}%
\pgfpathlineto{\pgfqpoint{-0.048611in}{0.000000in}}%
\pgfusepath{stroke,fill}%
}%
\begin{pgfscope}%
\pgfsys@transformshift{0.428083in}{1.200727in}%
\pgfsys@useobject{currentmarker}{}%
\end{pgfscope}%
\end{pgfscope}%
\begin{pgfscope}%
\definecolor{textcolor}{rgb}{0.000000,0.000000,0.000000}%
\pgfsetstrokecolor{textcolor}%
\pgfsetfillcolor{textcolor}%
\pgftext[x=0.229011in, y=1.169070in, left, base]{\color{textcolor}{\rmfamily\fontsize{6.000000}{7.200000}\selectfont\catcode`\^=\active\def^{\ifmmode\sp\else\^{}\fi}\catcode`\%=\active\def
\end{pgfscope}%
\begin{pgfscope}%
\pgfsetbuttcap%
\pgfsetroundjoin%
\definecolor{currentfill}{rgb}{0.000000,0.000000,0.000000}%
\pgfsetfillcolor{currentfill}%
\pgfsetlinewidth{0.803000pt}%
\definecolor{currentstroke}{rgb}{0.000000,0.000000,0.000000}%
\pgfsetstrokecolor{currentstroke}%
\pgfsetdash{}{0pt}%
\pgfsys@defobject{currentmarker}{\pgfqpoint{-0.048611in}{0.000000in}}{\pgfqpoint{-0.000000in}{0.000000in}}{%
\pgfpathmoveto{\pgfqpoint{-0.000000in}{0.000000in}}%
\pgfpathlineto{\pgfqpoint{-0.048611in}{0.000000in}}%
\pgfusepath{stroke,fill}%
}%
\begin{pgfscope}%
\pgfsys@transformshift{0.428083in}{1.482318in}%
\pgfsys@useobject{currentmarker}{}%
\end{pgfscope}%
\end{pgfscope}%
\begin{pgfscope}%
\definecolor{textcolor}{rgb}{0.000000,0.000000,0.000000}%
\pgfsetstrokecolor{textcolor}%
\pgfsetfillcolor{textcolor}%
\pgftext[x=0.229011in, y=1.450661in, left, base]{\color{textcolor}{\rmfamily\fontsize{6.000000}{7.200000}\selectfont\catcode`\^=\active\def^{\ifmmode\sp\else\^{}\fi}\catcode`\%=\active\def
\end{pgfscope}%
\begin{pgfscope}%
\pgfsetbuttcap%
\pgfsetroundjoin%
\definecolor{currentfill}{rgb}{0.000000,0.000000,0.000000}%
\pgfsetfillcolor{currentfill}%
\pgfsetlinewidth{0.803000pt}%
\definecolor{currentstroke}{rgb}{0.000000,0.000000,0.000000}%
\pgfsetstrokecolor{currentstroke}%
\pgfsetdash{}{0pt}%
\pgfsys@defobject{currentmarker}{\pgfqpoint{-0.048611in}{0.000000in}}{\pgfqpoint{-0.000000in}{0.000000in}}{%
\pgfpathmoveto{\pgfqpoint{-0.000000in}{0.000000in}}%
\pgfpathlineto{\pgfqpoint{-0.048611in}{0.000000in}}%
\pgfusepath{stroke,fill}%
}%
\begin{pgfscope}%
\pgfsys@transformshift{0.428083in}{1.763908in}%
\pgfsys@useobject{currentmarker}{}%
\end{pgfscope}%
\end{pgfscope}%
\begin{pgfscope}%
\definecolor{textcolor}{rgb}{0.000000,0.000000,0.000000}%
\pgfsetstrokecolor{textcolor}%
\pgfsetfillcolor{textcolor}%
\pgftext[x=0.178086in, y=1.732251in, left, base]{\color{textcolor}{\rmfamily\fontsize{6.000000}{7.200000}\selectfont\catcode`\^=\active\def^{\ifmmode\sp\else\^{}\fi}\catcode`\%=\active\def
\end{pgfscope}%
\begin{pgfscope}%
\definecolor{textcolor}{rgb}{0.000000,0.000000,0.000000}%
\pgfsetstrokecolor{textcolor}%
\pgfsetfillcolor{textcolor}%
\pgftext[x=0.122530in,y=1.059932in,,bottom,rotate=90.000000]{\color{textcolor}{\rmfamily\fontsize{8.000000}{9.600000}\selectfont\catcode`\^=\active\def^{\ifmmode\sp\else\^{}\fi}\catcode`\%=\active\def
\end{pgfscope}%
\begin{pgfscope}%
\pgfpathrectangle{\pgfqpoint{0.428083in}{0.355956in}}{\pgfqpoint{2.753577in}{1.407952in}}%
\pgfusepath{clip}%
\pgfsetrectcap%
\pgfsetroundjoin%
\pgfsetlinewidth{1.505625pt}%
\definecolor{currentstroke}{rgb}{0.200000,0.400000,1.000000}%
\pgfsetstrokecolor{currentstroke}%
\pgfsetstrokeopacity{0.600000}%
\pgfsetdash{}{0pt}%
\pgfpathmoveto{\pgfqpoint{0.553246in}{0.498909in}}%
\pgfpathlineto{\pgfqpoint{0.654388in}{0.586110in}}%
\pgfpathlineto{\pgfqpoint{0.780814in}{0.631388in}}%
\pgfpathlineto{\pgfqpoint{1.033668in}{0.672206in}}%
\pgfpathlineto{\pgfqpoint{1.792229in}{0.767508in}}%
\pgfpathlineto{\pgfqpoint{3.056497in}{0.846502in}}%
\pgfusepath{stroke}%
\end{pgfscope}%
\begin{pgfscope}%
\pgfpathrectangle{\pgfqpoint{0.428083in}{0.355956in}}{\pgfqpoint{2.753577in}{1.407952in}}%
\pgfusepath{clip}%
\pgfsetbuttcap%
\pgfsetroundjoin%
\definecolor{currentfill}{rgb}{0.200000,0.400000,1.000000}%
\pgfsetfillcolor{currentfill}%
\pgfsetfillopacity{0.600000}%
\pgfsetlinewidth{0.752812pt}%
\definecolor{currentstroke}{rgb}{1.000000,1.000000,1.000000}%
\pgfsetstrokecolor{currentstroke}%
\pgfsetstrokeopacity{0.600000}%
\pgfsetdash{}{0pt}%
\pgfsys@defobject{currentmarker}{\pgfqpoint{-0.041667in}{-0.041667in}}{\pgfqpoint{0.041667in}{0.041667in}}{%
\pgfpathmoveto{\pgfqpoint{0.000000in}{-0.041667in}}%
\pgfpathcurveto{\pgfqpoint{0.011050in}{-0.041667in}}{\pgfqpoint{0.021649in}{-0.037276in}}{\pgfqpoint{0.029463in}{-0.029463in}}%
\pgfpathcurveto{\pgfqpoint{0.037276in}{-0.021649in}}{\pgfqpoint{0.041667in}{-0.011050in}}{\pgfqpoint{0.041667in}{0.000000in}}%
\pgfpathcurveto{\pgfqpoint{0.041667in}{0.011050in}}{\pgfqpoint{0.037276in}{0.021649in}}{\pgfqpoint{0.029463in}{0.029463in}}%
\pgfpathcurveto{\pgfqpoint{0.021649in}{0.037276in}}{\pgfqpoint{0.011050in}{0.041667in}}{\pgfqpoint{0.000000in}{0.041667in}}%
\pgfpathcurveto{\pgfqpoint{-0.011050in}{0.041667in}}{\pgfqpoint{-0.021649in}{0.037276in}}{\pgfqpoint{-0.029463in}{0.029463in}}%
\pgfpathcurveto{\pgfqpoint{-0.037276in}{0.021649in}}{\pgfqpoint{-0.041667in}{0.011050in}}{\pgfqpoint{-0.041667in}{0.000000in}}%
\pgfpathcurveto{\pgfqpoint{-0.041667in}{-0.011050in}}{\pgfqpoint{-0.037276in}{-0.021649in}}{\pgfqpoint{-0.029463in}{-0.029463in}}%
\pgfpathcurveto{\pgfqpoint{-0.021649in}{-0.037276in}}{\pgfqpoint{-0.011050in}{-0.041667in}}{\pgfqpoint{0.000000in}{-0.041667in}}%
\pgfpathlineto{\pgfqpoint{0.000000in}{-0.041667in}}%
\pgfpathclose%
\pgfusepath{stroke,fill}%
}%
\begin{pgfscope}%
\pgfsys@transformshift{0.553246in}{0.498909in}%
\pgfsys@useobject{currentmarker}{}%
\end{pgfscope}%
\begin{pgfscope}%
\pgfsys@transformshift{0.654388in}{0.586110in}%
\pgfsys@useobject{currentmarker}{}%
\end{pgfscope}%
\begin{pgfscope}%
\pgfsys@transformshift{0.780814in}{0.631388in}%
\pgfsys@useobject{currentmarker}{}%
\end{pgfscope}%
\begin{pgfscope}%
\pgfsys@transformshift{1.033668in}{0.672206in}%
\pgfsys@useobject{currentmarker}{}%
\end{pgfscope}%
\begin{pgfscope}%
\pgfsys@transformshift{1.792229in}{0.767508in}%
\pgfsys@useobject{currentmarker}{}%
\end{pgfscope}%
\begin{pgfscope}%
\pgfsys@transformshift{3.056497in}{0.846502in}%
\pgfsys@useobject{currentmarker}{}%
\end{pgfscope}%
\end{pgfscope}%
\begin{pgfscope}%
\pgfpathrectangle{\pgfqpoint{0.428083in}{0.355956in}}{\pgfqpoint{2.753577in}{1.407952in}}%
\pgfusepath{clip}%
\pgfsetrectcap%
\pgfsetroundjoin%
\pgfsetlinewidth{1.505625pt}%
\definecolor{currentstroke}{rgb}{0.600000,0.400000,0.200000}%
\pgfsetstrokecolor{currentstroke}%
\pgfsetstrokeopacity{0.600000}%
\pgfsetdash{}{0pt}%
\pgfpathmoveto{\pgfqpoint{0.553246in}{0.359192in}}%
\pgfpathlineto{\pgfqpoint{0.654388in}{0.361890in}}%
\pgfpathlineto{\pgfqpoint{0.780814in}{0.373892in}}%
\pgfpathlineto{\pgfqpoint{1.033668in}{0.390885in}}%
\pgfpathlineto{\pgfqpoint{1.792229in}{0.436313in}}%
\pgfpathlineto{\pgfqpoint{3.056497in}{0.483620in}}%
\pgfusepath{stroke}%
\end{pgfscope}%
\begin{pgfscope}%
\pgfpathrectangle{\pgfqpoint{0.428083in}{0.355956in}}{\pgfqpoint{2.753577in}{1.407952in}}%
\pgfusepath{clip}%
\pgfsetbuttcap%
\pgfsetroundjoin%
\definecolor{currentfill}{rgb}{0.600000,0.400000,0.200000}%
\pgfsetfillcolor{currentfill}%
\pgfsetfillopacity{0.600000}%
\pgfsetlinewidth{0.752812pt}%
\definecolor{currentstroke}{rgb}{1.000000,1.000000,1.000000}%
\pgfsetstrokecolor{currentstroke}%
\pgfsetstrokeopacity{0.600000}%
\pgfsetdash{}{0pt}%
\pgfsys@defobject{currentmarker}{\pgfqpoint{-0.041667in}{-0.041667in}}{\pgfqpoint{0.041667in}{0.041667in}}{%
\pgfpathmoveto{\pgfqpoint{0.000000in}{-0.041667in}}%
\pgfpathcurveto{\pgfqpoint{0.011050in}{-0.041667in}}{\pgfqpoint{0.021649in}{-0.037276in}}{\pgfqpoint{0.029463in}{-0.029463in}}%
\pgfpathcurveto{\pgfqpoint{0.037276in}{-0.021649in}}{\pgfqpoint{0.041667in}{-0.011050in}}{\pgfqpoint{0.041667in}{0.000000in}}%
\pgfpathcurveto{\pgfqpoint{0.041667in}{0.011050in}}{\pgfqpoint{0.037276in}{0.021649in}}{\pgfqpoint{0.029463in}{0.029463in}}%
\pgfpathcurveto{\pgfqpoint{0.021649in}{0.037276in}}{\pgfqpoint{0.011050in}{0.041667in}}{\pgfqpoint{0.000000in}{0.041667in}}%
\pgfpathcurveto{\pgfqpoint{-0.011050in}{0.041667in}}{\pgfqpoint{-0.021649in}{0.037276in}}{\pgfqpoint{-0.029463in}{0.029463in}}%
\pgfpathcurveto{\pgfqpoint{-0.037276in}{0.021649in}}{\pgfqpoint{-0.041667in}{0.011050in}}{\pgfqpoint{-0.041667in}{0.000000in}}%
\pgfpathcurveto{\pgfqpoint{-0.041667in}{-0.011050in}}{\pgfqpoint{-0.037276in}{-0.021649in}}{\pgfqpoint{-0.029463in}{-0.029463in}}%
\pgfpathcurveto{\pgfqpoint{-0.021649in}{-0.037276in}}{\pgfqpoint{-0.011050in}{-0.041667in}}{\pgfqpoint{0.000000in}{-0.041667in}}%
\pgfpathlineto{\pgfqpoint{0.000000in}{-0.041667in}}%
\pgfpathclose%
\pgfusepath{stroke,fill}%
}%
\begin{pgfscope}%
\pgfsys@transformshift{0.553246in}{0.359192in}%
\pgfsys@useobject{currentmarker}{}%
\end{pgfscope}%
\begin{pgfscope}%
\pgfsys@transformshift{0.654388in}{0.361890in}%
\pgfsys@useobject{currentmarker}{}%
\end{pgfscope}%
\begin{pgfscope}%
\pgfsys@transformshift{0.780814in}{0.373892in}%
\pgfsys@useobject{currentmarker}{}%
\end{pgfscope}%
\begin{pgfscope}%
\pgfsys@transformshift{1.033668in}{0.390885in}%
\pgfsys@useobject{currentmarker}{}%
\end{pgfscope}%
\begin{pgfscope}%
\pgfsys@transformshift{1.792229in}{0.436313in}%
\pgfsys@useobject{currentmarker}{}%
\end{pgfscope}%
\begin{pgfscope}%
\pgfsys@transformshift{3.056497in}{0.483620in}%
\pgfsys@useobject{currentmarker}{}%
\end{pgfscope}%
\end{pgfscope}%
\begin{pgfscope}%
\pgfpathrectangle{\pgfqpoint{0.428083in}{0.355956in}}{\pgfqpoint{2.753577in}{1.407952in}}%
\pgfusepath{clip}%
\pgfsetrectcap%
\pgfsetroundjoin%
\pgfsetlinewidth{1.505625pt}%
\definecolor{currentstroke}{rgb}{0.349020,0.349020,0.349020}%
\pgfsetstrokecolor{currentstroke}%
\pgfsetstrokeopacity{0.600000}%
\pgfsetdash{}{0pt}%
\pgfpathmoveto{\pgfqpoint{0.553246in}{0.564964in}}%
\pgfpathlineto{\pgfqpoint{0.654388in}{0.772029in}}%
\pgfpathlineto{\pgfqpoint{0.780814in}{0.889880in}}%
\pgfpathlineto{\pgfqpoint{1.033668in}{0.994704in}}%
\pgfpathlineto{\pgfqpoint{1.792229in}{1.133059in}}%
\pgfpathlineto{\pgfqpoint{3.056497in}{1.284381in}}%
\pgfusepath{stroke}%
\end{pgfscope}%
\begin{pgfscope}%
\pgfpathrectangle{\pgfqpoint{0.428083in}{0.355956in}}{\pgfqpoint{2.753577in}{1.407952in}}%
\pgfusepath{clip}%
\pgfsetbuttcap%
\pgfsetroundjoin%
\definecolor{currentfill}{rgb}{0.349020,0.349020,0.349020}%
\pgfsetfillcolor{currentfill}%
\pgfsetfillopacity{0.600000}%
\pgfsetlinewidth{0.752812pt}%
\definecolor{currentstroke}{rgb}{1.000000,1.000000,1.000000}%
\pgfsetstrokecolor{currentstroke}%
\pgfsetstrokeopacity{0.600000}%
\pgfsetdash{}{0pt}%
\pgfsys@defobject{currentmarker}{\pgfqpoint{-0.041667in}{-0.041667in}}{\pgfqpoint{0.041667in}{0.041667in}}{%
\pgfpathmoveto{\pgfqpoint{0.000000in}{-0.041667in}}%
\pgfpathcurveto{\pgfqpoint{0.011050in}{-0.041667in}}{\pgfqpoint{0.021649in}{-0.037276in}}{\pgfqpoint{0.029463in}{-0.029463in}}%
\pgfpathcurveto{\pgfqpoint{0.037276in}{-0.021649in}}{\pgfqpoint{0.041667in}{-0.011050in}}{\pgfqpoint{0.041667in}{0.000000in}}%
\pgfpathcurveto{\pgfqpoint{0.041667in}{0.011050in}}{\pgfqpoint{0.037276in}{0.021649in}}{\pgfqpoint{0.029463in}{0.029463in}}%
\pgfpathcurveto{\pgfqpoint{0.021649in}{0.037276in}}{\pgfqpoint{0.011050in}{0.041667in}}{\pgfqpoint{0.000000in}{0.041667in}}%
\pgfpathcurveto{\pgfqpoint{-0.011050in}{0.041667in}}{\pgfqpoint{-0.021649in}{0.037276in}}{\pgfqpoint{-0.029463in}{0.029463in}}%
\pgfpathcurveto{\pgfqpoint{-0.037276in}{0.021649in}}{\pgfqpoint{-0.041667in}{0.011050in}}{\pgfqpoint{-0.041667in}{0.000000in}}%
\pgfpathcurveto{\pgfqpoint{-0.041667in}{-0.011050in}}{\pgfqpoint{-0.037276in}{-0.021649in}}{\pgfqpoint{-0.029463in}{-0.029463in}}%
\pgfpathcurveto{\pgfqpoint{-0.021649in}{-0.037276in}}{\pgfqpoint{-0.011050in}{-0.041667in}}{\pgfqpoint{0.000000in}{-0.041667in}}%
\pgfpathlineto{\pgfqpoint{0.000000in}{-0.041667in}}%
\pgfpathclose%
\pgfusepath{stroke,fill}%
}%
\begin{pgfscope}%
\pgfsys@transformshift{0.553246in}{0.564964in}%
\pgfsys@useobject{currentmarker}{}%
\end{pgfscope}%
\begin{pgfscope}%
\pgfsys@transformshift{0.654388in}{0.772029in}%
\pgfsys@useobject{currentmarker}{}%
\end{pgfscope}%
\begin{pgfscope}%
\pgfsys@transformshift{0.780814in}{0.889880in}%
\pgfsys@useobject{currentmarker}{}%
\end{pgfscope}%
\begin{pgfscope}%
\pgfsys@transformshift{1.033668in}{0.994704in}%
\pgfsys@useobject{currentmarker}{}%
\end{pgfscope}%
\begin{pgfscope}%
\pgfsys@transformshift{1.792229in}{1.133059in}%
\pgfsys@useobject{currentmarker}{}%
\end{pgfscope}%
\begin{pgfscope}%
\pgfsys@transformshift{3.056497in}{1.284381in}%
\pgfsys@useobject{currentmarker}{}%
\end{pgfscope}%
\end{pgfscope}%
\begin{pgfscope}%
\pgfpathrectangle{\pgfqpoint{0.428083in}{0.355956in}}{\pgfqpoint{2.753577in}{1.407952in}}%
\pgfusepath{clip}%
\pgfsetrectcap%
\pgfsetroundjoin%
\pgfsetlinewidth{1.505625pt}%
\definecolor{currentstroke}{rgb}{0.294118,0.000000,0.431373}%
\pgfsetstrokecolor{currentstroke}%
\pgfsetstrokeopacity{0.600000}%
\pgfsetdash{}{0pt}%
\pgfpathmoveto{\pgfqpoint{0.553246in}{0.625604in}}%
\pgfpathlineto{\pgfqpoint{0.654388in}{0.916436in}}%
\pgfpathlineto{\pgfqpoint{0.780814in}{1.041954in}}%
\pgfpathlineto{\pgfqpoint{1.033668in}{1.161334in}}%
\pgfpathlineto{\pgfqpoint{1.792229in}{1.330665in}}%
\pgfpathlineto{\pgfqpoint{3.056497in}{1.443336in}}%
\pgfusepath{stroke}%
\end{pgfscope}%
\begin{pgfscope}%
\pgfpathrectangle{\pgfqpoint{0.428083in}{0.355956in}}{\pgfqpoint{2.753577in}{1.407952in}}%
\pgfusepath{clip}%
\pgfsetbuttcap%
\pgfsetroundjoin%
\definecolor{currentfill}{rgb}{0.294118,0.000000,0.431373}%
\pgfsetfillcolor{currentfill}%
\pgfsetfillopacity{0.600000}%
\pgfsetlinewidth{0.752812pt}%
\definecolor{currentstroke}{rgb}{1.000000,1.000000,1.000000}%
\pgfsetstrokecolor{currentstroke}%
\pgfsetstrokeopacity{0.600000}%
\pgfsetdash{}{0pt}%
\pgfsys@defobject{currentmarker}{\pgfqpoint{-0.041667in}{-0.041667in}}{\pgfqpoint{0.041667in}{0.041667in}}{%
\pgfpathmoveto{\pgfqpoint{0.000000in}{-0.041667in}}%
\pgfpathcurveto{\pgfqpoint{0.011050in}{-0.041667in}}{\pgfqpoint{0.021649in}{-0.037276in}}{\pgfqpoint{0.029463in}{-0.029463in}}%
\pgfpathcurveto{\pgfqpoint{0.037276in}{-0.021649in}}{\pgfqpoint{0.041667in}{-0.011050in}}{\pgfqpoint{0.041667in}{0.000000in}}%
\pgfpathcurveto{\pgfqpoint{0.041667in}{0.011050in}}{\pgfqpoint{0.037276in}{0.021649in}}{\pgfqpoint{0.029463in}{0.029463in}}%
\pgfpathcurveto{\pgfqpoint{0.021649in}{0.037276in}}{\pgfqpoint{0.011050in}{0.041667in}}{\pgfqpoint{0.000000in}{0.041667in}}%
\pgfpathcurveto{\pgfqpoint{-0.011050in}{0.041667in}}{\pgfqpoint{-0.021649in}{0.037276in}}{\pgfqpoint{-0.029463in}{0.029463in}}%
\pgfpathcurveto{\pgfqpoint{-0.037276in}{0.021649in}}{\pgfqpoint{-0.041667in}{0.011050in}}{\pgfqpoint{-0.041667in}{0.000000in}}%
\pgfpathcurveto{\pgfqpoint{-0.041667in}{-0.011050in}}{\pgfqpoint{-0.037276in}{-0.021649in}}{\pgfqpoint{-0.029463in}{-0.029463in}}%
\pgfpathcurveto{\pgfqpoint{-0.021649in}{-0.037276in}}{\pgfqpoint{-0.011050in}{-0.041667in}}{\pgfqpoint{0.000000in}{-0.041667in}}%
\pgfpathlineto{\pgfqpoint{0.000000in}{-0.041667in}}%
\pgfpathclose%
\pgfusepath{stroke,fill}%
}%
\begin{pgfscope}%
\pgfsys@transformshift{0.553246in}{0.625604in}%
\pgfsys@useobject{currentmarker}{}%
\end{pgfscope}%
\begin{pgfscope}%
\pgfsys@transformshift{0.654388in}{0.916436in}%
\pgfsys@useobject{currentmarker}{}%
\end{pgfscope}%
\begin{pgfscope}%
\pgfsys@transformshift{0.780814in}{1.041954in}%
\pgfsys@useobject{currentmarker}{}%
\end{pgfscope}%
\begin{pgfscope}%
\pgfsys@transformshift{1.033668in}{1.161334in}%
\pgfsys@useobject{currentmarker}{}%
\end{pgfscope}%
\begin{pgfscope}%
\pgfsys@transformshift{1.792229in}{1.330665in}%
\pgfsys@useobject{currentmarker}{}%
\end{pgfscope}%
\begin{pgfscope}%
\pgfsys@transformshift{3.056497in}{1.443336in}%
\pgfsys@useobject{currentmarker}{}%
\end{pgfscope}%
\end{pgfscope}%
\begin{pgfscope}%
\pgfpathrectangle{\pgfqpoint{0.428083in}{0.355956in}}{\pgfqpoint{2.753577in}{1.407952in}}%
\pgfusepath{clip}%
\pgfsetrectcap%
\pgfsetroundjoin%
\pgfsetlinewidth{1.505625pt}%
\definecolor{currentstroke}{rgb}{0.800000,0.000000,0.000000}%
\pgfsetstrokecolor{currentstroke}%
\pgfsetstrokeopacity{0.600000}%
\pgfsetdash{}{0pt}%
\pgfpathmoveto{\pgfqpoint{0.553246in}{0.595698in}}%
\pgfpathlineto{\pgfqpoint{0.654388in}{0.841516in}}%
\pgfpathlineto{\pgfqpoint{0.780814in}{0.992507in}}%
\pgfpathlineto{\pgfqpoint{1.033668in}{1.160122in}}%
\pgfpathlineto{\pgfqpoint{1.792229in}{1.338867in}}%
\pgfpathlineto{\pgfqpoint{3.056497in}{1.424816in}}%
\pgfusepath{stroke}%
\end{pgfscope}%
\begin{pgfscope}%
\pgfpathrectangle{\pgfqpoint{0.428083in}{0.355956in}}{\pgfqpoint{2.753577in}{1.407952in}}%
\pgfusepath{clip}%
\pgfsetbuttcap%
\pgfsetroundjoin%
\definecolor{currentfill}{rgb}{0.800000,0.000000,0.000000}%
\pgfsetfillcolor{currentfill}%
\pgfsetfillopacity{0.600000}%
\pgfsetlinewidth{0.752812pt}%
\definecolor{currentstroke}{rgb}{1.000000,1.000000,1.000000}%
\pgfsetstrokecolor{currentstroke}%
\pgfsetstrokeopacity{0.600000}%
\pgfsetdash{}{0pt}%
\pgfsys@defobject{currentmarker}{\pgfqpoint{-0.041667in}{-0.041667in}}{\pgfqpoint{0.041667in}{0.041667in}}{%
\pgfpathmoveto{\pgfqpoint{0.000000in}{-0.041667in}}%
\pgfpathcurveto{\pgfqpoint{0.011050in}{-0.041667in}}{\pgfqpoint{0.021649in}{-0.037276in}}{\pgfqpoint{0.029463in}{-0.029463in}}%
\pgfpathcurveto{\pgfqpoint{0.037276in}{-0.021649in}}{\pgfqpoint{0.041667in}{-0.011050in}}{\pgfqpoint{0.041667in}{0.000000in}}%
\pgfpathcurveto{\pgfqpoint{0.041667in}{0.011050in}}{\pgfqpoint{0.037276in}{0.021649in}}{\pgfqpoint{0.029463in}{0.029463in}}%
\pgfpathcurveto{\pgfqpoint{0.021649in}{0.037276in}}{\pgfqpoint{0.011050in}{0.041667in}}{\pgfqpoint{0.000000in}{0.041667in}}%
\pgfpathcurveto{\pgfqpoint{-0.011050in}{0.041667in}}{\pgfqpoint{-0.021649in}{0.037276in}}{\pgfqpoint{-0.029463in}{0.029463in}}%
\pgfpathcurveto{\pgfqpoint{-0.037276in}{0.021649in}}{\pgfqpoint{-0.041667in}{0.011050in}}{\pgfqpoint{-0.041667in}{0.000000in}}%
\pgfpathcurveto{\pgfqpoint{-0.041667in}{-0.011050in}}{\pgfqpoint{-0.037276in}{-0.021649in}}{\pgfqpoint{-0.029463in}{-0.029463in}}%
\pgfpathcurveto{\pgfqpoint{-0.021649in}{-0.037276in}}{\pgfqpoint{-0.011050in}{-0.041667in}}{\pgfqpoint{0.000000in}{-0.041667in}}%
\pgfpathlineto{\pgfqpoint{0.000000in}{-0.041667in}}%
\pgfpathclose%
\pgfusepath{stroke,fill}%
}%
\begin{pgfscope}%
\pgfsys@transformshift{0.553246in}{0.595698in}%
\pgfsys@useobject{currentmarker}{}%
\end{pgfscope}%
\begin{pgfscope}%
\pgfsys@transformshift{0.654388in}{0.841516in}%
\pgfsys@useobject{currentmarker}{}%
\end{pgfscope}%
\begin{pgfscope}%
\pgfsys@transformshift{0.780814in}{0.992507in}%
\pgfsys@useobject{currentmarker}{}%
\end{pgfscope}%
\begin{pgfscope}%
\pgfsys@transformshift{1.033668in}{1.160122in}%
\pgfsys@useobject{currentmarker}{}%
\end{pgfscope}%
\begin{pgfscope}%
\pgfsys@transformshift{1.792229in}{1.338867in}%
\pgfsys@useobject{currentmarker}{}%
\end{pgfscope}%
\begin{pgfscope}%
\pgfsys@transformshift{3.056497in}{1.424816in}%
\pgfsys@useobject{currentmarker}{}%
\end{pgfscope}%
\end{pgfscope}%
\begin{pgfscope}%
\pgfsetrectcap%
\pgfsetmiterjoin%
\pgfsetlinewidth{0.803000pt}%
\definecolor{currentstroke}{rgb}{0.000000,0.000000,0.000000}%
\pgfsetstrokecolor{currentstroke}%
\pgfsetdash{}{0pt}%
\pgfpathmoveto{\pgfqpoint{0.428083in}{0.355956in}}%
\pgfpathlineto{\pgfqpoint{0.428083in}{1.763908in}}%
\pgfusepath{stroke}%
\end{pgfscope}%
\begin{pgfscope}%
\pgfsetrectcap%
\pgfsetmiterjoin%
\pgfsetlinewidth{0.803000pt}%
\definecolor{currentstroke}{rgb}{0.000000,0.000000,0.000000}%
\pgfsetstrokecolor{currentstroke}%
\pgfsetdash{}{0pt}%
\pgfpathmoveto{\pgfqpoint{3.181660in}{0.355956in}}%
\pgfpathlineto{\pgfqpoint{3.181660in}{1.763908in}}%
\pgfusepath{stroke}%
\end{pgfscope}%
\begin{pgfscope}%
\pgfsetrectcap%
\pgfsetmiterjoin%
\pgfsetlinewidth{0.803000pt}%
\definecolor{currentstroke}{rgb}{0.000000,0.000000,0.000000}%
\pgfsetstrokecolor{currentstroke}%
\pgfsetdash{}{0pt}%
\pgfpathmoveto{\pgfqpoint{0.428083in}{0.355956in}}%
\pgfpathlineto{\pgfqpoint{3.181660in}{0.355956in}}%
\pgfusepath{stroke}%
\end{pgfscope}%
\begin{pgfscope}%
\pgfsetrectcap%
\pgfsetmiterjoin%
\pgfsetlinewidth{0.803000pt}%
\definecolor{currentstroke}{rgb}{0.000000,0.000000,0.000000}%
\pgfsetstrokecolor{currentstroke}%
\pgfsetdash{}{0pt}%
\pgfpathmoveto{\pgfqpoint{0.428083in}{1.763908in}}%
\pgfpathlineto{\pgfqpoint{3.181660in}{1.763908in}}%
\pgfusepath{stroke}%
\end{pgfscope}%
\begin{pgfscope}%
\definecolor{textcolor}{rgb}{0.000000,0.000000,0.000000}%
\pgfsetstrokecolor{textcolor}%
\pgfsetfillcolor{textcolor}%
\pgftext[x=1.804872in,y=1.847242in,,base]{\color{textcolor}{\rmfamily\fontsize{8.000000}{9.600000}\selectfont\catcode`\^=\active\def^{\ifmmode\sp\else\^{}\fi}\catcode`\%=\active\def
\end{pgfscope}%
\begin{pgfscope}%
\pgfsetbuttcap%
\pgfsetmiterjoin%
\definecolor{currentfill}{rgb}{1.000000,1.000000,1.000000}%
\pgfsetfillcolor{currentfill}%
\pgfsetfillopacity{0.800000}%
\pgfsetlinewidth{1.003750pt}%
\definecolor{currentstroke}{rgb}{0.800000,0.800000,0.800000}%
\pgfsetstrokecolor{currentstroke}%
\pgfsetstrokeopacity{0.800000}%
\pgfsetdash{}{0pt}%
\pgfpathmoveto{\pgfqpoint{0.486417in}{1.083310in}}%
\pgfpathlineto{\pgfqpoint{1.276602in}{1.083310in}}%
\pgfpathquadraticcurveto{\pgfqpoint{1.293269in}{1.083310in}}{\pgfqpoint{1.293269in}{1.099976in}}%
\pgfpathlineto{\pgfqpoint{1.293269in}{1.705575in}}%
\pgfpathquadraticcurveto{\pgfqpoint{1.293269in}{1.722242in}}{\pgfqpoint{1.276602in}{1.722242in}}%
\pgfpathlineto{\pgfqpoint{0.486417in}{1.722242in}}%
\pgfpathquadraticcurveto{\pgfqpoint{0.469750in}{1.722242in}}{\pgfqpoint{0.469750in}{1.705575in}}%
\pgfpathlineto{\pgfqpoint{0.469750in}{1.099976in}}%
\pgfpathquadraticcurveto{\pgfqpoint{0.469750in}{1.083310in}}{\pgfqpoint{0.486417in}{1.083310in}}%
\pgfpathlineto{\pgfqpoint{0.486417in}{1.083310in}}%
\pgfpathclose%
\pgfusepath{stroke,fill}%
\end{pgfscope}%
\begin{pgfscope}%
\pgfsetrectcap%
\pgfsetroundjoin%
\pgfsetlinewidth{1.505625pt}%
\definecolor{currentstroke}{rgb}{0.200000,0.400000,1.000000}%
\pgfsetstrokecolor{currentstroke}%
\pgfsetstrokeopacity{0.600000}%
\pgfsetdash{}{0pt}%
\pgfpathmoveto{\pgfqpoint{0.503083in}{1.654761in}}%
\pgfpathlineto{\pgfqpoint{0.586417in}{1.654761in}}%
\pgfpathlineto{\pgfqpoint{0.669750in}{1.654761in}}%
\pgfusepath{stroke}%
\end{pgfscope}%
\begin{pgfscope}%
\pgfsetbuttcap%
\pgfsetroundjoin%
\definecolor{currentfill}{rgb}{0.200000,0.400000,1.000000}%
\pgfsetfillcolor{currentfill}%
\pgfsetfillopacity{0.600000}%
\pgfsetlinewidth{0.752812pt}%
\definecolor{currentstroke}{rgb}{1.000000,1.000000,1.000000}%
\pgfsetstrokecolor{currentstroke}%
\pgfsetstrokeopacity{0.600000}%
\pgfsetdash{}{0pt}%
\pgfsys@defobject{currentmarker}{\pgfqpoint{-0.041667in}{-0.041667in}}{\pgfqpoint{0.041667in}{0.041667in}}{%
\pgfpathmoveto{\pgfqpoint{0.000000in}{-0.041667in}}%
\pgfpathcurveto{\pgfqpoint{0.011050in}{-0.041667in}}{\pgfqpoint{0.021649in}{-0.037276in}}{\pgfqpoint{0.029463in}{-0.029463in}}%
\pgfpathcurveto{\pgfqpoint{0.037276in}{-0.021649in}}{\pgfqpoint{0.041667in}{-0.011050in}}{\pgfqpoint{0.041667in}{0.000000in}}%
\pgfpathcurveto{\pgfqpoint{0.041667in}{0.011050in}}{\pgfqpoint{0.037276in}{0.021649in}}{\pgfqpoint{0.029463in}{0.029463in}}%
\pgfpathcurveto{\pgfqpoint{0.021649in}{0.037276in}}{\pgfqpoint{0.011050in}{0.041667in}}{\pgfqpoint{0.000000in}{0.041667in}}%
\pgfpathcurveto{\pgfqpoint{-0.011050in}{0.041667in}}{\pgfqpoint{-0.021649in}{0.037276in}}{\pgfqpoint{-0.029463in}{0.029463in}}%
\pgfpathcurveto{\pgfqpoint{-0.037276in}{0.021649in}}{\pgfqpoint{-0.041667in}{0.011050in}}{\pgfqpoint{-0.041667in}{0.000000in}}%
\pgfpathcurveto{\pgfqpoint{-0.041667in}{-0.011050in}}{\pgfqpoint{-0.037276in}{-0.021649in}}{\pgfqpoint{-0.029463in}{-0.029463in}}%
\pgfpathcurveto{\pgfqpoint{-0.021649in}{-0.037276in}}{\pgfqpoint{-0.011050in}{-0.041667in}}{\pgfqpoint{0.000000in}{-0.041667in}}%
\pgfpathlineto{\pgfqpoint{0.000000in}{-0.041667in}}%
\pgfpathclose%
\pgfusepath{stroke,fill}%
}%
\begin{pgfscope}%
\pgfsys@transformshift{0.586417in}{1.654761in}%
\pgfsys@useobject{currentmarker}{}%
\end{pgfscope}%
\end{pgfscope}%
\begin{pgfscope}%
\definecolor{textcolor}{rgb}{0.000000,0.000000,0.000000}%
\pgfsetstrokecolor{textcolor}%
\pgfsetfillcolor{textcolor}%
\pgftext[x=0.736417in,y=1.625594in,left,base]{\color{textcolor}{\rmfamily\fontsize{6.000000}{7.200000}\selectfont\catcode`\^=\active\def^{\ifmmode\sp\else\^{}\fi}\catcode`\%=\active\def
\end{pgfscope}%
\begin{pgfscope}%
\pgfsetrectcap%
\pgfsetroundjoin%
\pgfsetlinewidth{1.505625pt}%
\definecolor{currentstroke}{rgb}{0.600000,0.400000,0.200000}%
\pgfsetstrokecolor{currentstroke}%
\pgfsetstrokeopacity{0.600000}%
\pgfsetdash{}{0pt}%
\pgfpathmoveto{\pgfqpoint{0.503083in}{1.532447in}}%
\pgfpathlineto{\pgfqpoint{0.586417in}{1.532447in}}%
\pgfpathlineto{\pgfqpoint{0.669750in}{1.532447in}}%
\pgfusepath{stroke}%
\end{pgfscope}%
\begin{pgfscope}%
\pgfsetbuttcap%
\pgfsetroundjoin%
\definecolor{currentfill}{rgb}{0.600000,0.400000,0.200000}%
\pgfsetfillcolor{currentfill}%
\pgfsetfillopacity{0.600000}%
\pgfsetlinewidth{0.752812pt}%
\definecolor{currentstroke}{rgb}{1.000000,1.000000,1.000000}%
\pgfsetstrokecolor{currentstroke}%
\pgfsetstrokeopacity{0.600000}%
\pgfsetdash{}{0pt}%
\pgfsys@defobject{currentmarker}{\pgfqpoint{-0.041667in}{-0.041667in}}{\pgfqpoint{0.041667in}{0.041667in}}{%
\pgfpathmoveto{\pgfqpoint{0.000000in}{-0.041667in}}%
\pgfpathcurveto{\pgfqpoint{0.011050in}{-0.041667in}}{\pgfqpoint{0.021649in}{-0.037276in}}{\pgfqpoint{0.029463in}{-0.029463in}}%
\pgfpathcurveto{\pgfqpoint{0.037276in}{-0.021649in}}{\pgfqpoint{0.041667in}{-0.011050in}}{\pgfqpoint{0.041667in}{0.000000in}}%
\pgfpathcurveto{\pgfqpoint{0.041667in}{0.011050in}}{\pgfqpoint{0.037276in}{0.021649in}}{\pgfqpoint{0.029463in}{0.029463in}}%
\pgfpathcurveto{\pgfqpoint{0.021649in}{0.037276in}}{\pgfqpoint{0.011050in}{0.041667in}}{\pgfqpoint{0.000000in}{0.041667in}}%
\pgfpathcurveto{\pgfqpoint{-0.011050in}{0.041667in}}{\pgfqpoint{-0.021649in}{0.037276in}}{\pgfqpoint{-0.029463in}{0.029463in}}%
\pgfpathcurveto{\pgfqpoint{-0.037276in}{0.021649in}}{\pgfqpoint{-0.041667in}{0.011050in}}{\pgfqpoint{-0.041667in}{0.000000in}}%
\pgfpathcurveto{\pgfqpoint{-0.041667in}{-0.011050in}}{\pgfqpoint{-0.037276in}{-0.021649in}}{\pgfqpoint{-0.029463in}{-0.029463in}}%
\pgfpathcurveto{\pgfqpoint{-0.021649in}{-0.037276in}}{\pgfqpoint{-0.011050in}{-0.041667in}}{\pgfqpoint{0.000000in}{-0.041667in}}%
\pgfpathlineto{\pgfqpoint{0.000000in}{-0.041667in}}%
\pgfpathclose%
\pgfusepath{stroke,fill}%
}%
\begin{pgfscope}%
\pgfsys@transformshift{0.586417in}{1.532447in}%
\pgfsys@useobject{currentmarker}{}%
\end{pgfscope}%
\end{pgfscope}%
\begin{pgfscope}%
\definecolor{textcolor}{rgb}{0.000000,0.000000,0.000000}%
\pgfsetstrokecolor{textcolor}%
\pgfsetfillcolor{textcolor}%
\pgftext[x=0.736417in,y=1.503280in,left,base]{\color{textcolor}{\rmfamily\fontsize{6.000000}{7.200000}\selectfont\catcode`\^=\active\def^{\ifmmode\sp\else\^{}\fi}\catcode`\%=\active\def
\end{pgfscope}%
\begin{pgfscope}%
\pgfsetrectcap%
\pgfsetroundjoin%
\pgfsetlinewidth{1.505625pt}%
\definecolor{currentstroke}{rgb}{0.349020,0.349020,0.349020}%
\pgfsetstrokecolor{currentstroke}%
\pgfsetstrokeopacity{0.600000}%
\pgfsetdash{}{0pt}%
\pgfpathmoveto{\pgfqpoint{0.503083in}{1.410132in}}%
\pgfpathlineto{\pgfqpoint{0.586417in}{1.410132in}}%
\pgfpathlineto{\pgfqpoint{0.669750in}{1.410132in}}%
\pgfusepath{stroke}%
\end{pgfscope}%
\begin{pgfscope}%
\pgfsetbuttcap%
\pgfsetroundjoin%
\definecolor{currentfill}{rgb}{0.349020,0.349020,0.349020}%
\pgfsetfillcolor{currentfill}%
\pgfsetfillopacity{0.600000}%
\pgfsetlinewidth{0.752812pt}%
\definecolor{currentstroke}{rgb}{1.000000,1.000000,1.000000}%
\pgfsetstrokecolor{currentstroke}%
\pgfsetstrokeopacity{0.600000}%
\pgfsetdash{}{0pt}%
\pgfsys@defobject{currentmarker}{\pgfqpoint{-0.041667in}{-0.041667in}}{\pgfqpoint{0.041667in}{0.041667in}}{%
\pgfpathmoveto{\pgfqpoint{0.000000in}{-0.041667in}}%
\pgfpathcurveto{\pgfqpoint{0.011050in}{-0.041667in}}{\pgfqpoint{0.021649in}{-0.037276in}}{\pgfqpoint{0.029463in}{-0.029463in}}%
\pgfpathcurveto{\pgfqpoint{0.037276in}{-0.021649in}}{\pgfqpoint{0.041667in}{-0.011050in}}{\pgfqpoint{0.041667in}{0.000000in}}%
\pgfpathcurveto{\pgfqpoint{0.041667in}{0.011050in}}{\pgfqpoint{0.037276in}{0.021649in}}{\pgfqpoint{0.029463in}{0.029463in}}%
\pgfpathcurveto{\pgfqpoint{0.021649in}{0.037276in}}{\pgfqpoint{0.011050in}{0.041667in}}{\pgfqpoint{0.000000in}{0.041667in}}%
\pgfpathcurveto{\pgfqpoint{-0.011050in}{0.041667in}}{\pgfqpoint{-0.021649in}{0.037276in}}{\pgfqpoint{-0.029463in}{0.029463in}}%
\pgfpathcurveto{\pgfqpoint{-0.037276in}{0.021649in}}{\pgfqpoint{-0.041667in}{0.011050in}}{\pgfqpoint{-0.041667in}{0.000000in}}%
\pgfpathcurveto{\pgfqpoint{-0.041667in}{-0.011050in}}{\pgfqpoint{-0.037276in}{-0.021649in}}{\pgfqpoint{-0.029463in}{-0.029463in}}%
\pgfpathcurveto{\pgfqpoint{-0.021649in}{-0.037276in}}{\pgfqpoint{-0.011050in}{-0.041667in}}{\pgfqpoint{0.000000in}{-0.041667in}}%
\pgfpathlineto{\pgfqpoint{0.000000in}{-0.041667in}}%
\pgfpathclose%
\pgfusepath{stroke,fill}%
}%
\begin{pgfscope}%
\pgfsys@transformshift{0.586417in}{1.410132in}%
\pgfsys@useobject{currentmarker}{}%
\end{pgfscope}%
\end{pgfscope}%
\begin{pgfscope}%
\definecolor{textcolor}{rgb}{0.000000,0.000000,0.000000}%
\pgfsetstrokecolor{textcolor}%
\pgfsetfillcolor{textcolor}%
\pgftext[x=0.736417in,y=1.380966in,left,base]{\color{textcolor}{\rmfamily\fontsize{6.000000}{7.200000}\selectfont\catcode`\^=\active\def^{\ifmmode\sp\else\^{}\fi}\catcode`\%=\active\def
\end{pgfscope}%
\begin{pgfscope}%
\pgfsetrectcap%
\pgfsetroundjoin%
\pgfsetlinewidth{1.505625pt}%
\definecolor{currentstroke}{rgb}{0.294118,0.000000,0.431373}%
\pgfsetstrokecolor{currentstroke}%
\pgfsetstrokeopacity{0.600000}%
\pgfsetdash{}{0pt}%
\pgfpathmoveto{\pgfqpoint{0.503083in}{1.287818in}}%
\pgfpathlineto{\pgfqpoint{0.586417in}{1.287818in}}%
\pgfpathlineto{\pgfqpoint{0.669750in}{1.287818in}}%
\pgfusepath{stroke}%
\end{pgfscope}%
\begin{pgfscope}%
\pgfsetbuttcap%
\pgfsetroundjoin%
\definecolor{currentfill}{rgb}{0.294118,0.000000,0.431373}%
\pgfsetfillcolor{currentfill}%
\pgfsetfillopacity{0.600000}%
\pgfsetlinewidth{0.752812pt}%
\definecolor{currentstroke}{rgb}{1.000000,1.000000,1.000000}%
\pgfsetstrokecolor{currentstroke}%
\pgfsetstrokeopacity{0.600000}%
\pgfsetdash{}{0pt}%
\pgfsys@defobject{currentmarker}{\pgfqpoint{-0.041667in}{-0.041667in}}{\pgfqpoint{0.041667in}{0.041667in}}{%
\pgfpathmoveto{\pgfqpoint{0.000000in}{-0.041667in}}%
\pgfpathcurveto{\pgfqpoint{0.011050in}{-0.041667in}}{\pgfqpoint{0.021649in}{-0.037276in}}{\pgfqpoint{0.029463in}{-0.029463in}}%
\pgfpathcurveto{\pgfqpoint{0.037276in}{-0.021649in}}{\pgfqpoint{0.041667in}{-0.011050in}}{\pgfqpoint{0.041667in}{0.000000in}}%
\pgfpathcurveto{\pgfqpoint{0.041667in}{0.011050in}}{\pgfqpoint{0.037276in}{0.021649in}}{\pgfqpoint{0.029463in}{0.029463in}}%
\pgfpathcurveto{\pgfqpoint{0.021649in}{0.037276in}}{\pgfqpoint{0.011050in}{0.041667in}}{\pgfqpoint{0.000000in}{0.041667in}}%
\pgfpathcurveto{\pgfqpoint{-0.011050in}{0.041667in}}{\pgfqpoint{-0.021649in}{0.037276in}}{\pgfqpoint{-0.029463in}{0.029463in}}%
\pgfpathcurveto{\pgfqpoint{-0.037276in}{0.021649in}}{\pgfqpoint{-0.041667in}{0.011050in}}{\pgfqpoint{-0.041667in}{0.000000in}}%
\pgfpathcurveto{\pgfqpoint{-0.041667in}{-0.011050in}}{\pgfqpoint{-0.037276in}{-0.021649in}}{\pgfqpoint{-0.029463in}{-0.029463in}}%
\pgfpathcurveto{\pgfqpoint{-0.021649in}{-0.037276in}}{\pgfqpoint{-0.011050in}{-0.041667in}}{\pgfqpoint{0.000000in}{-0.041667in}}%
\pgfpathlineto{\pgfqpoint{0.000000in}{-0.041667in}}%
\pgfpathclose%
\pgfusepath{stroke,fill}%
}%
\begin{pgfscope}%
\pgfsys@transformshift{0.586417in}{1.287818in}%
\pgfsys@useobject{currentmarker}{}%
\end{pgfscope}%
\end{pgfscope}%
\begin{pgfscope}%
\definecolor{textcolor}{rgb}{0.000000,0.000000,0.000000}%
\pgfsetstrokecolor{textcolor}%
\pgfsetfillcolor{textcolor}%
\pgftext[x=0.736417in,y=1.258651in,left,base]{\color{textcolor}{\rmfamily\fontsize{6.000000}{7.200000}\selectfont\catcode`\^=\active\def^{\ifmmode\sp\else\^{}\fi}\catcode`\%=\active\def
\end{pgfscope}%
\begin{pgfscope}%
\pgfsetrectcap%
\pgfsetroundjoin%
\pgfsetlinewidth{1.505625pt}%
\definecolor{currentstroke}{rgb}{0.800000,0.000000,0.000000}%
\pgfsetstrokecolor{currentstroke}%
\pgfsetstrokeopacity{0.600000}%
\pgfsetdash{}{0pt}%
\pgfpathmoveto{\pgfqpoint{0.503083in}{1.164324in}}%
\pgfpathlineto{\pgfqpoint{0.586417in}{1.164324in}}%
\pgfpathlineto{\pgfqpoint{0.669750in}{1.164324in}}%
\pgfusepath{stroke}%
\end{pgfscope}%
\begin{pgfscope}%
\pgfsetbuttcap%
\pgfsetroundjoin%
\definecolor{currentfill}{rgb}{0.800000,0.000000,0.000000}%
\pgfsetfillcolor{currentfill}%
\pgfsetfillopacity{0.600000}%
\pgfsetlinewidth{0.752812pt}%
\definecolor{currentstroke}{rgb}{1.000000,1.000000,1.000000}%
\pgfsetstrokecolor{currentstroke}%
\pgfsetstrokeopacity{0.600000}%
\pgfsetdash{}{0pt}%
\pgfsys@defobject{currentmarker}{\pgfqpoint{-0.041667in}{-0.041667in}}{\pgfqpoint{0.041667in}{0.041667in}}{%
\pgfpathmoveto{\pgfqpoint{0.000000in}{-0.041667in}}%
\pgfpathcurveto{\pgfqpoint{0.011050in}{-0.041667in}}{\pgfqpoint{0.021649in}{-0.037276in}}{\pgfqpoint{0.029463in}{-0.029463in}}%
\pgfpathcurveto{\pgfqpoint{0.037276in}{-0.021649in}}{\pgfqpoint{0.041667in}{-0.011050in}}{\pgfqpoint{0.041667in}{0.000000in}}%
\pgfpathcurveto{\pgfqpoint{0.041667in}{0.011050in}}{\pgfqpoint{0.037276in}{0.021649in}}{\pgfqpoint{0.029463in}{0.029463in}}%
\pgfpathcurveto{\pgfqpoint{0.021649in}{0.037276in}}{\pgfqpoint{0.011050in}{0.041667in}}{\pgfqpoint{0.000000in}{0.041667in}}%
\pgfpathcurveto{\pgfqpoint{-0.011050in}{0.041667in}}{\pgfqpoint{-0.021649in}{0.037276in}}{\pgfqpoint{-0.029463in}{0.029463in}}%
\pgfpathcurveto{\pgfqpoint{-0.037276in}{0.021649in}}{\pgfqpoint{-0.041667in}{0.011050in}}{\pgfqpoint{-0.041667in}{0.000000in}}%
\pgfpathcurveto{\pgfqpoint{-0.041667in}{-0.011050in}}{\pgfqpoint{-0.037276in}{-0.021649in}}{\pgfqpoint{-0.029463in}{-0.029463in}}%
\pgfpathcurveto{\pgfqpoint{-0.021649in}{-0.037276in}}{\pgfqpoint{-0.011050in}{-0.041667in}}{\pgfqpoint{0.000000in}{-0.041667in}}%
\pgfpathlineto{\pgfqpoint{0.000000in}{-0.041667in}}%
\pgfpathclose%
\pgfusepath{stroke,fill}%
}%
\begin{pgfscope}%
\pgfsys@transformshift{0.586417in}{1.164324in}%
\pgfsys@useobject{currentmarker}{}%
\end{pgfscope}%
\end{pgfscope}%
\begin{pgfscope}%
\definecolor{textcolor}{rgb}{0.000000,0.000000,0.000000}%
\pgfsetstrokecolor{textcolor}%
\pgfsetfillcolor{textcolor}%
\pgftext[x=0.736417in,y=1.135157in,left,base]{\color{textcolor}{\rmfamily\fontsize{6.000000}{7.200000}\selectfont\catcode`\^=\active\def^{\ifmmode\sp\else\^{}\fi}\catcode`\%=\active\def
\end{pgfscope}%
\end{pgfpicture}%
\makeatother%
\endgroup%

%% file: SWE-Bench-Verified-File_recall_graph.pgf
\begingroup%
\makeatletter%
\begin{pgfpicture}%
\pgfpathrectangle{\pgfpointorigin}{\pgfqpoint{3.196660in}{1.946660in}}%
\pgfusepath{use as bounding box, clip}%
\begin{pgfscope}%
\pgfsetbuttcap%
\pgfsetmiterjoin%
\definecolor{currentfill}{rgb}{1.000000,1.000000,1.000000}%
\pgfsetfillcolor{currentfill}%
\pgfsetlinewidth{0.000000pt}%
\definecolor{currentstroke}{rgb}{1.000000,1.000000,1.000000}%
\pgfsetstrokecolor{currentstroke}%
\pgfsetdash{}{0pt}%
\pgfpathmoveto{\pgfqpoint{0.000000in}{0.000000in}}%
\pgfpathlineto{\pgfqpoint{3.196660in}{0.000000in}}%
\pgfpathlineto{\pgfqpoint{3.196660in}{1.946660in}}%
\pgfpathlineto{\pgfqpoint{0.000000in}{1.946660in}}%
\pgfpathlineto{\pgfqpoint{0.000000in}{0.000000in}}%
\pgfpathclose%
\pgfusepath{fill}%
\end{pgfscope}%
\begin{pgfscope}%
\pgfsetbuttcap%
\pgfsetmiterjoin%
\definecolor{currentfill}{rgb}{1.000000,1.000000,1.000000}%
\pgfsetfillcolor{currentfill}%
\pgfsetlinewidth{0.000000pt}%
\definecolor{currentstroke}{rgb}{0.000000,0.000000,0.000000}%
\pgfsetstrokecolor{currentstroke}%
\pgfsetstrokeopacity{0.000000}%
\pgfsetdash{}{0pt}%
\pgfpathmoveto{\pgfqpoint{0.428083in}{0.355956in}}%
\pgfpathlineto{\pgfqpoint{3.181660in}{0.355956in}}%
\pgfpathlineto{\pgfqpoint{3.181660in}{1.763908in}}%
\pgfpathlineto{\pgfqpoint{0.428083in}{1.763908in}}%
\pgfpathlineto{\pgfqpoint{0.428083in}{0.355956in}}%
\pgfpathclose%
\pgfusepath{fill}%
\end{pgfscope}%
\begin{pgfscope}%
\pgfsetbuttcap%
\pgfsetroundjoin%
\definecolor{currentfill}{rgb}{0.000000,0.000000,0.000000}%
\pgfsetfillcolor{currentfill}%
\pgfsetlinewidth{0.803000pt}%
\definecolor{currentstroke}{rgb}{0.000000,0.000000,0.000000}%
\pgfsetstrokecolor{currentstroke}%
\pgfsetdash{}{0pt}%
\pgfsys@defobject{currentmarker}{\pgfqpoint{0.000000in}{-0.048611in}}{\pgfqpoint{0.000000in}{0.000000in}}{%
\pgfpathmoveto{\pgfqpoint{0.000000in}{0.000000in}}%
\pgfpathlineto{\pgfqpoint{0.000000in}{-0.048611in}}%
\pgfusepath{stroke,fill}%
}%
\begin{pgfscope}%
\pgfsys@transformshift{0.527961in}{0.355956in}%
\pgfsys@useobject{currentmarker}{}%
\end{pgfscope}%
\end{pgfscope}%
\begin{pgfscope}%
\definecolor{textcolor}{rgb}{0.000000,0.000000,0.000000}%
\pgfsetstrokecolor{textcolor}%
\pgfsetfillcolor{textcolor}%
\pgftext[x=0.527961in,y=0.258733in,,top]{\color{textcolor}{\rmfamily\fontsize{6.000000}{7.200000}\selectfont\catcode`\^=\active\def^{\ifmmode\sp\else\^{}\fi}\catcode`\%=\active\def
\end{pgfscope}%
\begin{pgfscope}%
\pgfsetbuttcap%
\pgfsetroundjoin%
\definecolor{currentfill}{rgb}{0.000000,0.000000,0.000000}%
\pgfsetfillcolor{currentfill}%
\pgfsetlinewidth{0.803000pt}%
\definecolor{currentstroke}{rgb}{0.000000,0.000000,0.000000}%
\pgfsetstrokecolor{currentstroke}%
\pgfsetdash{}{0pt}%
\pgfsys@defobject{currentmarker}{\pgfqpoint{0.000000in}{-0.048611in}}{\pgfqpoint{0.000000in}{0.000000in}}{%
\pgfpathmoveto{\pgfqpoint{0.000000in}{0.000000in}}%
\pgfpathlineto{\pgfqpoint{0.000000in}{-0.048611in}}%
\pgfusepath{stroke,fill}%
}%
\begin{pgfscope}%
\pgfsys@transformshift{1.033668in}{0.355956in}%
\pgfsys@useobject{currentmarker}{}%
\end{pgfscope}%
\end{pgfscope}%
\begin{pgfscope}%
\definecolor{textcolor}{rgb}{0.000000,0.000000,0.000000}%
\pgfsetstrokecolor{textcolor}%
\pgfsetfillcolor{textcolor}%
\pgftext[x=1.033668in,y=0.258733in,,top]{\color{textcolor}{\rmfamily\fontsize{6.000000}{7.200000}\selectfont\catcode`\^=\active\def^{\ifmmode\sp\else\^{}\fi}\catcode`\%=\active\def
\end{pgfscope}%
\begin{pgfscope}%
\pgfsetbuttcap%
\pgfsetroundjoin%
\definecolor{currentfill}{rgb}{0.000000,0.000000,0.000000}%
\pgfsetfillcolor{currentfill}%
\pgfsetlinewidth{0.803000pt}%
\definecolor{currentstroke}{rgb}{0.000000,0.000000,0.000000}%
\pgfsetstrokecolor{currentstroke}%
\pgfsetdash{}{0pt}%
\pgfsys@defobject{currentmarker}{\pgfqpoint{0.000000in}{-0.048611in}}{\pgfqpoint{0.000000in}{0.000000in}}{%
\pgfpathmoveto{\pgfqpoint{0.000000in}{0.000000in}}%
\pgfpathlineto{\pgfqpoint{0.000000in}{-0.048611in}}%
\pgfusepath{stroke,fill}%
}%
\begin{pgfscope}%
\pgfsys@transformshift{1.539375in}{0.355956in}%
\pgfsys@useobject{currentmarker}{}%
\end{pgfscope}%
\end{pgfscope}%
\begin{pgfscope}%
\definecolor{textcolor}{rgb}{0.000000,0.000000,0.000000}%
\pgfsetstrokecolor{textcolor}%
\pgfsetfillcolor{textcolor}%
\pgftext[x=1.539375in,y=0.258733in,,top]{\color{textcolor}{\rmfamily\fontsize{6.000000}{7.200000}\selectfont\catcode`\^=\active\def^{\ifmmode\sp\else\^{}\fi}\catcode`\%=\active\def
\end{pgfscope}%
\begin{pgfscope}%
\pgfsetbuttcap%
\pgfsetroundjoin%
\definecolor{currentfill}{rgb}{0.000000,0.000000,0.000000}%
\pgfsetfillcolor{currentfill}%
\pgfsetlinewidth{0.803000pt}%
\definecolor{currentstroke}{rgb}{0.000000,0.000000,0.000000}%
\pgfsetstrokecolor{currentstroke}%
\pgfsetdash{}{0pt}%
\pgfsys@defobject{currentmarker}{\pgfqpoint{0.000000in}{-0.048611in}}{\pgfqpoint{0.000000in}{0.000000in}}{%
\pgfpathmoveto{\pgfqpoint{0.000000in}{0.000000in}}%
\pgfpathlineto{\pgfqpoint{0.000000in}{-0.048611in}}%
\pgfusepath{stroke,fill}%
}%
\begin{pgfscope}%
\pgfsys@transformshift{2.045083in}{0.355956in}%
\pgfsys@useobject{currentmarker}{}%
\end{pgfscope}%
\end{pgfscope}%
\begin{pgfscope}%
\definecolor{textcolor}{rgb}{0.000000,0.000000,0.000000}%
\pgfsetstrokecolor{textcolor}%
\pgfsetfillcolor{textcolor}%
\pgftext[x=2.045083in,y=0.258733in,,top]{\color{textcolor}{\rmfamily\fontsize{6.000000}{7.200000}\selectfont\catcode`\^=\active\def^{\ifmmode\sp\else\^{}\fi}\catcode`\%=\active\def
\end{pgfscope}%
\begin{pgfscope}%
\pgfsetbuttcap%
\pgfsetroundjoin%
\definecolor{currentfill}{rgb}{0.000000,0.000000,0.000000}%
\pgfsetfillcolor{currentfill}%
\pgfsetlinewidth{0.803000pt}%
\definecolor{currentstroke}{rgb}{0.000000,0.000000,0.000000}%
\pgfsetstrokecolor{currentstroke}%
\pgfsetdash{}{0pt}%
\pgfsys@defobject{currentmarker}{\pgfqpoint{0.000000in}{-0.048611in}}{\pgfqpoint{0.000000in}{0.000000in}}{%
\pgfpathmoveto{\pgfqpoint{0.000000in}{0.000000in}}%
\pgfpathlineto{\pgfqpoint{0.000000in}{-0.048611in}}%
\pgfusepath{stroke,fill}%
}%
\begin{pgfscope}%
\pgfsys@transformshift{2.550790in}{0.355956in}%
\pgfsys@useobject{currentmarker}{}%
\end{pgfscope}%
\end{pgfscope}%
\begin{pgfscope}%
\definecolor{textcolor}{rgb}{0.000000,0.000000,0.000000}%
\pgfsetstrokecolor{textcolor}%
\pgfsetfillcolor{textcolor}%
\pgftext[x=2.550790in,y=0.258733in,,top]{\color{textcolor}{\rmfamily\fontsize{6.000000}{7.200000}\selectfont\catcode`\^=\active\def^{\ifmmode\sp\else\^{}\fi}\catcode`\%=\active\def
\end{pgfscope}%
\begin{pgfscope}%
\pgfsetbuttcap%
\pgfsetroundjoin%
\definecolor{currentfill}{rgb}{0.000000,0.000000,0.000000}%
\pgfsetfillcolor{currentfill}%
\pgfsetlinewidth{0.803000pt}%
\definecolor{currentstroke}{rgb}{0.000000,0.000000,0.000000}%
\pgfsetstrokecolor{currentstroke}%
\pgfsetdash{}{0pt}%
\pgfsys@defobject{currentmarker}{\pgfqpoint{0.000000in}{-0.048611in}}{\pgfqpoint{0.000000in}{0.000000in}}{%
\pgfpathmoveto{\pgfqpoint{0.000000in}{0.000000in}}%
\pgfpathlineto{\pgfqpoint{0.000000in}{-0.048611in}}%
\pgfusepath{stroke,fill}%
}%
\begin{pgfscope}%
\pgfsys@transformshift{3.056497in}{0.355956in}%
\pgfsys@useobject{currentmarker}{}%
\end{pgfscope}%
\end{pgfscope}%
\begin{pgfscope}%
\definecolor{textcolor}{rgb}{0.000000,0.000000,0.000000}%
\pgfsetstrokecolor{textcolor}%
\pgfsetfillcolor{textcolor}%
\pgftext[x=3.056497in,y=0.258733in,,top]{\color{textcolor}{\rmfamily\fontsize{6.000000}{7.200000}\selectfont\catcode`\^=\active\def^{\ifmmode\sp\else\^{}\fi}\catcode`\%=\active\def
\end{pgfscope}%
\begin{pgfscope}%
\definecolor{textcolor}{rgb}{0.000000,0.000000,0.000000}%
\pgfsetstrokecolor{textcolor}%
\pgfsetfillcolor{textcolor}%
\pgftext[x=1.804872in,y=0.122530in,,top]{\color{textcolor}{\rmfamily\fontsize{8.000000}{9.600000}\selectfont\catcode`\^=\active\def^{\ifmmode\sp\else\^{}\fi}\catcode`\%=\active\def
\end{pgfscope}%
\begin{pgfscope}%
\pgfsetbuttcap%
\pgfsetroundjoin%
\definecolor{currentfill}{rgb}{0.000000,0.000000,0.000000}%
\pgfsetfillcolor{currentfill}%
\pgfsetlinewidth{0.803000pt}%
\definecolor{currentstroke}{rgb}{0.000000,0.000000,0.000000}%
\pgfsetstrokecolor{currentstroke}%
\pgfsetdash{}{0pt}%
\pgfsys@defobject{currentmarker}{\pgfqpoint{-0.048611in}{0.000000in}}{\pgfqpoint{-0.000000in}{0.000000in}}{%
\pgfpathmoveto{\pgfqpoint{-0.000000in}{0.000000in}}%
\pgfpathlineto{\pgfqpoint{-0.048611in}{0.000000in}}%
\pgfusepath{stroke,fill}%
}%
\begin{pgfscope}%
\pgfsys@transformshift{0.428083in}{0.355956in}%
\pgfsys@useobject{currentmarker}{}%
\end{pgfscope}%
\end{pgfscope}%
\begin{pgfscope}%
\definecolor{textcolor}{rgb}{0.000000,0.000000,0.000000}%
\pgfsetstrokecolor{textcolor}%
\pgfsetfillcolor{textcolor}%
\pgftext[x=0.279936in, y=0.324299in, left, base]{\color{textcolor}{\rmfamily\fontsize{6.000000}{7.200000}\selectfont\catcode`\^=\active\def^{\ifmmode\sp\else\^{}\fi}\catcode`\%=\active\def
\end{pgfscope}%
\begin{pgfscope}%
\pgfsetbuttcap%
\pgfsetroundjoin%
\definecolor{currentfill}{rgb}{0.000000,0.000000,0.000000}%
\pgfsetfillcolor{currentfill}%
\pgfsetlinewidth{0.803000pt}%
\definecolor{currentstroke}{rgb}{0.000000,0.000000,0.000000}%
\pgfsetstrokecolor{currentstroke}%
\pgfsetdash{}{0pt}%
\pgfsys@defobject{currentmarker}{\pgfqpoint{-0.048611in}{0.000000in}}{\pgfqpoint{-0.000000in}{0.000000in}}{%
\pgfpathmoveto{\pgfqpoint{-0.000000in}{0.000000in}}%
\pgfpathlineto{\pgfqpoint{-0.048611in}{0.000000in}}%
\pgfusepath{stroke,fill}%
}%
\begin{pgfscope}%
\pgfsys@transformshift{0.428083in}{0.637546in}%
\pgfsys@useobject{currentmarker}{}%
\end{pgfscope}%
\end{pgfscope}%
\begin{pgfscope}%
\definecolor{textcolor}{rgb}{0.000000,0.000000,0.000000}%
\pgfsetstrokecolor{textcolor}%
\pgfsetfillcolor{textcolor}%
\pgftext[x=0.229011in, y=0.605889in, left, base]{\color{textcolor}{\rmfamily\fontsize{6.000000}{7.200000}\selectfont\catcode`\^=\active\def^{\ifmmode\sp\else\^{}\fi}\catcode`\%=\active\def
\end{pgfscope}%
\begin{pgfscope}%
\pgfsetbuttcap%
\pgfsetroundjoin%
\definecolor{currentfill}{rgb}{0.000000,0.000000,0.000000}%
\pgfsetfillcolor{currentfill}%
\pgfsetlinewidth{0.803000pt}%
\definecolor{currentstroke}{rgb}{0.000000,0.000000,0.000000}%
\pgfsetstrokecolor{currentstroke}%
\pgfsetdash{}{0pt}%
\pgfsys@defobject{currentmarker}{\pgfqpoint{-0.048611in}{0.000000in}}{\pgfqpoint{-0.000000in}{0.000000in}}{%
\pgfpathmoveto{\pgfqpoint{-0.000000in}{0.000000in}}%
\pgfpathlineto{\pgfqpoint{-0.048611in}{0.000000in}}%
\pgfusepath{stroke,fill}%
}%
\begin{pgfscope}%
\pgfsys@transformshift{0.428083in}{0.919137in}%
\pgfsys@useobject{currentmarker}{}%
\end{pgfscope}%
\end{pgfscope}%
\begin{pgfscope}%
\definecolor{textcolor}{rgb}{0.000000,0.000000,0.000000}%
\pgfsetstrokecolor{textcolor}%
\pgfsetfillcolor{textcolor}%
\pgftext[x=0.229011in, y=0.887480in, left, base]{\color{textcolor}{\rmfamily\fontsize{6.000000}{7.200000}\selectfont\catcode`\^=\active\def^{\ifmmode\sp\else\^{}\fi}\catcode`\%=\active\def
\end{pgfscope}%
\begin{pgfscope}%
\pgfsetbuttcap%
\pgfsetroundjoin%
\definecolor{currentfill}{rgb}{0.000000,0.000000,0.000000}%
\pgfsetfillcolor{currentfill}%
\pgfsetlinewidth{0.803000pt}%
\definecolor{currentstroke}{rgb}{0.000000,0.000000,0.000000}%
\pgfsetstrokecolor{currentstroke}%
\pgfsetdash{}{0pt}%
\pgfsys@defobject{currentmarker}{\pgfqpoint{-0.048611in}{0.000000in}}{\pgfqpoint{-0.000000in}{0.000000in}}{%
\pgfpathmoveto{\pgfqpoint{-0.000000in}{0.000000in}}%
\pgfpathlineto{\pgfqpoint{-0.048611in}{0.000000in}}%
\pgfusepath{stroke,fill}%
}%
\begin{pgfscope}%
\pgfsys@transformshift{0.428083in}{1.200727in}%
\pgfsys@useobject{currentmarker}{}%
\end{pgfscope}%
\end{pgfscope}%
\begin{pgfscope}%
\definecolor{textcolor}{rgb}{0.000000,0.000000,0.000000}%
\pgfsetstrokecolor{textcolor}%
\pgfsetfillcolor{textcolor}%
\pgftext[x=0.229011in, y=1.169070in, left, base]{\color{textcolor}{\rmfamily\fontsize{6.000000}{7.200000}\selectfont\catcode`\^=\active\def^{\ifmmode\sp\else\^{}\fi}\catcode`\%=\active\def
\end{pgfscope}%
\begin{pgfscope}%
\pgfsetbuttcap%
\pgfsetroundjoin%
\definecolor{currentfill}{rgb}{0.000000,0.000000,0.000000}%
\pgfsetfillcolor{currentfill}%
\pgfsetlinewidth{0.803000pt}%
\definecolor{currentstroke}{rgb}{0.000000,0.000000,0.000000}%
\pgfsetstrokecolor{currentstroke}%
\pgfsetdash{}{0pt}%
\pgfsys@defobject{currentmarker}{\pgfqpoint{-0.048611in}{0.000000in}}{\pgfqpoint{-0.000000in}{0.000000in}}{%
\pgfpathmoveto{\pgfqpoint{-0.000000in}{0.000000in}}%
\pgfpathlineto{\pgfqpoint{-0.048611in}{0.000000in}}%
\pgfusepath{stroke,fill}%
}%
\begin{pgfscope}%
\pgfsys@transformshift{0.428083in}{1.482318in}%
\pgfsys@useobject{currentmarker}{}%
\end{pgfscope}%
\end{pgfscope}%
\begin{pgfscope}%
\definecolor{textcolor}{rgb}{0.000000,0.000000,0.000000}%
\pgfsetstrokecolor{textcolor}%
\pgfsetfillcolor{textcolor}%
\pgftext[x=0.229011in, y=1.450661in, left, base]{\color{textcolor}{\rmfamily\fontsize{6.000000}{7.200000}\selectfont\catcode`\^=\active\def^{\ifmmode\sp\else\^{}\fi}\catcode`\%=\active\def
\end{pgfscope}%
\begin{pgfscope}%
\pgfsetbuttcap%
\pgfsetroundjoin%
\definecolor{currentfill}{rgb}{0.000000,0.000000,0.000000}%
\pgfsetfillcolor{currentfill}%
\pgfsetlinewidth{0.803000pt}%
\definecolor{currentstroke}{rgb}{0.000000,0.000000,0.000000}%
\pgfsetstrokecolor{currentstroke}%
\pgfsetdash{}{0pt}%
\pgfsys@defobject{currentmarker}{\pgfqpoint{-0.048611in}{0.000000in}}{\pgfqpoint{-0.000000in}{0.000000in}}{%
\pgfpathmoveto{\pgfqpoint{-0.000000in}{0.000000in}}%
\pgfpathlineto{\pgfqpoint{-0.048611in}{0.000000in}}%
\pgfusepath{stroke,fill}%
}%
\begin{pgfscope}%
\pgfsys@transformshift{0.428083in}{1.763908in}%
\pgfsys@useobject{currentmarker}{}%
\end{pgfscope}%
\end{pgfscope}%
\begin{pgfscope}%
\definecolor{textcolor}{rgb}{0.000000,0.000000,0.000000}%
\pgfsetstrokecolor{textcolor}%
\pgfsetfillcolor{textcolor}%
\pgftext[x=0.178086in, y=1.732251in, left, base]{\color{textcolor}{\rmfamily\fontsize{6.000000}{7.200000}\selectfont\catcode`\^=\active\def^{\ifmmode\sp\else\^{}\fi}\catcode`\%=\active\def
\end{pgfscope}%
\begin{pgfscope}%
\definecolor{textcolor}{rgb}{0.000000,0.000000,0.000000}%
\pgfsetstrokecolor{textcolor}%
\pgfsetfillcolor{textcolor}%
\pgftext[x=0.122530in,y=1.059932in,,bottom,rotate=90.000000]{\color{textcolor}{\rmfamily\fontsize{8.000000}{9.600000}\selectfont\catcode`\^=\active\def^{\ifmmode\sp\else\^{}\fi}\catcode`\%=\active\def
\end{pgfscope}%
\begin{pgfscope}%
\pgfpathrectangle{\pgfqpoint{0.428083in}{0.355956in}}{\pgfqpoint{2.753577in}{1.407952in}}%
\pgfusepath{clip}%
\pgfsetbuttcap%
\pgfsetroundjoin%
\pgfsetlinewidth{1.505625pt}%
\definecolor{currentstroke}{rgb}{0.200000,0.400000,1.000000}%
\pgfsetstrokecolor{currentstroke}%
\pgfsetstrokeopacity{0.600000}%
\pgfsetdash{{5.550000pt}{2.400000pt}}{0.000000pt}%
\pgfpathmoveto{\pgfqpoint{0.553246in}{0.611653in}}%
\pgfpathlineto{\pgfqpoint{0.654388in}{0.867512in}}%
\pgfpathlineto{\pgfqpoint{0.780814in}{1.012838in}}%
\pgfpathlineto{\pgfqpoint{1.033668in}{1.156439in}}%
\pgfpathlineto{\pgfqpoint{1.792229in}{1.316438in}}%
\pgfpathlineto{\pgfqpoint{3.056497in}{1.438677in}}%
\pgfusepath{stroke}%
\end{pgfscope}%
\begin{pgfscope}%
\pgfpathrectangle{\pgfqpoint{0.428083in}{0.355956in}}{\pgfqpoint{2.753577in}{1.407952in}}%
\pgfusepath{clip}%
\pgfsetbuttcap%
\pgfsetmiterjoin%
\definecolor{currentfill}{rgb}{0.200000,0.400000,1.000000}%
\pgfsetfillcolor{currentfill}%
\pgfsetfillopacity{0.600000}%
\pgfsetlinewidth{0.752812pt}%
\definecolor{currentstroke}{rgb}{1.000000,1.000000,1.000000}%
\pgfsetstrokecolor{currentstroke}%
\pgfsetstrokeopacity{0.600000}%
\pgfsetdash{}{0pt}%
\pgfsys@defobject{currentmarker}{\pgfqpoint{-0.041667in}{-0.041667in}}{\pgfqpoint{0.041667in}{0.041667in}}{%
\pgfpathmoveto{\pgfqpoint{-0.020833in}{-0.041667in}}%
\pgfpathlineto{\pgfqpoint{0.000000in}{-0.020833in}}%
\pgfpathlineto{\pgfqpoint{0.020833in}{-0.041667in}}%
\pgfpathlineto{\pgfqpoint{0.041667in}{-0.020833in}}%
\pgfpathlineto{\pgfqpoint{0.020833in}{0.000000in}}%
\pgfpathlineto{\pgfqpoint{0.041667in}{0.020833in}}%
\pgfpathlineto{\pgfqpoint{0.020833in}{0.041667in}}%
\pgfpathlineto{\pgfqpoint{0.000000in}{0.020833in}}%
\pgfpathlineto{\pgfqpoint{-0.020833in}{0.041667in}}%
\pgfpathlineto{\pgfqpoint{-0.041667in}{0.020833in}}%
\pgfpathlineto{\pgfqpoint{-0.020833in}{0.000000in}}%
\pgfpathlineto{\pgfqpoint{-0.041667in}{-0.020833in}}%
\pgfpathlineto{\pgfqpoint{-0.020833in}{-0.041667in}}%
\pgfpathclose%
\pgfusepath{stroke,fill}%
}%
\begin{pgfscope}%
\pgfsys@transformshift{0.553246in}{0.611653in}%
\pgfsys@useobject{currentmarker}{}%
\end{pgfscope}%
\begin{pgfscope}%
\pgfsys@transformshift{0.654388in}{0.867512in}%
\pgfsys@useobject{currentmarker}{}%
\end{pgfscope}%
\begin{pgfscope}%
\pgfsys@transformshift{0.780814in}{1.012838in}%
\pgfsys@useobject{currentmarker}{}%
\end{pgfscope}%
\begin{pgfscope}%
\pgfsys@transformshift{1.033668in}{1.156439in}%
\pgfsys@useobject{currentmarker}{}%
\end{pgfscope}%
\begin{pgfscope}%
\pgfsys@transformshift{1.792229in}{1.316438in}%
\pgfsys@useobject{currentmarker}{}%
\end{pgfscope}%
\begin{pgfscope}%
\pgfsys@transformshift{3.056497in}{1.438677in}%
\pgfsys@useobject{currentmarker}{}%
\end{pgfscope}%
\end{pgfscope}%
\begin{pgfscope}%
\pgfpathrectangle{\pgfqpoint{0.428083in}{0.355956in}}{\pgfqpoint{2.753577in}{1.407952in}}%
\pgfusepath{clip}%
\pgfsetbuttcap%
\pgfsetroundjoin%
\pgfsetlinewidth{1.505625pt}%
\definecolor{currentstroke}{rgb}{0.600000,0.400000,0.200000}%
\pgfsetstrokecolor{currentstroke}%
\pgfsetstrokeopacity{0.600000}%
\pgfsetdash{{5.550000pt}{2.400000pt}}{0.000000pt}%
\pgfpathmoveto{\pgfqpoint{0.553246in}{0.395605in}}%
\pgfpathlineto{\pgfqpoint{0.654388in}{0.480568in}}%
\pgfpathlineto{\pgfqpoint{0.780814in}{0.585382in}}%
\pgfpathlineto{\pgfqpoint{1.033668in}{0.717276in}}%
\pgfpathlineto{\pgfqpoint{1.792229in}{0.936615in}}%
\pgfpathlineto{\pgfqpoint{3.056497in}{1.111827in}}%
\pgfusepath{stroke}%
\end{pgfscope}%
\begin{pgfscope}%
\pgfpathrectangle{\pgfqpoint{0.428083in}{0.355956in}}{\pgfqpoint{2.753577in}{1.407952in}}%
\pgfusepath{clip}%
\pgfsetbuttcap%
\pgfsetmiterjoin%
\definecolor{currentfill}{rgb}{0.600000,0.400000,0.200000}%
\pgfsetfillcolor{currentfill}%
\pgfsetfillopacity{0.600000}%
\pgfsetlinewidth{0.752812pt}%
\definecolor{currentstroke}{rgb}{1.000000,1.000000,1.000000}%
\pgfsetstrokecolor{currentstroke}%
\pgfsetstrokeopacity{0.600000}%
\pgfsetdash{}{0pt}%
\pgfsys@defobject{currentmarker}{\pgfqpoint{-0.041667in}{-0.041667in}}{\pgfqpoint{0.041667in}{0.041667in}}{%
\pgfpathmoveto{\pgfqpoint{-0.020833in}{-0.041667in}}%
\pgfpathlineto{\pgfqpoint{0.000000in}{-0.020833in}}%
\pgfpathlineto{\pgfqpoint{0.020833in}{-0.041667in}}%
\pgfpathlineto{\pgfqpoint{0.041667in}{-0.020833in}}%
\pgfpathlineto{\pgfqpoint{0.020833in}{0.000000in}}%
\pgfpathlineto{\pgfqpoint{0.041667in}{0.020833in}}%
\pgfpathlineto{\pgfqpoint{0.020833in}{0.041667in}}%
\pgfpathlineto{\pgfqpoint{0.000000in}{0.020833in}}%
\pgfpathlineto{\pgfqpoint{-0.020833in}{0.041667in}}%
\pgfpathlineto{\pgfqpoint{-0.041667in}{0.020833in}}%
\pgfpathlineto{\pgfqpoint{-0.020833in}{0.000000in}}%
\pgfpathlineto{\pgfqpoint{-0.041667in}{-0.020833in}}%
\pgfpathlineto{\pgfqpoint{-0.020833in}{-0.041667in}}%
\pgfpathclose%
\pgfusepath{stroke,fill}%
}%
\begin{pgfscope}%
\pgfsys@transformshift{0.553246in}{0.395605in}%
\pgfsys@useobject{currentmarker}{}%
\end{pgfscope}%
\begin{pgfscope}%
\pgfsys@transformshift{0.654388in}{0.480568in}%
\pgfsys@useobject{currentmarker}{}%
\end{pgfscope}%
\begin{pgfscope}%
\pgfsys@transformshift{0.780814in}{0.585382in}%
\pgfsys@useobject{currentmarker}{}%
\end{pgfscope}%
\begin{pgfscope}%
\pgfsys@transformshift{1.033668in}{0.717276in}%
\pgfsys@useobject{currentmarker}{}%
\end{pgfscope}%
\begin{pgfscope}%
\pgfsys@transformshift{1.792229in}{0.936615in}%
\pgfsys@useobject{currentmarker}{}%
\end{pgfscope}%
\begin{pgfscope}%
\pgfsys@transformshift{3.056497in}{1.111827in}%
\pgfsys@useobject{currentmarker}{}%
\end{pgfscope}%
\end{pgfscope}%
\begin{pgfscope}%
\pgfpathrectangle{\pgfqpoint{0.428083in}{0.355956in}}{\pgfqpoint{2.753577in}{1.407952in}}%
\pgfusepath{clip}%
\pgfsetbuttcap%
\pgfsetroundjoin%
\pgfsetlinewidth{1.505625pt}%
\definecolor{currentstroke}{rgb}{0.349020,0.349020,0.349020}%
\pgfsetstrokecolor{currentstroke}%
\pgfsetstrokeopacity{0.600000}%
\pgfsetdash{{5.550000pt}{2.400000pt}}{0.000000pt}%
\pgfpathmoveto{\pgfqpoint{0.553246in}{0.854134in}}%
\pgfpathlineto{\pgfqpoint{0.654388in}{1.158974in}}%
\pgfpathlineto{\pgfqpoint{0.780814in}{1.320106in}}%
\pgfpathlineto{\pgfqpoint{1.033668in}{1.441320in}}%
\pgfpathlineto{\pgfqpoint{1.792229in}{1.570517in}}%
\pgfpathlineto{\pgfqpoint{3.056497in}{1.636923in}}%
\pgfusepath{stroke}%
\end{pgfscope}%
\begin{pgfscope}%
\pgfpathrectangle{\pgfqpoint{0.428083in}{0.355956in}}{\pgfqpoint{2.753577in}{1.407952in}}%
\pgfusepath{clip}%
\pgfsetbuttcap%
\pgfsetmiterjoin%
\definecolor{currentfill}{rgb}{0.349020,0.349020,0.349020}%
\pgfsetfillcolor{currentfill}%
\pgfsetfillopacity{0.600000}%
\pgfsetlinewidth{0.752812pt}%
\definecolor{currentstroke}{rgb}{1.000000,1.000000,1.000000}%
\pgfsetstrokecolor{currentstroke}%
\pgfsetstrokeopacity{0.600000}%
\pgfsetdash{}{0pt}%
\pgfsys@defobject{currentmarker}{\pgfqpoint{-0.041667in}{-0.041667in}}{\pgfqpoint{0.041667in}{0.041667in}}{%
\pgfpathmoveto{\pgfqpoint{-0.020833in}{-0.041667in}}%
\pgfpathlineto{\pgfqpoint{0.000000in}{-0.020833in}}%
\pgfpathlineto{\pgfqpoint{0.020833in}{-0.041667in}}%
\pgfpathlineto{\pgfqpoint{0.041667in}{-0.020833in}}%
\pgfpathlineto{\pgfqpoint{0.020833in}{0.000000in}}%
\pgfpathlineto{\pgfqpoint{0.041667in}{0.020833in}}%
\pgfpathlineto{\pgfqpoint{0.020833in}{0.041667in}}%
\pgfpathlineto{\pgfqpoint{0.000000in}{0.020833in}}%
\pgfpathlineto{\pgfqpoint{-0.020833in}{0.041667in}}%
\pgfpathlineto{\pgfqpoint{-0.041667in}{0.020833in}}%
\pgfpathlineto{\pgfqpoint{-0.020833in}{0.000000in}}%
\pgfpathlineto{\pgfqpoint{-0.041667in}{-0.020833in}}%
\pgfpathlineto{\pgfqpoint{-0.020833in}{-0.041667in}}%
\pgfpathclose%
\pgfusepath{stroke,fill}%
}%
\begin{pgfscope}%
\pgfsys@transformshift{0.553246in}{0.854134in}%
\pgfsys@useobject{currentmarker}{}%
\end{pgfscope}%
\begin{pgfscope}%
\pgfsys@transformshift{0.654388in}{1.158974in}%
\pgfsys@useobject{currentmarker}{}%
\end{pgfscope}%
\begin{pgfscope}%
\pgfsys@transformshift{0.780814in}{1.320106in}%
\pgfsys@useobject{currentmarker}{}%
\end{pgfscope}%
\begin{pgfscope}%
\pgfsys@transformshift{1.033668in}{1.441320in}%
\pgfsys@useobject{currentmarker}{}%
\end{pgfscope}%
\begin{pgfscope}%
\pgfsys@transformshift{1.792229in}{1.570517in}%
\pgfsys@useobject{currentmarker}{}%
\end{pgfscope}%
\begin{pgfscope}%
\pgfsys@transformshift{3.056497in}{1.636923in}%
\pgfsys@useobject{currentmarker}{}%
\end{pgfscope}%
\end{pgfscope}%
\begin{pgfscope}%
\pgfpathrectangle{\pgfqpoint{0.428083in}{0.355956in}}{\pgfqpoint{2.753577in}{1.407952in}}%
\pgfusepath{clip}%
\pgfsetbuttcap%
\pgfsetroundjoin%
\pgfsetlinewidth{1.505625pt}%
\definecolor{currentstroke}{rgb}{0.294118,0.000000,0.431373}%
\pgfsetstrokecolor{currentstroke}%
\pgfsetstrokeopacity{0.600000}%
\pgfsetdash{{5.550000pt}{2.400000pt}}{0.000000pt}%
\pgfpathmoveto{\pgfqpoint{0.553246in}{1.032420in}}%
\pgfpathlineto{\pgfqpoint{0.654388in}{1.341684in}}%
\pgfpathlineto{\pgfqpoint{0.780814in}{1.449304in}}%
\pgfpathlineto{\pgfqpoint{1.033668in}{1.526552in}}%
\pgfpathlineto{\pgfqpoint{1.792229in}{1.621710in}}%
\pgfpathlineto{\pgfqpoint{3.056497in}{1.671933in}}%
\pgfusepath{stroke}%
\end{pgfscope}%
\begin{pgfscope}%
\pgfpathrectangle{\pgfqpoint{0.428083in}{0.355956in}}{\pgfqpoint{2.753577in}{1.407952in}}%
\pgfusepath{clip}%
\pgfsetbuttcap%
\pgfsetmiterjoin%
\definecolor{currentfill}{rgb}{0.294118,0.000000,0.431373}%
\pgfsetfillcolor{currentfill}%
\pgfsetfillopacity{0.600000}%
\pgfsetlinewidth{0.752812pt}%
\definecolor{currentstroke}{rgb}{1.000000,1.000000,1.000000}%
\pgfsetstrokecolor{currentstroke}%
\pgfsetstrokeopacity{0.600000}%
\pgfsetdash{}{0pt}%
\pgfsys@defobject{currentmarker}{\pgfqpoint{-0.041667in}{-0.041667in}}{\pgfqpoint{0.041667in}{0.041667in}}{%
\pgfpathmoveto{\pgfqpoint{-0.020833in}{-0.041667in}}%
\pgfpathlineto{\pgfqpoint{0.000000in}{-0.020833in}}%
\pgfpathlineto{\pgfqpoint{0.020833in}{-0.041667in}}%
\pgfpathlineto{\pgfqpoint{0.041667in}{-0.020833in}}%
\pgfpathlineto{\pgfqpoint{0.020833in}{0.000000in}}%
\pgfpathlineto{\pgfqpoint{0.041667in}{0.020833in}}%
\pgfpathlineto{\pgfqpoint{0.020833in}{0.041667in}}%
\pgfpathlineto{\pgfqpoint{0.000000in}{0.020833in}}%
\pgfpathlineto{\pgfqpoint{-0.020833in}{0.041667in}}%
\pgfpathlineto{\pgfqpoint{-0.041667in}{0.020833in}}%
\pgfpathlineto{\pgfqpoint{-0.020833in}{0.000000in}}%
\pgfpathlineto{\pgfqpoint{-0.041667in}{-0.020833in}}%
\pgfpathlineto{\pgfqpoint{-0.020833in}{-0.041667in}}%
\pgfpathclose%
\pgfusepath{stroke,fill}%
}%
\begin{pgfscope}%
\pgfsys@transformshift{0.553246in}{1.032420in}%
\pgfsys@useobject{currentmarker}{}%
\end{pgfscope}%
\begin{pgfscope}%
\pgfsys@transformshift{0.654388in}{1.341684in}%
\pgfsys@useobject{currentmarker}{}%
\end{pgfscope}%
\begin{pgfscope}%
\pgfsys@transformshift{0.780814in}{1.449304in}%
\pgfsys@useobject{currentmarker}{}%
\end{pgfscope}%
\begin{pgfscope}%
\pgfsys@transformshift{1.033668in}{1.526552in}%
\pgfsys@useobject{currentmarker}{}%
\end{pgfscope}%
\begin{pgfscope}%
\pgfsys@transformshift{1.792229in}{1.621710in}%
\pgfsys@useobject{currentmarker}{}%
\end{pgfscope}%
\begin{pgfscope}%
\pgfsys@transformshift{3.056497in}{1.671933in}%
\pgfsys@useobject{currentmarker}{}%
\end{pgfscope}%
\end{pgfscope}%
\begin{pgfscope}%
\pgfpathrectangle{\pgfqpoint{0.428083in}{0.355956in}}{\pgfqpoint{2.753577in}{1.407952in}}%
\pgfusepath{clip}%
\pgfsetbuttcap%
\pgfsetroundjoin%
\pgfsetlinewidth{1.505625pt}%
\definecolor{currentstroke}{rgb}{0.800000,0.000000,0.000000}%
\pgfsetstrokecolor{currentstroke}%
\pgfsetstrokeopacity{0.600000}%
\pgfsetdash{{5.550000pt}{2.400000pt}}{0.000000pt}%
\pgfpathmoveto{\pgfqpoint{0.553246in}{1.063007in}}%
\pgfpathlineto{\pgfqpoint{0.654388in}{1.337746in}}%
\pgfpathlineto{\pgfqpoint{0.780814in}{1.457395in}}%
\pgfpathlineto{\pgfqpoint{1.033668in}{1.541279in}}%
\pgfpathlineto{\pgfqpoint{1.792229in}{1.632715in}}%
\pgfpathlineto{\pgfqpoint{3.056497in}{1.686768in}}%
\pgfusepath{stroke}%
\end{pgfscope}%
\begin{pgfscope}%
\pgfpathrectangle{\pgfqpoint{0.428083in}{0.355956in}}{\pgfqpoint{2.753577in}{1.407952in}}%
\pgfusepath{clip}%
\pgfsetbuttcap%
\pgfsetmiterjoin%
\definecolor{currentfill}{rgb}{0.800000,0.000000,0.000000}%
\pgfsetfillcolor{currentfill}%
\pgfsetfillopacity{0.600000}%
\pgfsetlinewidth{0.752812pt}%
\definecolor{currentstroke}{rgb}{1.000000,1.000000,1.000000}%
\pgfsetstrokecolor{currentstroke}%
\pgfsetstrokeopacity{0.600000}%
\pgfsetdash{}{0pt}%
\pgfsys@defobject{currentmarker}{\pgfqpoint{-0.041667in}{-0.041667in}}{\pgfqpoint{0.041667in}{0.041667in}}{%
\pgfpathmoveto{\pgfqpoint{-0.020833in}{-0.041667in}}%
\pgfpathlineto{\pgfqpoint{0.000000in}{-0.020833in}}%
\pgfpathlineto{\pgfqpoint{0.020833in}{-0.041667in}}%
\pgfpathlineto{\pgfqpoint{0.041667in}{-0.020833in}}%
\pgfpathlineto{\pgfqpoint{0.020833in}{0.000000in}}%
\pgfpathlineto{\pgfqpoint{0.041667in}{0.020833in}}%
\pgfpathlineto{\pgfqpoint{0.020833in}{0.041667in}}%
\pgfpathlineto{\pgfqpoint{0.000000in}{0.020833in}}%
\pgfpathlineto{\pgfqpoint{-0.020833in}{0.041667in}}%
\pgfpathlineto{\pgfqpoint{-0.041667in}{0.020833in}}%
\pgfpathlineto{\pgfqpoint{-0.020833in}{0.000000in}}%
\pgfpathlineto{\pgfqpoint{-0.041667in}{-0.020833in}}%
\pgfpathlineto{\pgfqpoint{-0.020833in}{-0.041667in}}%
\pgfpathclose%
\pgfusepath{stroke,fill}%
}%
\begin{pgfscope}%
\pgfsys@transformshift{0.553246in}{1.063007in}%
\pgfsys@useobject{currentmarker}{}%
\end{pgfscope}%
\begin{pgfscope}%
\pgfsys@transformshift{0.654388in}{1.337746in}%
\pgfsys@useobject{currentmarker}{}%
\end{pgfscope}%
\begin{pgfscope}%
\pgfsys@transformshift{0.780814in}{1.457395in}%
\pgfsys@useobject{currentmarker}{}%
\end{pgfscope}%
\begin{pgfscope}%
\pgfsys@transformshift{1.033668in}{1.541279in}%
\pgfsys@useobject{currentmarker}{}%
\end{pgfscope}%
\begin{pgfscope}%
\pgfsys@transformshift{1.792229in}{1.632715in}%
\pgfsys@useobject{currentmarker}{}%
\end{pgfscope}%
\begin{pgfscope}%
\pgfsys@transformshift{3.056497in}{1.686768in}%
\pgfsys@useobject{currentmarker}{}%
\end{pgfscope}%
\end{pgfscope}%
\begin{pgfscope}%
\pgfsetrectcap%
\pgfsetmiterjoin%
\pgfsetlinewidth{0.803000pt}%
\definecolor{currentstroke}{rgb}{0.000000,0.000000,0.000000}%
\pgfsetstrokecolor{currentstroke}%
\pgfsetdash{}{0pt}%
\pgfpathmoveto{\pgfqpoint{0.428083in}{0.355956in}}%
\pgfpathlineto{\pgfqpoint{0.428083in}{1.763908in}}%
\pgfusepath{stroke}%
\end{pgfscope}%
\begin{pgfscope}%
\pgfsetrectcap%
\pgfsetmiterjoin%
\pgfsetlinewidth{0.803000pt}%
\definecolor{currentstroke}{rgb}{0.000000,0.000000,0.000000}%
\pgfsetstrokecolor{currentstroke}%
\pgfsetdash{}{0pt}%
\pgfpathmoveto{\pgfqpoint{3.181660in}{0.355956in}}%
\pgfpathlineto{\pgfqpoint{3.181660in}{1.763908in}}%
\pgfusepath{stroke}%
\end{pgfscope}%
\begin{pgfscope}%
\pgfsetrectcap%
\pgfsetmiterjoin%
\pgfsetlinewidth{0.803000pt}%
\definecolor{currentstroke}{rgb}{0.000000,0.000000,0.000000}%
\pgfsetstrokecolor{currentstroke}%
\pgfsetdash{}{0pt}%
\pgfpathmoveto{\pgfqpoint{0.428083in}{0.355956in}}%
\pgfpathlineto{\pgfqpoint{3.181660in}{0.355956in}}%
\pgfusepath{stroke}%
\end{pgfscope}%
\begin{pgfscope}%
\pgfsetrectcap%
\pgfsetmiterjoin%
\pgfsetlinewidth{0.803000pt}%
\definecolor{currentstroke}{rgb}{0.000000,0.000000,0.000000}%
\pgfsetstrokecolor{currentstroke}%
\pgfsetdash{}{0pt}%
\pgfpathmoveto{\pgfqpoint{0.428083in}{1.763908in}}%
\pgfpathlineto{\pgfqpoint{3.181660in}{1.763908in}}%
\pgfusepath{stroke}%
\end{pgfscope}%
\begin{pgfscope}%
\definecolor{textcolor}{rgb}{0.000000,0.000000,0.000000}%
\pgfsetstrokecolor{textcolor}%
\pgfsetfillcolor{textcolor}%
\pgftext[x=1.804872in,y=1.847242in,,base]{\color{textcolor}{\rmfamily\fontsize{8.000000}{9.600000}\selectfont\catcode`\^=\active\def^{\ifmmode\sp\else\^{}\fi}\catcode`\%=\active\def
\end{pgfscope}%
\begin{pgfscope}%
\pgfsetbuttcap%
\pgfsetmiterjoin%
\definecolor{currentfill}{rgb}{1.000000,1.000000,1.000000}%
\pgfsetfillcolor{currentfill}%
\pgfsetfillopacity{0.800000}%
\pgfsetlinewidth{1.003750pt}%
\definecolor{currentstroke}{rgb}{0.800000,0.800000,0.800000}%
\pgfsetstrokecolor{currentstroke}%
\pgfsetstrokeopacity{0.800000}%
\pgfsetdash{}{0pt}%
\pgfpathmoveto{\pgfqpoint{2.333141in}{0.397622in}}%
\pgfpathlineto{\pgfqpoint{3.123327in}{0.397622in}}%
\pgfpathquadraticcurveto{\pgfqpoint{3.139993in}{0.397622in}}{\pgfqpoint{3.139993in}{0.414289in}}%
\pgfpathlineto{\pgfqpoint{3.139993in}{1.019888in}}%
\pgfpathquadraticcurveto{\pgfqpoint{3.139993in}{1.036554in}}{\pgfqpoint{3.123327in}{1.036554in}}%
\pgfpathlineto{\pgfqpoint{2.333141in}{1.036554in}}%
\pgfpathquadraticcurveto{\pgfqpoint{2.316474in}{1.036554in}}{\pgfqpoint{2.316474in}{1.019888in}}%
\pgfpathlineto{\pgfqpoint{2.316474in}{0.414289in}}%
\pgfpathquadraticcurveto{\pgfqpoint{2.316474in}{0.397622in}}{\pgfqpoint{2.333141in}{0.397622in}}%
\pgfpathlineto{\pgfqpoint{2.333141in}{0.397622in}}%
\pgfpathclose%
\pgfusepath{stroke,fill}%
\end{pgfscope}%
\begin{pgfscope}%
\pgfsetbuttcap%
\pgfsetroundjoin%
\pgfsetlinewidth{1.505625pt}%
\definecolor{currentstroke}{rgb}{0.200000,0.400000,1.000000}%
\pgfsetstrokecolor{currentstroke}%
\pgfsetstrokeopacity{0.600000}%
\pgfsetdash{{5.550000pt}{2.400000pt}}{0.000000pt}%
\pgfpathmoveto{\pgfqpoint{2.349808in}{0.969074in}}%
\pgfpathlineto{\pgfqpoint{2.433141in}{0.969074in}}%
\pgfpathlineto{\pgfqpoint{2.516474in}{0.969074in}}%
\pgfusepath{stroke}%
\end{pgfscope}%
\begin{pgfscope}%
\pgfsetbuttcap%
\pgfsetmiterjoin%
\definecolor{currentfill}{rgb}{0.200000,0.400000,1.000000}%
\pgfsetfillcolor{currentfill}%
\pgfsetfillopacity{0.600000}%
\pgfsetlinewidth{0.752812pt}%
\definecolor{currentstroke}{rgb}{1.000000,1.000000,1.000000}%
\pgfsetstrokecolor{currentstroke}%
\pgfsetstrokeopacity{0.600000}%
\pgfsetdash{}{0pt}%
\pgfsys@defobject{currentmarker}{\pgfqpoint{-0.041667in}{-0.041667in}}{\pgfqpoint{0.041667in}{0.041667in}}{%
\pgfpathmoveto{\pgfqpoint{-0.020833in}{-0.041667in}}%
\pgfpathlineto{\pgfqpoint{0.000000in}{-0.020833in}}%
\pgfpathlineto{\pgfqpoint{0.020833in}{-0.041667in}}%
\pgfpathlineto{\pgfqpoint{0.041667in}{-0.020833in}}%
\pgfpathlineto{\pgfqpoint{0.020833in}{0.000000in}}%
\pgfpathlineto{\pgfqpoint{0.041667in}{0.020833in}}%
\pgfpathlineto{\pgfqpoint{0.020833in}{0.041667in}}%
\pgfpathlineto{\pgfqpoint{0.000000in}{0.020833in}}%
\pgfpathlineto{\pgfqpoint{-0.020833in}{0.041667in}}%
\pgfpathlineto{\pgfqpoint{-0.041667in}{0.020833in}}%
\pgfpathlineto{\pgfqpoint{-0.020833in}{0.000000in}}%
\pgfpathlineto{\pgfqpoint{-0.041667in}{-0.020833in}}%
\pgfpathlineto{\pgfqpoint{-0.020833in}{-0.041667in}}%
\pgfpathclose%
\pgfusepath{stroke,fill}%
}%
\begin{pgfscope}%
\pgfsys@transformshift{2.433141in}{0.969074in}%
\pgfsys@useobject{currentmarker}{}%
\end{pgfscope}%
\end{pgfscope}%
\begin{pgfscope}%
\definecolor{textcolor}{rgb}{0.000000,0.000000,0.000000}%
\pgfsetstrokecolor{textcolor}%
\pgfsetfillcolor{textcolor}%
\pgftext[x=2.583141in,y=0.939907in,left,base]{\color{textcolor}{\rmfamily\fontsize{6.000000}{7.200000}\selectfont\catcode`\^=\active\def^{\ifmmode\sp\else\^{}\fi}\catcode`\%=\active\def
\end{pgfscope}%
\begin{pgfscope}%
\pgfsetbuttcap%
\pgfsetroundjoin%
\pgfsetlinewidth{1.505625pt}%
\definecolor{currentstroke}{rgb}{0.600000,0.400000,0.200000}%
\pgfsetstrokecolor{currentstroke}%
\pgfsetstrokeopacity{0.600000}%
\pgfsetdash{{5.550000pt}{2.400000pt}}{0.000000pt}%
\pgfpathmoveto{\pgfqpoint{2.349808in}{0.846759in}}%
\pgfpathlineto{\pgfqpoint{2.433141in}{0.846759in}}%
\pgfpathlineto{\pgfqpoint{2.516474in}{0.846759in}}%
\pgfusepath{stroke}%
\end{pgfscope}%
\begin{pgfscope}%
\pgfsetbuttcap%
\pgfsetmiterjoin%
\definecolor{currentfill}{rgb}{0.600000,0.400000,0.200000}%
\pgfsetfillcolor{currentfill}%
\pgfsetfillopacity{0.600000}%
\pgfsetlinewidth{0.752812pt}%
\definecolor{currentstroke}{rgb}{1.000000,1.000000,1.000000}%
\pgfsetstrokecolor{currentstroke}%
\pgfsetstrokeopacity{0.600000}%
\pgfsetdash{}{0pt}%
\pgfsys@defobject{currentmarker}{\pgfqpoint{-0.041667in}{-0.041667in}}{\pgfqpoint{0.041667in}{0.041667in}}{%
\pgfpathmoveto{\pgfqpoint{-0.020833in}{-0.041667in}}%
\pgfpathlineto{\pgfqpoint{0.000000in}{-0.020833in}}%
\pgfpathlineto{\pgfqpoint{0.020833in}{-0.041667in}}%
\pgfpathlineto{\pgfqpoint{0.041667in}{-0.020833in}}%
\pgfpathlineto{\pgfqpoint{0.020833in}{0.000000in}}%
\pgfpathlineto{\pgfqpoint{0.041667in}{0.020833in}}%
\pgfpathlineto{\pgfqpoint{0.020833in}{0.041667in}}%
\pgfpathlineto{\pgfqpoint{0.000000in}{0.020833in}}%
\pgfpathlineto{\pgfqpoint{-0.020833in}{0.041667in}}%
\pgfpathlineto{\pgfqpoint{-0.041667in}{0.020833in}}%
\pgfpathlineto{\pgfqpoint{-0.020833in}{0.000000in}}%
\pgfpathlineto{\pgfqpoint{-0.041667in}{-0.020833in}}%
\pgfpathlineto{\pgfqpoint{-0.020833in}{-0.041667in}}%
\pgfpathclose%
\pgfusepath{stroke,fill}%
}%
\begin{pgfscope}%
\pgfsys@transformshift{2.433141in}{0.846759in}%
\pgfsys@useobject{currentmarker}{}%
\end{pgfscope}%
\end{pgfscope}%
\begin{pgfscope}%
\definecolor{textcolor}{rgb}{0.000000,0.000000,0.000000}%
\pgfsetstrokecolor{textcolor}%
\pgfsetfillcolor{textcolor}%
\pgftext[x=2.583141in,y=0.817593in,left,base]{\color{textcolor}{\rmfamily\fontsize{6.000000}{7.200000}\selectfont\catcode`\^=\active\def^{\ifmmode\sp\else\^{}\fi}\catcode`\%=\active\def
\end{pgfscope}%
\begin{pgfscope}%
\pgfsetbuttcap%
\pgfsetroundjoin%
\pgfsetlinewidth{1.505625pt}%
\definecolor{currentstroke}{rgb}{0.349020,0.349020,0.349020}%
\pgfsetstrokecolor{currentstroke}%
\pgfsetstrokeopacity{0.600000}%
\pgfsetdash{{5.550000pt}{2.400000pt}}{0.000000pt}%
\pgfpathmoveto{\pgfqpoint{2.349808in}{0.724445in}}%
\pgfpathlineto{\pgfqpoint{2.433141in}{0.724445in}}%
\pgfpathlineto{\pgfqpoint{2.516474in}{0.724445in}}%
\pgfusepath{stroke}%
\end{pgfscope}%
\begin{pgfscope}%
\pgfsetbuttcap%
\pgfsetmiterjoin%
\definecolor{currentfill}{rgb}{0.349020,0.349020,0.349020}%
\pgfsetfillcolor{currentfill}%
\pgfsetfillopacity{0.600000}%
\pgfsetlinewidth{0.752812pt}%
\definecolor{currentstroke}{rgb}{1.000000,1.000000,1.000000}%
\pgfsetstrokecolor{currentstroke}%
\pgfsetstrokeopacity{0.600000}%
\pgfsetdash{}{0pt}%
\pgfsys@defobject{currentmarker}{\pgfqpoint{-0.041667in}{-0.041667in}}{\pgfqpoint{0.041667in}{0.041667in}}{%
\pgfpathmoveto{\pgfqpoint{-0.020833in}{-0.041667in}}%
\pgfpathlineto{\pgfqpoint{0.000000in}{-0.020833in}}%
\pgfpathlineto{\pgfqpoint{0.020833in}{-0.041667in}}%
\pgfpathlineto{\pgfqpoint{0.041667in}{-0.020833in}}%
\pgfpathlineto{\pgfqpoint{0.020833in}{0.000000in}}%
\pgfpathlineto{\pgfqpoint{0.041667in}{0.020833in}}%
\pgfpathlineto{\pgfqpoint{0.020833in}{0.041667in}}%
\pgfpathlineto{\pgfqpoint{0.000000in}{0.020833in}}%
\pgfpathlineto{\pgfqpoint{-0.020833in}{0.041667in}}%
\pgfpathlineto{\pgfqpoint{-0.041667in}{0.020833in}}%
\pgfpathlineto{\pgfqpoint{-0.020833in}{0.000000in}}%
\pgfpathlineto{\pgfqpoint{-0.041667in}{-0.020833in}}%
\pgfpathlineto{\pgfqpoint{-0.020833in}{-0.041667in}}%
\pgfpathclose%
\pgfusepath{stroke,fill}%
}%
\begin{pgfscope}%
\pgfsys@transformshift{2.433141in}{0.724445in}%
\pgfsys@useobject{currentmarker}{}%
\end{pgfscope}%
\end{pgfscope}%
\begin{pgfscope}%
\definecolor{textcolor}{rgb}{0.000000,0.000000,0.000000}%
\pgfsetstrokecolor{textcolor}%
\pgfsetfillcolor{textcolor}%
\pgftext[x=2.583141in,y=0.695278in,left,base]{\color{textcolor}{\rmfamily\fontsize{6.000000}{7.200000}\selectfont\catcode`\^=\active\def^{\ifmmode\sp\else\^{}\fi}\catcode`\%=\active\def
\end{pgfscope}%
\begin{pgfscope}%
\pgfsetbuttcap%
\pgfsetroundjoin%
\pgfsetlinewidth{1.505625pt}%
\definecolor{currentstroke}{rgb}{0.294118,0.000000,0.431373}%
\pgfsetstrokecolor{currentstroke}%
\pgfsetstrokeopacity{0.600000}%
\pgfsetdash{{5.550000pt}{2.400000pt}}{0.000000pt}%
\pgfpathmoveto{\pgfqpoint{2.349808in}{0.602131in}}%
\pgfpathlineto{\pgfqpoint{2.433141in}{0.602131in}}%
\pgfpathlineto{\pgfqpoint{2.516474in}{0.602131in}}%
\pgfusepath{stroke}%
\end{pgfscope}%
\begin{pgfscope}%
\pgfsetbuttcap%
\pgfsetmiterjoin%
\definecolor{currentfill}{rgb}{0.294118,0.000000,0.431373}%
\pgfsetfillcolor{currentfill}%
\pgfsetfillopacity{0.600000}%
\pgfsetlinewidth{0.752812pt}%
\definecolor{currentstroke}{rgb}{1.000000,1.000000,1.000000}%
\pgfsetstrokecolor{currentstroke}%
\pgfsetstrokeopacity{0.600000}%
\pgfsetdash{}{0pt}%
\pgfsys@defobject{currentmarker}{\pgfqpoint{-0.041667in}{-0.041667in}}{\pgfqpoint{0.041667in}{0.041667in}}{%
\pgfpathmoveto{\pgfqpoint{-0.020833in}{-0.041667in}}%
\pgfpathlineto{\pgfqpoint{0.000000in}{-0.020833in}}%
\pgfpathlineto{\pgfqpoint{0.020833in}{-0.041667in}}%
\pgfpathlineto{\pgfqpoint{0.041667in}{-0.020833in}}%
\pgfpathlineto{\pgfqpoint{0.020833in}{0.000000in}}%
\pgfpathlineto{\pgfqpoint{0.041667in}{0.020833in}}%
\pgfpathlineto{\pgfqpoint{0.020833in}{0.041667in}}%
\pgfpathlineto{\pgfqpoint{0.000000in}{0.020833in}}%
\pgfpathlineto{\pgfqpoint{-0.020833in}{0.041667in}}%
\pgfpathlineto{\pgfqpoint{-0.041667in}{0.020833in}}%
\pgfpathlineto{\pgfqpoint{-0.020833in}{0.000000in}}%
\pgfpathlineto{\pgfqpoint{-0.041667in}{-0.020833in}}%
\pgfpathlineto{\pgfqpoint{-0.020833in}{-0.041667in}}%
\pgfpathclose%
\pgfusepath{stroke,fill}%
}%
\begin{pgfscope}%
\pgfsys@transformshift{2.433141in}{0.602131in}%
\pgfsys@useobject{currentmarker}{}%
\end{pgfscope}%
\end{pgfscope}%
\begin{pgfscope}%
\definecolor{textcolor}{rgb}{0.000000,0.000000,0.000000}%
\pgfsetstrokecolor{textcolor}%
\pgfsetfillcolor{textcolor}%
\pgftext[x=2.583141in,y=0.572964in,left,base]{\color{textcolor}{\rmfamily\fontsize{6.000000}{7.200000}\selectfont\catcode`\^=\active\def^{\ifmmode\sp\else\^{}\fi}\catcode`\%=\active\def
\end{pgfscope}%
\begin{pgfscope}%
\pgfsetbuttcap%
\pgfsetroundjoin%
\pgfsetlinewidth{1.505625pt}%
\definecolor{currentstroke}{rgb}{0.800000,0.000000,0.000000}%
\pgfsetstrokecolor{currentstroke}%
\pgfsetstrokeopacity{0.600000}%
\pgfsetdash{{5.550000pt}{2.400000pt}}{0.000000pt}%
\pgfpathmoveto{\pgfqpoint{2.349808in}{0.478636in}}%
\pgfpathlineto{\pgfqpoint{2.433141in}{0.478636in}}%
\pgfpathlineto{\pgfqpoint{2.516474in}{0.478636in}}%
\pgfusepath{stroke}%
\end{pgfscope}%
\begin{pgfscope}%
\pgfsetbuttcap%
\pgfsetmiterjoin%
\definecolor{currentfill}{rgb}{0.800000,0.000000,0.000000}%
\pgfsetfillcolor{currentfill}%
\pgfsetfillopacity{0.600000}%
\pgfsetlinewidth{0.752812pt}%
\definecolor{currentstroke}{rgb}{1.000000,1.000000,1.000000}%
\pgfsetstrokecolor{currentstroke}%
\pgfsetstrokeopacity{0.600000}%
\pgfsetdash{}{0pt}%
\pgfsys@defobject{currentmarker}{\pgfqpoint{-0.041667in}{-0.041667in}}{\pgfqpoint{0.041667in}{0.041667in}}{%
\pgfpathmoveto{\pgfqpoint{-0.020833in}{-0.041667in}}%
\pgfpathlineto{\pgfqpoint{0.000000in}{-0.020833in}}%
\pgfpathlineto{\pgfqpoint{0.020833in}{-0.041667in}}%
\pgfpathlineto{\pgfqpoint{0.041667in}{-0.020833in}}%
\pgfpathlineto{\pgfqpoint{0.020833in}{0.000000in}}%
\pgfpathlineto{\pgfqpoint{0.041667in}{0.020833in}}%
\pgfpathlineto{\pgfqpoint{0.020833in}{0.041667in}}%
\pgfpathlineto{\pgfqpoint{0.000000in}{0.020833in}}%
\pgfpathlineto{\pgfqpoint{-0.020833in}{0.041667in}}%
\pgfpathlineto{\pgfqpoint{-0.041667in}{0.020833in}}%
\pgfpathlineto{\pgfqpoint{-0.020833in}{0.000000in}}%
\pgfpathlineto{\pgfqpoint{-0.041667in}{-0.020833in}}%
\pgfpathlineto{\pgfqpoint{-0.020833in}{-0.041667in}}%
\pgfpathclose%
\pgfusepath{stroke,fill}%
}%
\begin{pgfscope}%
\pgfsys@transformshift{2.433141in}{0.478636in}%
\pgfsys@useobject{currentmarker}{}%
\end{pgfscope}%
\end{pgfscope}%
\begin{pgfscope}%
\definecolor{textcolor}{rgb}{0.000000,0.000000,0.000000}%
\pgfsetstrokecolor{textcolor}%
\pgfsetfillcolor{textcolor}%
\pgftext[x=2.583141in,y=0.449470in,left,base]{\color{textcolor}{\rmfamily\fontsize{6.000000}{7.200000}\selectfont\catcode`\^=\active\def^{\ifmmode\sp\else\^{}\fi}\catcode`\%=\active\def
\end{pgfscope}%
\end{pgfpicture}%
\makeatother%
\endgroup%